\documentclass[Afour,sageh,times]{sagej}

\usepackage{moreverb,url}

\usepackage[colorlinks,bookmarksopen,bookmarksnumbered,citecolor=red,urlcolor=red]{hyperref}

\usepackage[english]{babel}
\usepackage{float}
\usepackage{makecell}
\usepackage{multirow}
\usepackage{amsmath,amssymb,amsfonts}
\usepackage{algorithmic}
\usepackage{graphicx}
\usepackage{textcomp}
\usepackage{xcolor}
\usepackage{subfigure}
\usepackage{enumerate}
\usepackage{booktabs}
\usepackage{lettrine}
\usepackage{orcidlink}
\usepackage{tabularx}
\usepackage{array} 
\usepackage{booktabs} 
\usepackage{textcomp}
\usepackage{diagbox} 
\usepackage[T1]{fontenc}    
\usepackage{gensymb}
\usepackage{microtype}
\usepackage{color}

\newcommand\BibTeX{{\rmfamily B\kern-.05em \textsc{i\kern-.025em b}\kern-.08em
T\kern-.1667em\lower.7ex\hbox{E}\kern-.125emX}}

\def\volumeyear{2024}
\begin{document}

\runninghead{Wang et al.}

\title{{NeRFs} in Robotics: A Survey}


\author{Guangming~Wang\affilnum{1,*}\orcidlink{0000-0002-7675-543X}, Lei Pan\affilnum{2,*}\orcidlink{0000-0002-5342-5059}, Songyou Peng\affilnum{3}\orcidlink{0009-0007-6085-8059}, Shaohui Liu\affilnum{3}\orcidlink{0009-0006-8225-7257}, Chenfeng Xu\affilnum{4}\orcidlink{0000-0002-4941-6985}, Yanzi Miao\affilnum{2}\orcidlink{0000-0002-2688-7477}, Wei Zhan\affilnum{4}\orcidlink{0000-0002-1474-1200}, Masayoshi Tomizuka\affilnum{4}\orcidlink{0000-0003-0206-6639}, Marc Pollefeys\affilnum{3}\orcidlink{0000-0003-2448-2318}, and Hesheng Wang\affilnum{5}\orcidlink{0000-0002-9959-1634}}

\affiliation{\affilnum{1} University of Cambridge, UK. This paper was partially completed when he was visiting ETH Zurich, Switzerland.\\
\affilnum{2}China University of Mining and Technology, Xuzhou, China\\
\affilnum{3}ETH Zurich, Zurich, Switzerland\\
\affilnum{4}Mechanical Systems Control Laboratory, University of California, Berkeley, USA\\
\affilnum{5}Shanghai Jiao Tong University, Shanghai, China \\
*Authors are with equal contributions}


\corrauth{Hesheng Wang, is with the Department of Automation, Shanghai Jiao Tong University, Shanghai 200240, China and the Key Laboratory of System Control and Information Processing, Ministry of Education of China.}

\email{wanghesheng@sjtu.edu.cn}


\begin{abstract}
Detailed and realistic 3D environment representations have been a long-standing goal in the fields of computer vision and robotics. The recent emergence of neural implicit representations has introduced significant advances to these domains, enabling numerous novel capabilities. Among these, Neural Radiance Fields (NeRFs) have gained considerable attention because of their considerable representational advantages,
such as simplified mathematical models, low memory footprint, and continuous scene representations. In addition to computer vision, NeRFs have demonstrated significant potential in robotics. Thus, we present this survey to provide a comprehensive understanding of NeRFs in the field of robotics. By exploring the advantages and limitations of NeRF as well as its current applications and future potential, we aim to provide an overview of
this promising area of research. Our survey is divided into two main sections: \textit{Applications of NeRFs in Robotics} and \textit{Advances for NeRFs in Robotics}, from the perspective of how NeRF enters the field of robotics. In the first section, we introduce and analyze some works that have been or could be used in robotics for perception and interaction tasks. In the second section, we show some works related to improving NeRF's own properties, which are essential for deploying NeRFs in robotics. In the discussion section of the review, we summarize the existing challenges and provide valuable future research directions. 
\end{abstract}


\keywords{Robotics, neural radiance fields, scene understanding, scene interaction, deep learning}

\maketitle

\section{Introduction}
{Deep Learning is used as a tool to design and deploy state-of-the-art robotic systems in various fields.}
These robots are surpassing even the most experienced human experts \citep{lee2020learning, Kaufmann2023drone}. 
Neural networks are demonstrating potential by enabling robots to perform tasks more naturally and intelligently, thus changing the traditional paradigms of robot perception and motion \citep{karoly2020deep}.

\begin{figure*}[t]
	\vspace{-4mm}
	\centering
	\includegraphics[scale=0.6]{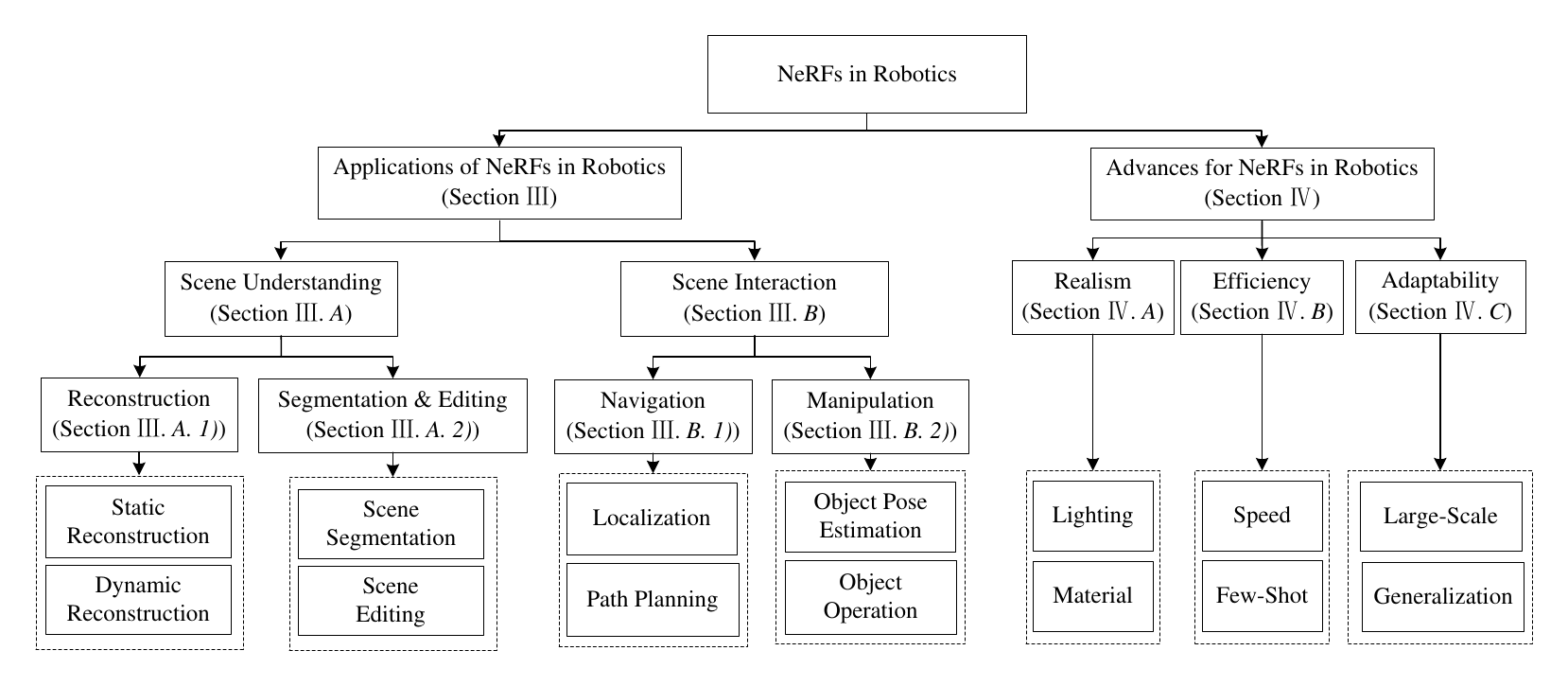}
	\vspace{-1mm}
	\caption{A taxonomy of {NeRFs} in robotics.}
	\label{Taxonomy}
\end{figure*}
\begin{figure*}[t]
	\vspace{-1mm}
	\centering
	\includegraphics[scale=0.4]{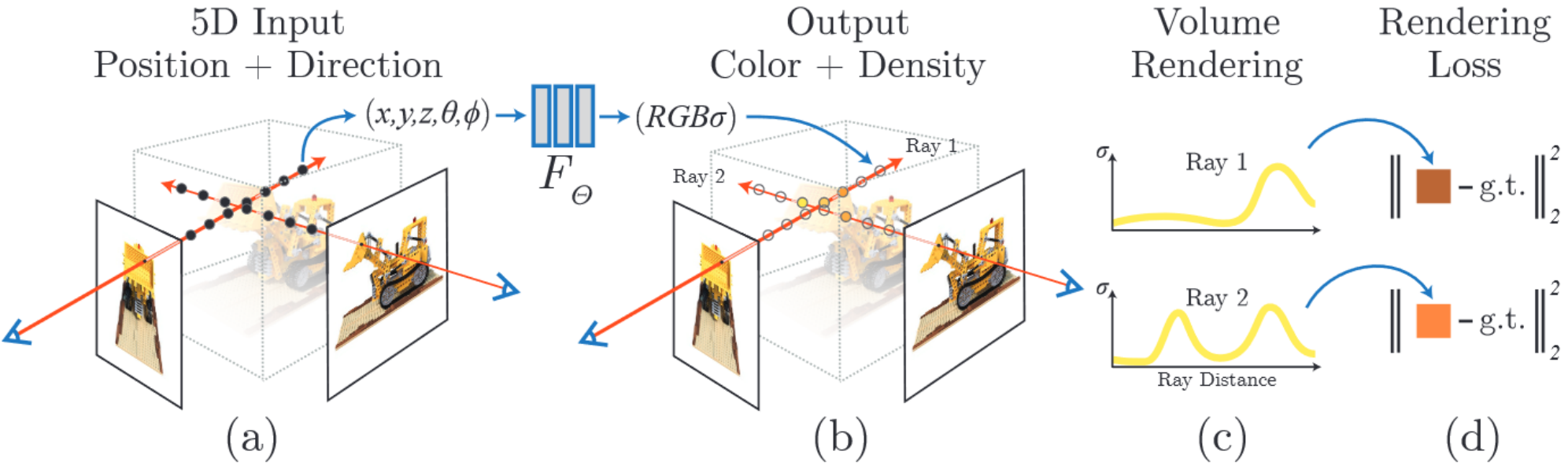}
	\vspace{-2mm}
	\caption
        {   
            The training process of NeRF. The image is sourced from \citep{mildenhall2020nerf}. 
            For each viewpoint, NeRF assumes a ray along the direction connecting the camera origin and a pixel of the target image. Multiple points are sampled along this ray in the reconstructed scene. 
            The 5D coordinates of these points (3D position $+$ 2D orientation) are input into an MLP, which outputs their corresponding colour and density values. 
            Next, the volume rendering is performed by integrating the colour and density of sampled points along a ray, producing the estimated colour of the target pixel. 
            Finally, the difference between the estimated colour and the ground truth is used to update the entire network through the rendering loss. The NeRF network is trained through this iterative process.
        }
        \label{nerf}
\end{figure*}

Neural rendering is a {family of methods} for generating images or videos 
{by combining machine learning with physical models from computer graphics.}
Neural rendering enables {generation} of realistic views while allowing explicit or implicit control of scene properties \citep{tewari2020state}.
{Neural Radiance Field} (NeRF) \citep{mildenhall2020nerf} trains a neural network whose {parameters encode} a specific implicit representation of scenes. 
Volume rendering \citep{kajiya1984ray}, {which serves as the core component of the NeRF framework}, enables NeRF to learn {a continuous 3D scene representation} from a set of 2D images with known camera poses, {and facilitates photorealistic rendering of novel views from arbitrary viewpoints}. {The remarkable capability of NeRF for novel view rendering has attracted significant interest from researchers and has inspired numerous subsequent studies} \citep{tancik2022block, zhu2022nice, adamkiewicz2022vision, maggio2022loc, shafiullah2022clip, hu2022template, zhu2022latitude, kundu2022panoptic}.
These works offer new opportunities for representing and processing perception and motion in robotics, and introduce a generalized NeRF paradigm with significant potential for robotic applications.

Since the debut of NeRF in 2020, several survey papers \citep{dellaert2020neural, xie2022neural, tewari2022advances, gao2022nerf, rabby2023beyondpixels} have been published to {highlight the rapid progress in this growing field. Among them, \cite{dellaert2020neural} presented the first survey on NeRFs in the same year as its introduction, reflecting the immediate impact and interest generated by the method.}
This concise survey {outlines} the background of NeRFs, analyzes the strengths and limitations of NeRFs, and {reviews} related work available that {proposed} extensions to various aspects of NeRFs.
Building on the previous survey \citep{tewari2020state}, \cite{tewari2022advances} supplemented recent advances in neural rendering, highlighting 3D consistency as a prominent feature in neural rendering development, particularly in methods {utilizing volumetric representations such as NeRFs}.
\cite{xie2022neural} conducted a survey that provides an extensive review of Neural Fields, {covering methods and applications.}
\cite{gao2022nerf} presented a comprehensive NeRF survey that contains several classical NeRF works as well as several typical datasets. \cite{rabby2023beyondpixels} focused on detailed summaries and comparisons of related work in terms of enhancement of NeRF attributes. Among them, \cite{xie2022neural} encompasses a wide range of background and theory knowledge. {The surveys by \cite{dellaert2020neural}, \cite{gao2022nerf}, and \cite{ rabby2023beyondpixels} focus} on {NeRFs} in various stages of development, summarizing the evolution of this field. {We recommend consulting the aforementioned works for a comprehensive and multifaceted understanding of neural fields}. 

{There has been significant adoption of NeRFs in robotics,}
with a lot of creative ideas. Unlike the focus on view synthesis in the surveys mentioned above \citep{dellaert2020neural, xie2022neural, gao2022nerf, rabby2023beyondpixels}, our survey is {positioned} within the context of robotics, providing a fresh perspective on {NeRFs}. We comprehensively introduce the {applications of NeRFs and promising related works in robotics}. In addition, we analyze {recent research efforts} to improve the performance of NeRFs for more effective deployment in robotic applications. Finally, we delve into the existing challenges within this emerging field and offer insight into future directions. {The general structure of this survey is illustrated in Figure \ref{Taxonomy}}.

Section \ref{section2} (Background) provides a brief overview of the background knowledge of NeRF, {focusing on} the core concepts and mathematical principles.
Section \ref{section3} ({Applications} of {NeRFs} in Robotics), as the main body of this survey, categorizes various application directions of NeRF in robotics. Related works are {reviewed} and meticulously analyzed. {Additionally}, we summarize the key evaluation metrics and {highlight} the achievements of some state-of-the-art (SOTA) methods.
Section \ref{section4} ({Advances of NeRFs in Robotics) introduces relevant enhancement efforts to improve the capabilities of NeRFs}. These enhancements aim to facilitate the effective deployment of {NeRFs} in robotics.
Section \ref{section5} (Discussion) {identifies} some of the challenges and future directions for NeRF in robotics as references for researchers.
Finally, Section \ref{section6} (Conclusion) provides a summary of the key findings and insights of this survey.

\section{Background} \label{section2}
\subsection{NeRF Theory}
NeRF \citep{mildenhall2020nerf} {models} a scene as a 5D vector-valued function, {approximated by} an MLP ${F_\Theta }:(\textbf{x},\textbf{d}) \to (\textbf{c},\sigma )$. 
The input to the network is a 5D vector $(x,y,z,\theta ,\phi )$, consisting of a 3D spatial coordinate $\textbf{x}=(x,y,z)$ and a 2D viewing direction $\textbf{d}=(\theta ,\phi )$. {The network outputs} an RGB color vector $\textbf{c} = (r,g,b)$ and a volume density $\sigma$. NeRF generates target images via volume rendering. 
{The entire network is trained by optimizing the learning weights $\Theta$ through comparing the rendered images and ground-truth observations.}

The NeRF training process is shown in Fig. \ref{nerf}, which is divided into four parts:
\begin{enumerate}
    \item[(a)] {NeRF assumes a set of rays originating from the camera center and passing through each pixel in the image into the scene.} A set of points are sampled along each ray. The 5D coordinates (3D position $+$ 2D orientation) of such sampled points are fed into the Multilayer Perceptron (MLP) after positional encoding.
    In the positional encoding, a set of basis functions {maps} the coordinates to a higher-dimensional space, enabling the MLP to capture high-frequency spatial information and {better represent fine-grained} scene representations.
    \item[(b)] The network outputs the volume density $\sigma $ and color $\textbf{c}$ of the sampled points. The volume density $\sigma $ is only related to the position, while the color $\textbf{c}$ is related to both the position and the viewing direction.
    \item[(c)] {Volume rendering computes the color of a target pixel by integrating the density-weighted colors of sampled points along the corresponding ray.}
    \item[(d)] The rendering loss is typically defined as the squared error between the predicted color and the ground-truth color of each target pixel, and is minimized to optimize the network parameters.
\end{enumerate}

\begin{figure*}[t]
	\vspace{-4mm}
	\centering
	\includegraphics[scale=0.628]{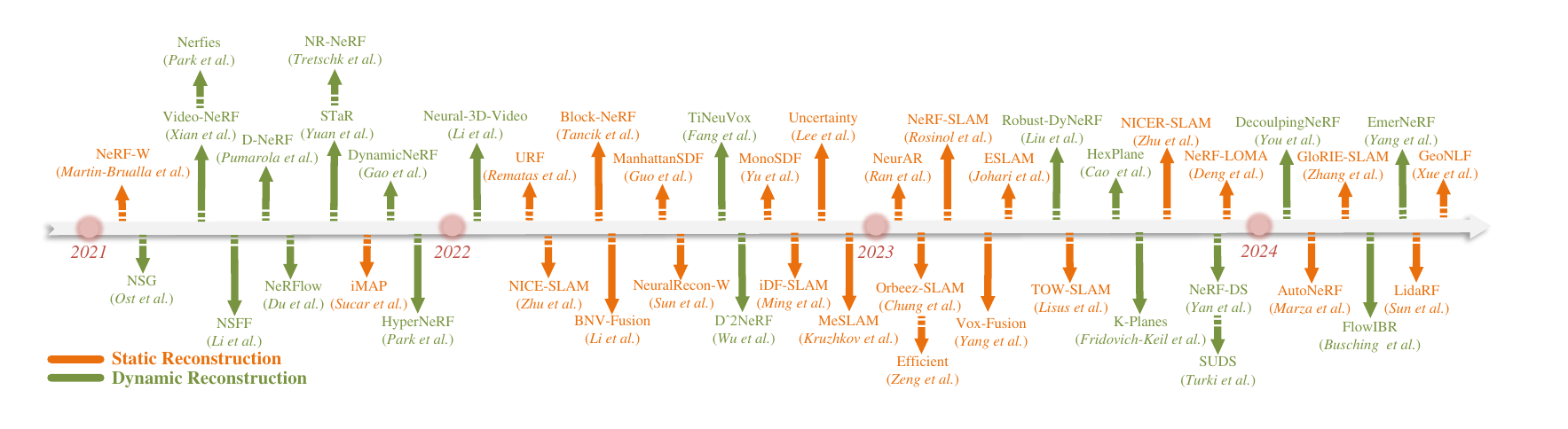}
	\vspace{-1mm}
	\caption{Chronological: {NeRFs} for 
                Scene Reconstruction in Section \ref{reconstruction}.}
	\label{Reconstruction}
\end{figure*}

{Specifically, volume rendering performs integration along each ray by accumulating the color contributions of all sampled points, weighted by their densities and visibility, to compute the final pixel value in the target image along the viewing direction} $\textbf{d}$:
\begin{equation}
    C(\textbf{r}) = \int_{{t_n}}^{{t_f}} {T(t)\sigma (\textbf{r}(t))\textbf{c}(\textbf{r}(t),\textbf{d})dt}, \label{1}    
\end{equation}

where ${t_n}$ and ${t_f}$ are near and far bounds of the camera ray $\textbf{r}(t) = \textbf{o} + t\textbf{d}$. $T(t)$ is calculated as the transmittance that the ray can travel from ${t_n}$ to $t$:
\begin{equation}
    T(t) = \exp \left( { - \int_{{t_n}}^t {\sigma (\textbf{r}(s))ds} } \right). \label{2}
\end{equation}

{Due to the discrete nature of point sampling, NeRF approximates the ideal continuous volume integration using a discrete formulation as follows:}
\begin{equation}
    \mathop C\limits^ \wedge  (\textbf{r}) = \sum\limits_{i = 1}^N {{T_i}{\alpha _i}{\textbf{c}_i}},\; where\; {T_i} = \exp \left( { - \sum\limits_{j = 1}^{i - 1} {{\sigma _j}{\delta _j}} } \right),  \label{3}
\end{equation}
where alpha values ${\alpha _i} = (1 - \exp ( - {\sigma _i}{\delta _i}))$. ${\delta _i} = {t_{i + 1}} - {t_i}$ is the distance between adjacent samples.

{Building on this design, NeRF incorporates two additional techniques: positional encoding to enhance representation quality, and hierarchical volume sampling to improve computational efficiency.}

A positional encoding, defined as ${F_\Theta } = {F'_\Theta } \circ \gamma $, uses $\gamma (p)$ to map
the input vector into a high-dimensional space to better represent the high-frequency changes in color and geometry of the scene:
\begin{equation}
 \begin{aligned}
    \gamma (p) =& (\sin ({2^0}\pi p),\cos ({2^0}\pi p),\\
                &...,\sin ({2^{L - 1}}\pi p),\cos ({2^{L - 1}}\pi p)), 
\end{aligned}   
\end{equation}
where $L$ is a hyperparameter. In NeRF \citep{mildenhall2020nerf}, $L = 10$ is used for $\gamma(\textbf{x})$, and $L = 4$ for $\gamma(\textbf{d})$. Note that $\gamma(\textbf{x})$ is injected into the network at the beginning of MLP and $\gamma(\textbf{d})$ is injected close to the end, which has been shown to {mitigate} degenerate solutions \citep{zhang2020nerf++}.

{Hierarchical volume rendering employs a coarse-to-fine strategy, where ${N_c}$ points are first sampled coarsely to generate an initial rendering. This coarse result then guides fine sampling to select ${N_f}$ fine-level points. The goal is to focus sampling on regions that contribute more significantly to the final pixel color.}

\begin{figure}[t]
	\centering
	\subfigure[Scene reconstruction using the indoor dataset]{
		\begin{minipage}[t]{0.95\linewidth}
			\centering
			\includegraphics[width=1\linewidth]{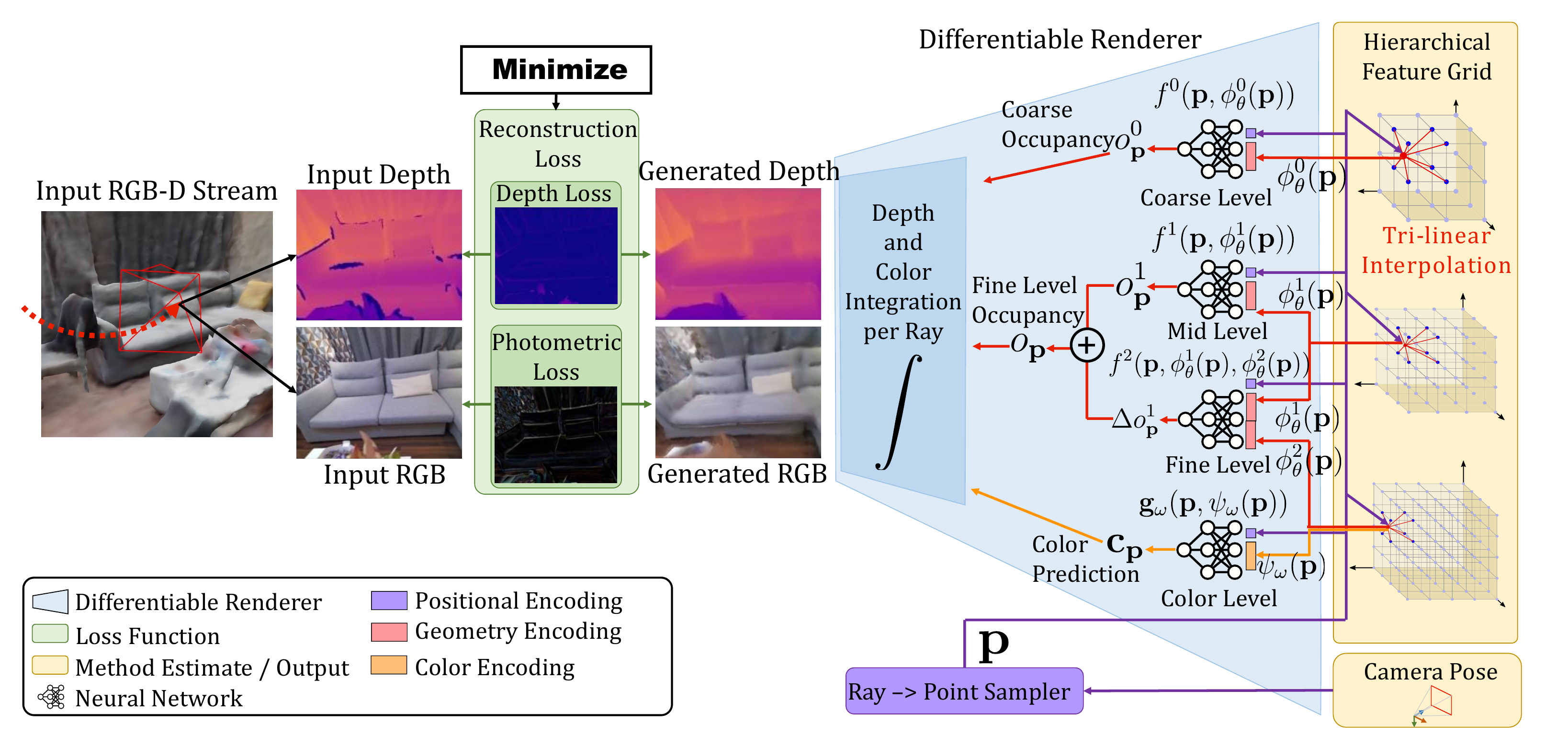}
                \label{Indoor_Static_Reconstruction}
		\end{minipage}
	}
        \subfigure[Scene reconstruction using the outdoor dataset]{
		\begin{minipage}[t]{0.95\linewidth}
			\centering
			\includegraphics[width=1\linewidth]{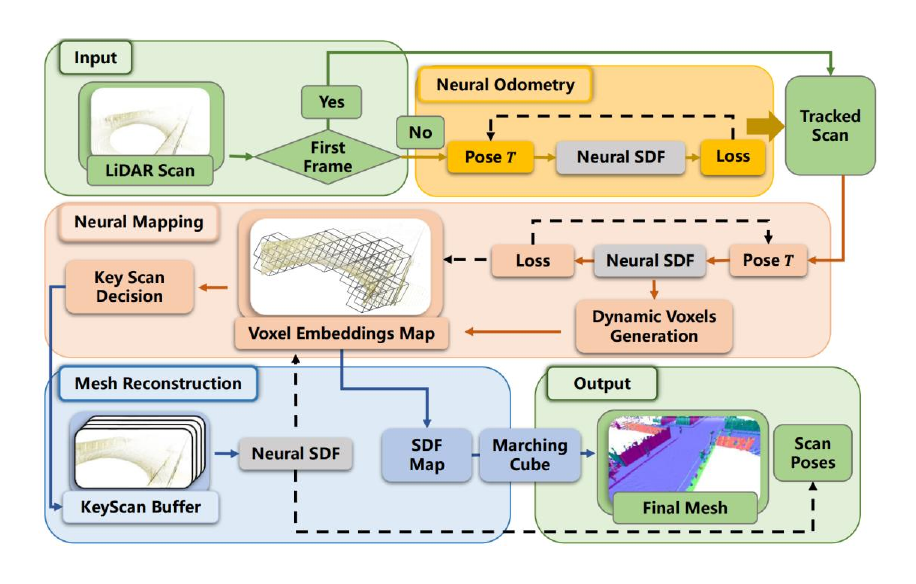}
                \label{Outdoor_Static_Reconstruction}
		\end{minipage}
	}
	\caption{An illustration of NeRF for static reconstruction. Fig. \ref{Indoor_Static_Reconstruction} and Fig. \ref{Outdoor_Static_Reconstruction} are originally shown in \citep{zhu2022nice} and \citep{deng2023nerf}, respectively.}
	\label{static_reconstruction}
\end{figure}

Finally, the loss function for hierarchical volume rendering is {defined as follows}:
\begin{equation}
    L = \sum\limits_{\textbf{r} \in R} {\left[ {\left\| {\mathop {{{\mathop C\limits^ \wedge  }_c}(\textbf{r}) - C(\textbf{r})}\limits^{} } \right\|_2^2 + \left\| {{{\mathop C\limits^ \wedge  }_f}(\textbf{r}) - C(\textbf{r})} \right\|_2^2} \right]},  
\end{equation}
where $R$ is the set of rays, and $C(\textbf{r})$ is the ground truth color, and ${{{\mathop C\limits^ \wedge  }_c}(\textbf{r})}$ and ${{{\mathop C\limits^ \wedge}_f}(\textbf{r})}$ are predicted colors from the coarse network and the fine network. 

In particular, the NeRF research community provides powerful open-source toolkits, such as Nerfstudio \citep{tancik2023nerfstudio} and NerfBridge \citep{yu2023nerfbridge}, to facilitate code development for researchers. NerfStudio \citep{tancik2023nerfstudio} offers a modular framework for NeRF development. Furthermore, NerfBridge \citep{yu2023nerfbridge} developed an interface between NerfStudio and {the Robot Operating System (ROS), enabling online robotic applications through real-time transmission of image and pose streams for training NeRF models on robotic platforms}.

\section{{Applications} of {NeRFs} in Robotics} \label{section3}
{The advantages of NeRFs, including their capabilities to facilitate simplified mathematical models, compact environment storage, and continuous scene representations, make them significantly suitable for robotics applications.}
These capabilities play a crucial role in achieving scene understanding in robotics and in completing specific tasks through interaction with the environment.

\subsection{Scene Understanding}  
\subsubsection{Reconstruction} \label{reconstruction}
We categorize the related work into static and dynamic reconstruction and present them using a timeline, as illustrated in Fig. \ref{Reconstruction}. 

\emph{(a) Static Reconstruction: } 
Scene reconstruction in robotics refers to the process of modeling a 3D representation of the environment by analyzing perceived sensor data. 
{In the context of scene reconstruction, research efforts can be broadly categorized into two groups based on the type of environment: indoor scenes (e.g., rooms) and outdoor scenes (e.g., roads), as illustrated in Fig. \ref{static_reconstruction}.}

In works utilizing indoor datasets,
iMAP \citep{sucar2021imap} {integrates an MLP architecture with a volumetric density representation, inspired by NeRF \citep{mildenhall2020nerf}, for Simultaneous Localization and Mapping (SLAM) tasks. By leveraging loss-guided sampling and a replay buffer mechanism, iMAP achieves competitive SLAM performance using only 2D images as input.}
However, the limited capacity of the MLP structure results in {issues such as catastrophic forgetting and slow inference, thereby restricting the scalability and efficiency of scene reconstruction.  In addition, volumetric density is a probabilistic representation and suffers from appearance-geometry ambiguity \citep{zhang2020nerf++}, which can result in low-precision reconstructions.}

To expand the scale of reconstruction, MeSLAM \citep{kruzhkov2022meslam} employs a multi-MLP structure to represent different parts of the scene. {In contrast, NICE-SLAM \citep{zhu2022nice} introduces a coarse-to-fine feature grid representation to extend iMAP’s capability from single-room to multi-room reconstruction.} Vox-Fusion \citep{yang2022vox} uses a tree-like structure to store grid embeddings, allowing dynamic allocation of new spatial voxels as the scene expands. \cite{lisus2023towards} demonstrates that {incorporating depth uncertainty and motion information can improve SLAM accuracy, and that a spherical background model can be employed to extend the scale of reconstructed scenes.} To enhance reconstruction efficiency, ESLAM \citep{johari2023eslam} replaces feature grids with perpendicular feature planes aligned on the multi-scale axis, reducing the growth of the scene scale from cubic to quadratic.

Orbeez-SLAM \citep{chung2023orbeez} and NeRF-SLAM \citep{rosinol2022nerf} utilize existing SLAM odometry modules for localization, improving efficiency. GloRIE-SLAM \citep{zhang2024glorie} is an RGB-only SLAM system that {leverages optical flow to integrate local and global Bundle Adjustment (BA), enabling accurate pose estimation and learning of adaptable neural point cloud representations.} The system merges predicted monocular depth priors with noisy depth maps obtained during tracking to compensate for the absence of geometric priors. 
Following BA optimization, the flexible neural point cloud updates according to the poses and depths of the keyframes. Global pose consistency is ensured by employing loop closure detection and online global BA.
{To evaluate the accuracy of pose estimation, precise ground-truth poses of robots and target objects are typically obtained using dedicated pose tracking systems, such as motion capture (MoCap) setups.}
The MoCap system uses observation devices to digitally track and re-encode the motion of objects in space, commonly by employing infrared cameras to capture the motion trajectories of specific markers on the target \citep{menolotto2020motion}.

In works utilizing outdoor datasets,
\cite{sun2022neural} employ appearance embeddings, such as NeRF-W \citep{martin2021nerf}, to model appearance variation and propose a combination of voxel-guided sampling and surface-guided sampling to improve efficiency in large-scale scenes. 
{Block-NeRF \citep{tancik2022block} partitions large-scale scenes into multiple spatially bounded and concatenated blocks to model long streets with complex intersections. The contribution of each block to rendering a target novel view is modulated by learned visibility weights.} \cite{rematas2022urban} fuse LiDAR data with image data and introduce a series of LiDAR-based losses to improve reconstruction quality.
NeRF-LOMA \citep{deng2023nerf} is a NeRF-based pure-LiDAR SLAM designed for outdoor driving environments. NeRF-LOMA incorporates a neural Signed Distance Function (SDF) that optimizes a neural implicit decoder to decode neural implicit embeddings within octree grids into SDF values. By minimizing SDF errors, NeRF-LOMA simultaneously optimizes the embeddings, poses, and decoder, ultimately enabling the reconstruction of dense smooth mesh maps.
Similarly, LidaRF \citep{sun2024lidarf} employs 3D sparse convolution to extract geometric features from point clouds and constructs a grid-based representation. Additionally, LidaRF generates augmented training data through LiDAR projections and trains geometric prediction using a robust depth supervision scheme.
GeoNLF \citep{xue2024geonlf} is a hybrid framework that alternates between global neural reconstruction and pure geometric pose optimization. By leveraging rich geometric features from LiDAR point clouds, GeoNLF incorporates an additional chamfer loss on inter-frame point cloud {correspondences, complementing standard BA optimization and photometric supervision, to jointly optimize camera poses and enhance mapping quality.}

Recently, {Truncated Signed Distance Function} (TSDF) and active scene reconstruction techniques based on the NeRF architecture have {achieved notable advances}.

{Unlike the volumetric density representation in vanilla NeRF, the TSDF encodes the distance from a sample point to the nearest surface, thereby enabling more explicit geometric reconstruction. TSDF-based methods recover surfaces by extracting the zero-level set, naturally capturing scene geometry with high sharpness and accuracy \citep{newcombe2011kinectfusion, bylow2013real}.}
However, the classical volume rendering formula is not directly applicable to TSDF. Fortunately, some recent rendering techniques are available that can be adapted to TSDF representations \citep{oechsle2021unisurf,wang2021neus,azinovic2022neural,yariv2021volume,or2022stylesdf}. 
In conjunction with these advances in rendering techniques, MonoSDF \citep{yu2022monosdf} integrates a general pre-trained monocular geometric prediction network, which predicts depth and normals as geometric priors, into neural implicit SDF surface reconstruction. {\cite{guo2022neural} improve SDF reconstruction quality in low-texture indoor regions by incorporating semantic guidance and leveraging the Manhattan world assumption.} BNV-Fusion \citep{li2022bnv} introduces a bilateral neural volumetric fusion algorithm that combines depth image features extracted at both local and global {scales}. The global geometry is supervised using the SDF loss. {IDF-SLAM \citep{ming2022idf} employs a pre-trained feature-based neural tracker \citep{el2021unsupervisedr} in combination with a neural implicit mapper that learns a TSDF-based scene representation.} Vox-Fusion \citep{yang2022vox} employs voxel feature embedding as input, generating RGB and SDF values as output. NICER-SLAM \citep{zhu2023nicer} replaces occupancy with TSDF in NICE-SLAM \citep{zhu2022nice} to achieve improved performance. 

Active scene reconstruction technologies aim to explore methods for empowering robots to actively select data that maximize benefits, thereby achieving a more intelligent reconstruction process. 
\cite{lee2022uncertainty} select the next observation view that can most effectively reduce uncertainty by estimating the volume uncertainty.
In NeurAR \citep{ran2023neurar}, pixel colors are modeled as {Gaussian-distributed random variables to explicitly represent observation uncertainty.} The uncertainty is directly associated with the Peak Signal-to-Noise Ratio (PSNR) metric and can be used as a proxy to measure the quality of candidate viewpoints.
{\cite{zeng2023efficient} propose an active reconstruction strategy that plans camera trajectories based on information gain, which is evaluated by comparing the current viewpoint with the partial 3D reconstruction accumulated so far.} 
AutoNeRF \citep{marza2023autonerf} uses a modular policy exploration approach to learn robotic autonomous data collection strategies, with scene semantics as evaluation criterion.

\begin{figure}[t]
	\centering
	\subfigure[Deformation-based dynamic reconstruction]{
		\begin{minipage}[t]{1\linewidth}
			\centering
			\includegraphics[width=0.9\linewidth]{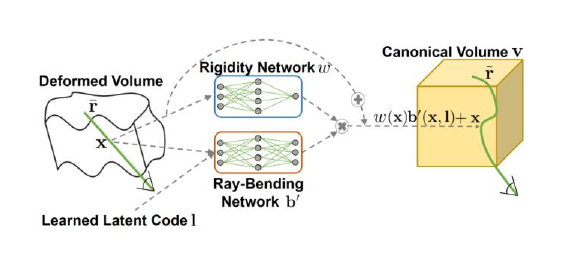}
                \label{Deformation-based}
		\end{minipage}
	}
        \subfigure[Flow-based dynamic reconstruction]{
		\begin{minipage}[t]{1\linewidth}
			\centering
			\includegraphics[width=0.9\linewidth]{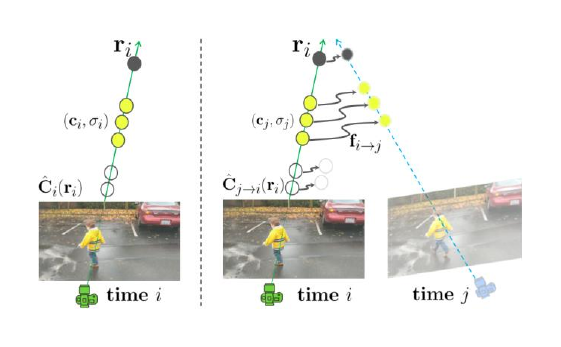}
                \label{Flow-based}
		\end{minipage}
	}
	\caption{An illustration of NeRF for dynamic reconstruction. Fig. \ref{Deformation-based} and Fig. \ref{Flow-based} are originally shown in \citep{tretschk2021non} and \citep{li2021neural}, respectively.}
	\label{dynamic_reconstruction}
\end{figure}

\emph{(b) Dynamic Reconstruction: } 
Long-term running robots usually face dynamic changes in complex environments. For the vanilla NeRF model based on static scene assumptions, dynamics undoubtedly disrupt the learning process, causing artifacts. {Moreover, in dynamic scenes, each moment provides only a single observation, resulting in a severe lack of spatial consistency constraints across different viewpoints.} Therefore, the NeRF-based models must be extended or learned differently in dynamic environments. Related works are as illustrated in Fig. \ref{dynamic_reconstruction}.

\begin{figure*}[t]
	\vspace{-4mm}
	\centering
	\includegraphics[scale=0.7]{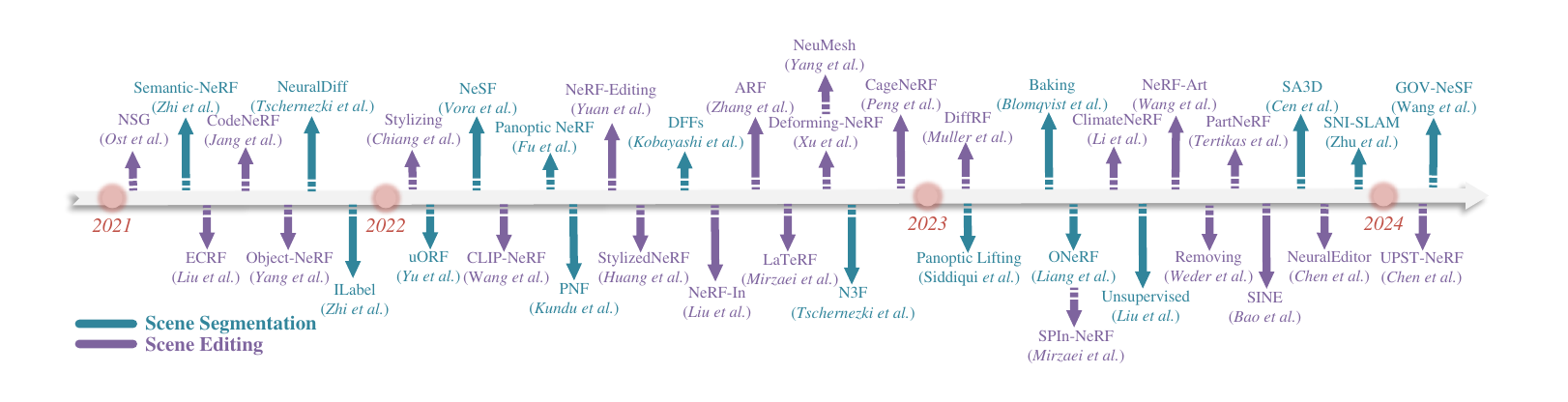}
	\vspace{-1mm}
	\caption{Chronological: NeRF for Scene Segmentation and Editing in Section \ref{segmentation_editing}.}
	\label{Segmentation_Editing}
\end{figure*}

{In the early stages of exploration, scene dynamics are modeled in an end-to-end manner by conditioning NeRF on additional inputs, such as time or camera pose transformations.}
STaR \citep{yuan2021star} models a rigidly dynamic NeRF to represent a single moving object within a scene and optimizes time-dependent rigid poses to track motion. 
To build the dynamic field, \cite{xian2021space} convert the original 3D spatial coordinates to 4D spatio-temporal coordinates. 
{DyNeRF \citep{li2022neural} employs time-dependent latent codes rather than explicit time inputs to model the dynamic field, enabling better representation of topological changes and transient effects.}
\citet{ost2021neural} build a dynamic scene representation {using a graph-based structure, where each leaf node corresponds to a local radiance field. Furthermore, objects belonging to the same category share the weights of their respective local fields.}

{As research progresses, dynamic representations based on deformation fields and motion flow are increasingly adopted to model scene dynamics, leading to improved reconstruction accuracy and temporal consistency.}

These deformation-based works \citep{pumarola2021d, tretschk2021non, park2021nerfies, park2021hypernerf, yan2023nerf, wu2022d, fang2022fast, liu2023robust} represent motion as deformations of the observed space relative to a multiframe consistent canonical space represented by a static field. 
The calculated deformations by deformation fields finely reflect local changes in the scene, including non-rigid deformations.
D-NeRF \citep{pumarola2021d} defines the canonical space based on the first frame.  The deformation network, conditioned on time, learns the displacements of ray sampling points in the observed space relative to the canonical space. In NR-NeRF \citep{tretschk2021non}, the canonical space is not predefined but is instead learned jointly from all observed frames. In addition, NR-NeRF employs time-based implicit encoding instead of directly inputting time for better rendering quality. 
NeRFies \citep{park2021nerfies} utilize a dense $SE(3)$ field to model scene deformations instead of using a displacement field, and introduces elastic energy constraints to alleviate ambiguities in optimization induced by motion. 
HyperNeRF \citep{park2021hypernerf} represents the scene in a hyperspace for topological variations, where each frame observation corresponds to a 3D NeRF as a slice of the hyperspace.  Based on HyperNeRF, NeRF-DS \citep{yan2023nerf} addresses the under-parameterization of reflections in dynamic specular objects by conditioning the color prediction branch on object surface positions and rotated surface normals.
To further enhance the quality of deformation-based dynamic scene representation, 
D$^{2}$NeRF \citep{wu2022d} introduces a shadow field to learn a shadow ratio for the static NeRF for rendering shadow variations.  
RoDynRF \citep{liu2023robust} learns deformation NeRFs while jointly estimating camera poses and focal lengths, achieving tracking in dynamic scenes that are difficult to achieve with the classical method COLMAP \citep{schonberger2016structure}. 
TiNeuVox \citep{fang2022fast} employs an explicit structure of time-sensitive neural voxels to improve efficiency, replacing the time-consuming feature inference process with a querying process.

Unlike deformations, flow is more commonly used to reflect the overall motion of objects in the scene, where some works \citep{li2021neural, gao2021dynamic, yang2023emernerf, turki2023suds, you2023decoupling,busching2024flowibr} use scene flow, while one work \citep{du2021neural} uses velocity flow.
NSFF \citep{li2021neural} predicts the scene flow and the occlusion weights between the current frame with both forward and backward frames. 
\cite{gao2021dynamic} separately model static NeRF and dynamic NeRF based on foreground masks. The dynamic NeRF predicts forward and backward scene flows while predicting a blending weight for mixing the results of dynamic and static NeRFs.
EmerNeRF \citep{yang2023emernerf} self-supervises the separation of static and dynamic scene components.
At the same time, EmerNeRF predicts 3D scene flow aggregating temporal displacement features to enhance cross-observation consistency for dynamic components.
SUDS \citep{turki2023suds} models static NeRF, dynamic NeRF, and far-field NeRF to adapt to large-scale dynamic urban scenes. The dynamic NeRF estimates 3D scene flow, which is projected onto the image plane and supervised by 2D optical flows predicted by DINO \citep{caron2021emerging}. 
{To eliminate the reliance on precomputed 2D optical flow,} \citet{you2023decoupling} propose surface consistency and patch-based multiview constraints as unsupervised regularization terms to {jointly} learn decoupled object motion and camera motion.
In addition, FlowIBR \citep{busching2024flowibr} combines a generalizable novel view synthesis model, pre-trained on a large corpus of static scenes, {with a scene-specific flow field learned for each dynamic scene}. The flow field extends the applicability of the epipolar line projection constraint between the source observations with target views for dynamic scenes. 
Unlike scene flows, \cite{du2021neural} predict velocity flows of sampled points, which are then integrated to predict future spatial positions of points in upcoming frames. 

In addition, the K-Planes \citep{fridovich2023k} and HexPlane \citep{cao2023hexplane} utilize six adaptive spatio-temporal feature planes to capture representations of dynamic environments effectively. This approach not only guarantees exceptional rendering quality for new view synthesis in dynamic scenarios but also markedly cuts down on training duration and memory usage.

\emph{(c) Conclusion for Reconstruction:} In summary, the evolution of NeRF-based reconstruction techniques in robotics shows a shift from small-scale, scene-specific methods to scalable, adaptive methods.

For static reconstruction, volumetric and TSDF methods provide early dense mapping, but face scalability and ambiguity issues. Neural implicit methods, like NeRF and neural SDFs, improve surface detail but require innovations such as multi-MLP \citep{kruzhkov2022meslam}, voxel grids \citep{yang2022vox}, and hierarchical feature planes \citep{johari2023eslam} to scale to large indoor and outdoor scenes. Accurate pose estimation, whether through SLAM or MoCap systems, remains fundamental for reliable reconstruction.

In dynamic reconstruction, early time-conditioned NeRFs are extended by deformation-based methods to model motions, and flow-based approaches to enhance temporal consistency. Recent works \citep{fridovich2023k,cao2023hexplane} further combine explicit spatio-temporal grids to balance quality and computational cost, which is crucial for real-time robotics applications. Overall, these trends reflect a move toward real-time, generalizable, and robot-oriented reconstruction systems capable of long-term robot operation in unstructured dynamic environments.

\subsubsection{Segmentation \& Editing} \label{segmentation_editing}
The {chronological development} for scene segmentation and editing is illustrated in Fig. \ref{Segmentation_Editing}.

\begin{figure}[t]
	\centering
        \subfigure[Semantic segmentation]{
		\begin{minipage}[t]{0.95\linewidth}
			\centering
			\includegraphics[width=1\linewidth]{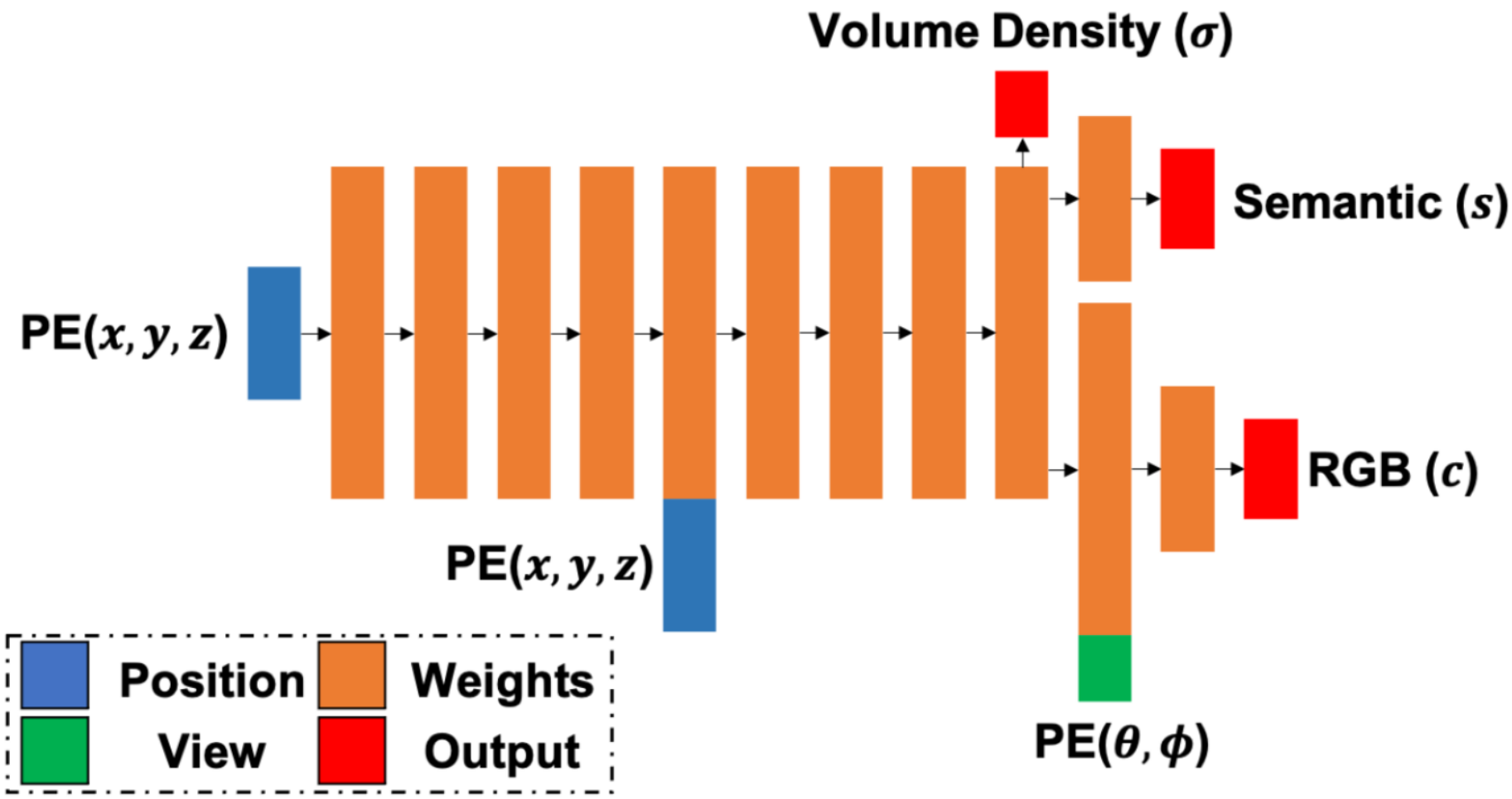}
                \label{Semantic_Segmentation}
		\end{minipage}
	}
	\subfigure[Instance segmentation]{
		\begin{minipage}[t]{0.98\linewidth}
			\centering
		\includegraphics[width=1\linewidth]{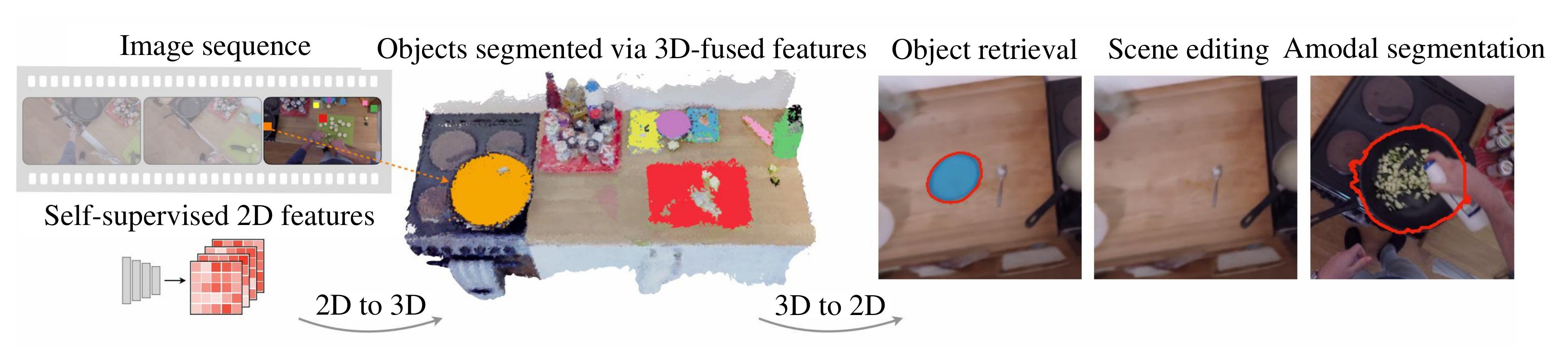}
                \label{Object_Segmentation}
		\end{minipage}
	}
         \subfigure[Panoptic segmentation]{
		\begin{minipage}[t]{0.95\linewidth}
			\centering
    		\includegraphics[width=1\linewidth]{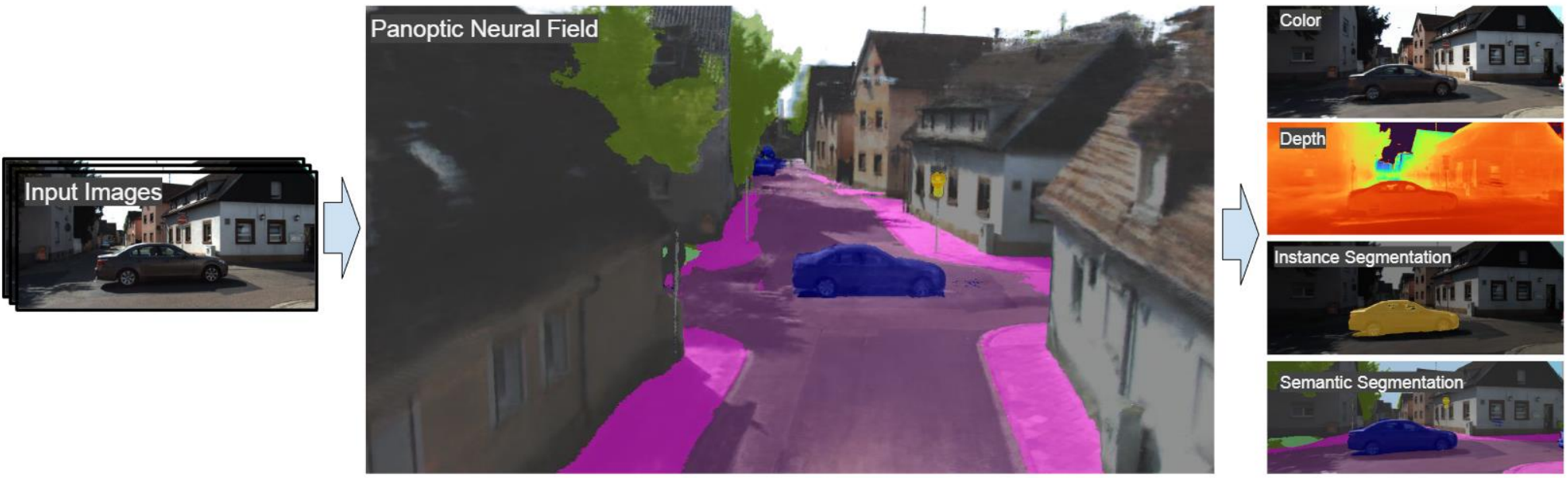}
                \label{Panoptic_Segmentation}
		\end{minipage}
	}
	\caption{An illustration of NeRF for scene segmentation. Fig. \ref{Semantic_Segmentation}, Fig. \ref{Object_Segmentation}, and Fig. \ref{Panoptic_Segmentation} are originally shown in \citep{zhi2021place}, \citep{tschernezki2021neuraldiff}, and \citep{kundu2022panoptic}, respectively.}
	\label{scene_segmentation}
\end{figure}

\emph{(a) Scene Segmentation: } 
Scene segmentation refers to partitioning a perceived scene into distinct components based on purpose-specific tasks.
Scene segmentation enhances a robot’s ability to accurately perceive and understand the surrounding environment. By identifying distinct scene components, scene segmentation facilitates goal-specific tasks such as object manipulation and navigation. Compared to 2D segmentation, 3D segmentation is better aligned with the operational demands of real-world robotic applications.
NeRF presents an innovative approach to supervise 3D segmentation from 2D posed images. Based on segmentation goals, the related work is classified into three groups: \emph{semantic segmentation}, \emph{instance segmentation}, and \emph{panoptic segmentation}, as illustrated in Fig. \ref{scene_segmentation}.

Semantic segmentation divides the scene into different components by assigning a semantic label to each 3D point.
Semantic-NeRF \citep{zhi2021place} integrates an additional semantic head alongside colour and density heads, allowing for the estimation of semantics at sampled points.  
To achieve generic semantic segmentation capability, NeSF \citep{vora2021nesf} trains a multi-scene shared 3D UNet \citep{cciccek20163d} to encode the pre-trained density field of NeRF, along with training a semantic MLP to decode features into semantic information. 
Generalization is achieved {by training on large-scale semantically labeled datasets, which requires high-quality annotations to ensure effectiveness.} To reduce the reliance on precise pixel-level semantic labels, iLabel \citep{zhi2021ilabel} and \cite{blomqvist2022baking} introduce methods for semantic segmentation using only sparse semantic labels from users. iLabel \citep{zhi2021ilabel} integrates a semantic prediction branch on top of iMAP \citep{sucar2021imap} to achieve online interactive 3D semantic SLAM. \cite{blomqvist2022baking} improve the quality of upstream features by baking pre-trained feature extractors on a large amount of data. \citet{liu2022unsupervised} {propose} a self-supervised semantic segmentation {framework comprising a segmentation model trained continuously across scenes and a set of scene-specific semantic-NeRF models \citep{zhi2021place}.} The segmentation model provides {pseudo ground-truth labels to supervise the training of the semantic-NeRF models. In turn, the consistency among semantic-NeRF models is leveraged to refine the semantic labels, further improving the segmentation model through iterative training.}
SNI-SLAM \citep{zhu2023sni} integrates multi-level features from colour, geometry, and semantics by feature interaction and collaboration, achieving more accurate results, including colour rendering, geometry representation, and semantic segmentation.
GOV-NeSF \citep{wang2024gov} uses only 2D images and utilizes LSeg \citep{li2022language}, an open-vocabulary 2D semantic segmentation model, for the extraction of semantic features. Then, GOV-NeSF \citep{wang2024gov} introduces a multiview joint fusion module to integrate texture and semantic features, {along with a cross-view attention module to model inter-view dependencies and aggregate multiview information.}

Instance segmentation aims to precisely delineate individual object instances within a scene, and its results are often used for object-level modeling or scene composition in novel view synthesis.
In this context, uORF \citep{yu2021unsupervised} leverages object-centric latent representations extracted from a single image to condition the training of a shared NeRF model in an unsupervised manner, enabling controllable rendering outputs such as instance-level segmentation and scene composition.
In robotic tasks, it is often necessary to focus on specific objects in a scene rather than all instances. When segmentation is performed by focusing solely on a designated object, instance segmentation can be referred to as object segmentation.
ONeRF \citep{liang2022onerf} achieves unsupervised object segmentation using iteratively clustering of features and 3D consistency of NeRF to generate accurate masks.
\cite{kobayashi2022decomposing} and N3F \citep{tschernezki2022neural} {employ a teacher-student distillation framework}, where semantic attributes are extracted by a 2D teacher network, such as CLIP \citep{radford2021learning}, LSeg \citep{li2022languagedriven}, or DINO \citep{caron2021emerging}. SA3D \citep{cen2023segment} combines the segmentation capability of SAM \citep{kirillov2023segany} with the 3D mask propagation capability of NeRF to segment the desired 3D models. {Mask-based} inverse rendering and cross-view self-prompting are iteratively applied across different novel views to progressively generate a detailed 3D object mask.
To complete object segmentation of egocentric videos, NeuralDiff \citep{tschernezki2021neuraldiff} incorporates inductive biases and employs a triple stream neural rendering network to segment the background, foreground and actor.

Panoptic segmentation can be understood as a combination of instance segmentation and semantic segmentation \citep{cheng2020panoptic,kirillov2019panoptic}, where all instances are segmented while assigned semantic labels and instance labels. 
Panoptic NeRF \citep{fu2022panoptic} is designed for outdoor driving scenes (e.g., KITTI-360 \citep{liao2022kitti}), assuming available 2D pseudo-semantic labels and 3D bounding primitives.
Panoptic NeRF \citep{fu2022panoptic} {constructs} dual semantic fields{: a fixed semantic field that enhances geometry estimation, and a learnable semantic field that refines semantic estimation.} Additionally, 3D bounding primitives {are introduced to provide supplementary 3D semantic supervision, helping suppress noise in pseudo-labels and facilitating instance-level annotation.}
PNF \citep{kundu2022panoptic} {replaces a shared MLP with instance-specific lightweight MLPs to represent individual foreground objects, removing the need for explicit object encodings. This design enables independent semantic prediction and object pose estimation, facilitating the tracking of object motions.} Each object is {modeled separately, and the resulting instance masks are combined with semantic segmentation outputs to achieve panoptic segmentation.}
Panoptic Lifting \citep{siddiqui2023panoptic} introduces a novel approach for acquiring a full 3D volume depiction from in-the-wild images, utilizing only 2D panoptic segmentation masks derived from pre-trained models. This technique operates on a neural field to craft coherent 3D panoptic representations that are unified and consistent across multiple views.

\begin{figure}[t]
	\centering
	\subfigure[Object appearance and geometry editing]{
		\begin{minipage}[t]{0.95\linewidth}
			\centering
			\includegraphics[width=1\linewidth]{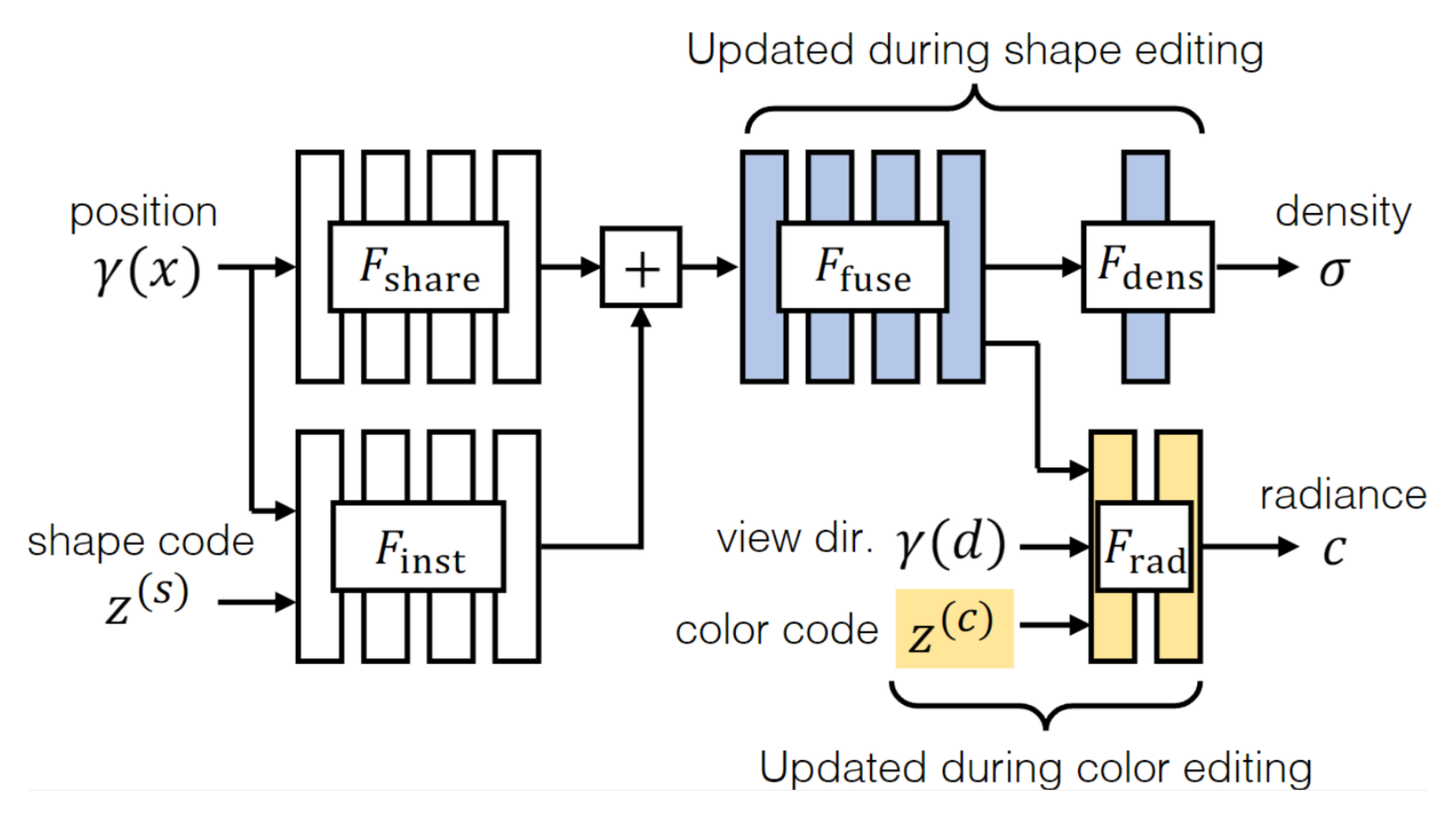}
                \label{Object_Appearance_and_Geometry_Editing}
		\end{minipage}
	}
        \subfigure[Object insertion and erasure editing]{
		\begin{minipage}[t]{0.95\linewidth}
			\centering
			\includegraphics[width=1\linewidth]{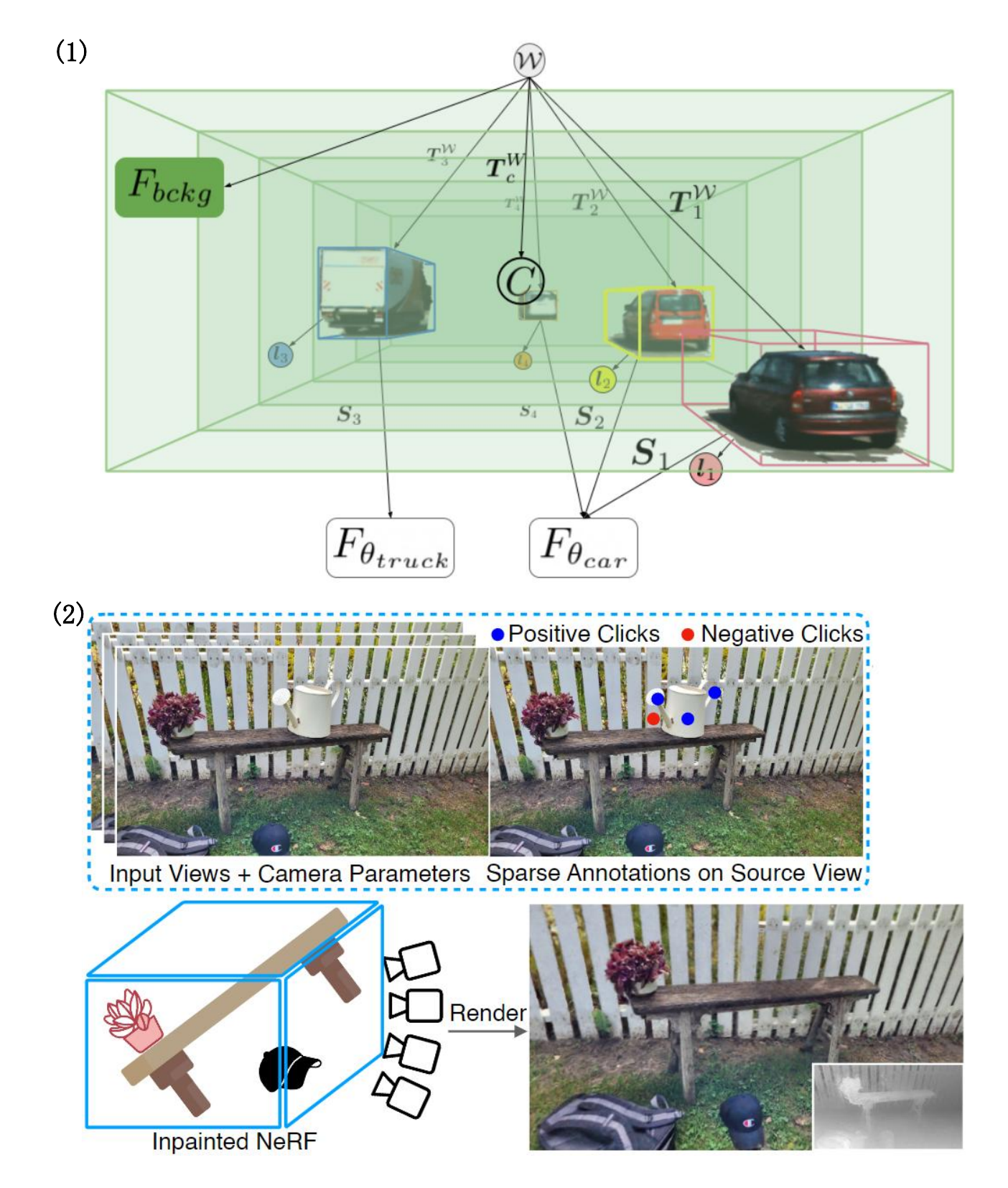}
                \label{Object_Insertion_and_Erasure_Editing}
		\end{minipage}
	}
         \subfigure[Scene stylization editing]{
		\begin{minipage}[t]{0.95\linewidth}
			\centering
    		\includegraphics[width=1\linewidth]{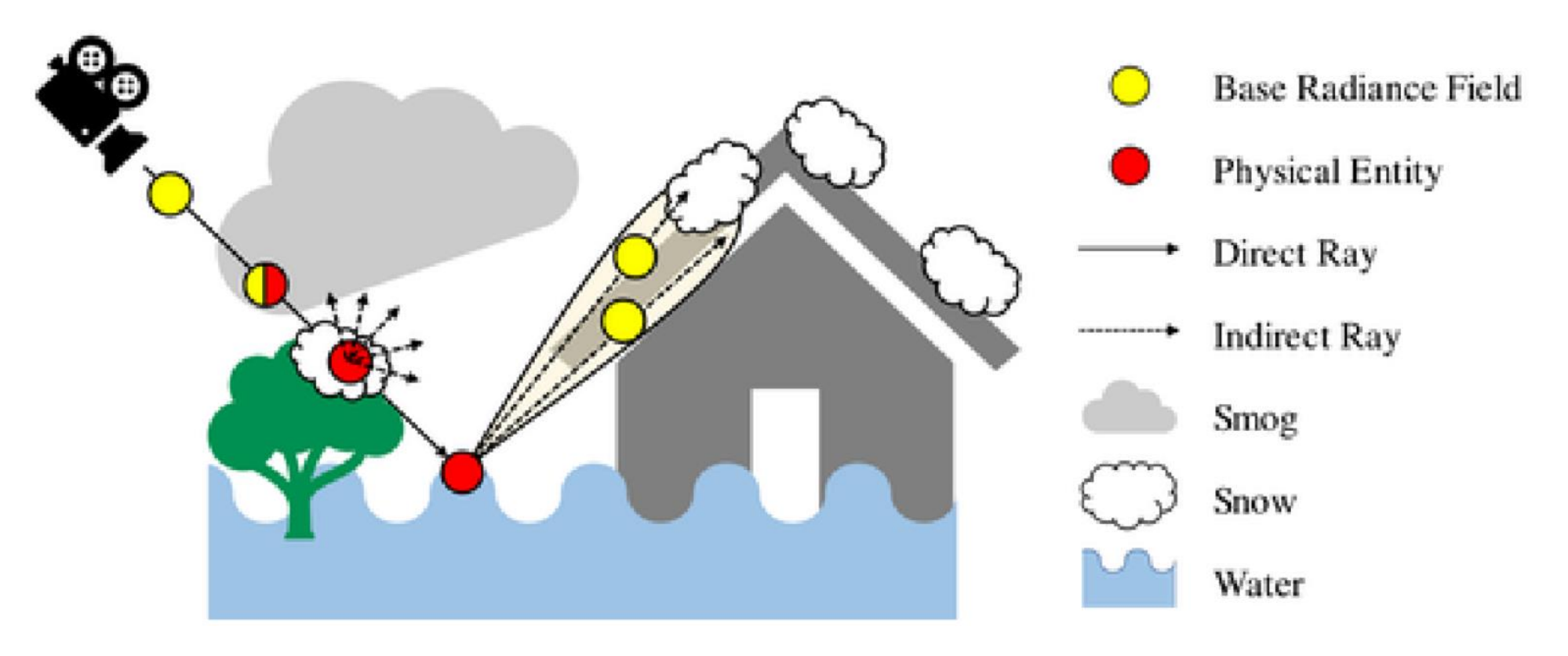}
                \label{Scene_Stylization_Editing}
		\end{minipage}
	}
	\caption{An illustration of NeRF for scene editing. Fig. \ref{Object_Appearance_and_Geometry_Editing} and Fig. \ref{Scene_Stylization_Editing} are originally shown in \citep{liu2021editing} and \citep{li2023climatenerf}, respectively, and Fig. \ref{Object_Insertion_and_Erasure_Editing} in order (1) and (2), sequentially correspond to \citep{ost2021neural} and \citep{mirzaei2023spin}.}
	\label{scene_editing}
\end{figure}

\emph{(b) Scene Editing: } \label{section-scene_editing}
Scene editing refers to the process of modifying scene content based on the prompts provided by the user to achieve the desired effects.
The edited scenes can serve as a source of training data for robots, and these data are often hard or time-consuming to collect in the real world.
NeRF plays a crucial role in enhancing the reality and 3D consistency of the edited results.
We categorize related works into \emph{object appearance and geometry editing}, \emph{object insertion and erasure editing}, and \emph{scene stylization editing}, depending on the editing objectives, as illustrated in Fig. \ref{scene_editing}.

To achieve appearance and geometry editing, a common approach is to construct appearance and geometry encodings as inputs to conditional NeRF. It is worth noting that to avoid mutual interference between appearance and geometry editing, both conditions should be disentangled.
To this end, CodeNeRF \citep{jang2021codenerf} learns to disentangle object shape and appearance encodings as conditions while learning NeRF weights. CodeNeRF achieves editing by adjusting ideal encodings. In addition to modifying the corresponding encodings, EditNeRF \citep{liu2021editing} simultaneously updates the weights of specified layers.
Without {reliance on a fixed prompt model,} CLIP-NeRF \citep{wang2022clip} leverages the multimodal {capabilities} of CLIP \citep{radford2021learning} to {guide the generation of appearance and geometry through} text prompts or image exemplars. SINE \citep{bao2023sine} employs a prior-guided editing field to adjust spatial point coordinates and colours for semantic-driven editing.
To {enable localized} editing of objects, PartNeRF \citep{tertikas2023partnerf} assigns {to each object part a NeRF representation defined within a local coordinate frame.} Each NeRF {representation is controlled by partial encodings derived from the global shape} and appearance codes.

The works mentioned above have effectively demonstrated the realism of implicit representations in editing tasks. However, it is challenging to achieve precise geometry editing using only implicit representations. Integrating implicit representations into the framework of explicit models is a promising direction that can mitigate this issue.
\cite{xu2022deforming} and CageNeRF \citep{peng2022cagenerf} both assume that a coarse polygonal mesh cage encloses objects. \cite{xu2022deforming} {perform} deformation by manipulating the cage vertices, {whereas} CageNeRF \citep{peng2022cagenerf} learns a network that takes the original cage and a novel pose as inputs to generate the deformed cage.
NeRF-Editing \citep{yuan2022nerf} {employs} the classical mesh deformation technique \citep{sorkine2007rigid} to {enable} users to {directly} edit the mesh representation derived from the density field of the canonical NeRF. {These edits are then used to compute the corresponding deformation} of the canonical space {for novel view rendering.}
NeuMesh \citep{yang2022neumesh} employs a mesh-based representation {in which} learnable geometry and appearance encodings, {along with} sign indicators {for positional identification,} are stored in the {mesh vertices.} Geometry and appearance are edited by {adjusting} the mesh vertices and updating the encodings {using the} corresponding decoders.
NeuralEditor \citep{chen2023neuraleditor} introduces a point-cloud-guided NeRF model based on a K-D tree structure, {enabling editing through the manipulation of the point cloud}. In this context, geometric editing is {defined} as the movement of each point in the point cloud to its final position. Simultaneously, the {Infinite Surface Transformation} (IST) is {proposed} to {adjust} the viewing direction of each point, {ensuring} the correct direction-appearance correspondence.

Object insertion and erasure editing involve {the flexible addition of} new objects or {the removal of} existing ones from a scene, while {preserving} scene coherence and harmony.
\cite{ost2021neural} achieve object insertion and erasure by adding and deleting {the corresponding} leaf nodes in the scene graph. LaTeRF \citep{mirzaei2022laterf} extracts interesting objects by introducing an additional output head to regress the probability of each point belonging to interesting objects. For occluded components, LaTeRF utilizes CLIP \citep{radford2021learning} to fill the gaps by {incorporating} semantic priors.
\cite{yang2021learning} construct a framework {consisting of} a scene branch and an object branch while maintaining a library of object activation codes. During rendering, Yang et al. select and switch the corresponding codes at the target position to control object movement, insertion, and erasure.
NeRF-In \citep{liu2022nerf}  updates a pre-trained NeRF model to achieve object erasure by using edited RGB-D priors guided by user-drawn erasure masks.
SPIn-NeRF \citep{mirzaei2023spin} {further employs} a semantic NeRF model to refine the erasure masks {ensuring} globally consistent object erasure.
On the other hand, \cite{weder2023removing} introduce confidence in the RGB-D views guided by masks, {selecting} views that ensure accurate painting and multiview consistency {for training} the object erasure NeRF. 
DiffRF \citep{muller2023diffrf} {employs} a denoising diffusion probabilistic model to construct NeRF based on a {well-defined} voxel grid structure. To reduce ambiguity during rendering, this method {incorporates} a volume rendering loss to the noise prediction equation, resulting in {improved} rendering outputs. In the process of modifying feature regions, DiffRF applies masks to the altered zones, and then reconstructs new shapes and appearances in the hidden areas using a completion strategy.

Stylization editing {generates diverse} stylistic scene data in response to style prompts. This can reduce overall data collection {time} and {enhance} the robustness of trained systems.
ClimateNeRF \citep{li2023climatenerf} achieves realistic rendering in {various} climate styles, such as fog, snow, and flooding, by {integrating} the instant-NGP framework \citep{muller2022instant} with physics simulation techniques. 
Moreover,  while these works \citep{chen2022upst, huang2022stylizednerf, wang2022nerf} primarily {focus on} artistic stylization, it is worth {investigating} relevant adaptations to generate style-specific data for robots.

\emph{(c) Conclusion for Segmentation \& Editing:} The advancements in NeRF-based scene segmentation and editing are {enhancing} robotic systems with richer perception and interaction capabilities.

Early 3D semantic segmentation methods extended NeRF with semantic heads or shared encoders but required large-scale datasets with semantic labels. Later approaches addressed label sparsity using sparse supervision \citep{zhi2021ilabel, blomqvist2022baking}, self-supervision \citep{liu2022unsupervised}, and open-vocabulary models \citep{wang2024gov}, thereby improving adaptability across scenes. 
Instance segmentation has evolved through various approaches, including unsupervised object discovery \citep{liang2022onerf}, teacher-student distillation \citep{kobayashi2022decomposing, tschernezki2022neural}, and 2D-to-3D mask propagation \citep{cen2023segment}, {facilitating} object-centric robotic tasks {such as} manipulation and rearrangement. 
Panoptic segmentation combines semantic and instance cues for comprehensive scene understanding, which is crucial for mobile robots in cluttered environments.

In scene editing, research has evolved from disentangling appearance and geometry to integrating implicit and explicit models (e.g., meshes, cages, point clouds) to enable controllable and physically plausible modifications. These editing techniques provide robots with access to diverse, augmented, and stylized training data, {facilitating} simulation-to-real transfer and robust policy learning. Overall, these trends highlight a {shift towards} more flexible, data-efficient, and robot-adaptive scene understanding and manipulation frameworks.

\subsection{Scene Interaction}
\begin{figure*}[t]
	\vspace{-4mm}
	\centering
	\includegraphics[scale=0.538]{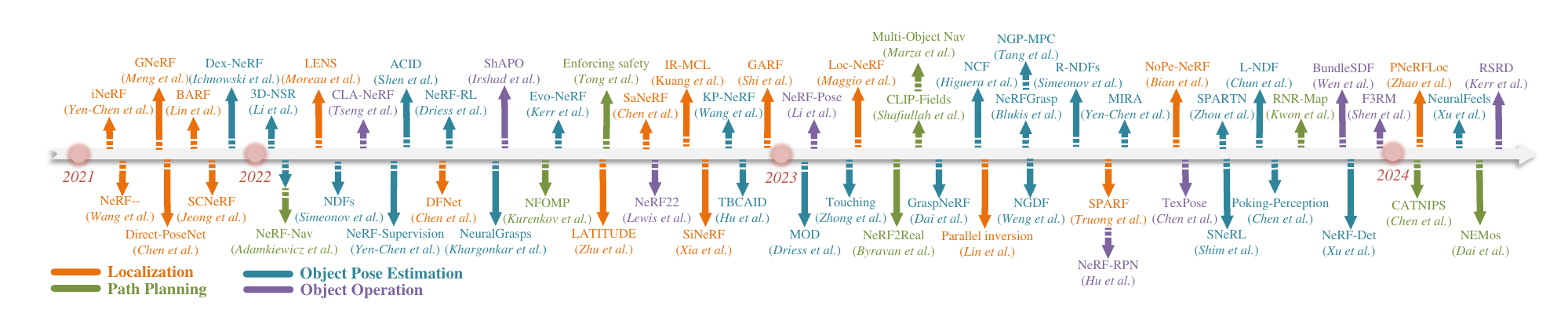}
	\vspace{-1mm}
	\caption{Chronological: NeRF for Robotic Navigation in  Section \ref{navigation} and Manipulation in Section \ref{manipulation}.}
	\label{Navigation_Manipulation}
\end{figure*}

Navigation and manipulation are typical scenarios in which robots interact with their environment or humans. The timeline of related work is depicted in Fig. \ref{Navigation_Manipulation}.

\begin{figure}[t]
	\centering
	\subfigure[Known map-based localization]{
		\begin{minipage}[t]{0.95\linewidth}
			\centering
			\includegraphics[width=1\linewidth]{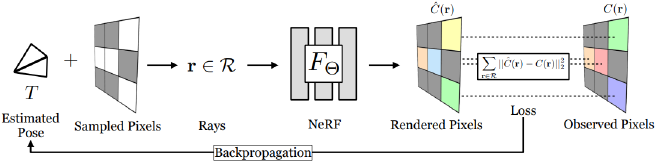}
                \label{KnownMap}
		\end{minipage}
	}
        \subfigure[Unknown map-based localization]{
		\begin{minipage}[t]{0.95\linewidth}
			\centering
			\includegraphics[width=1\linewidth]{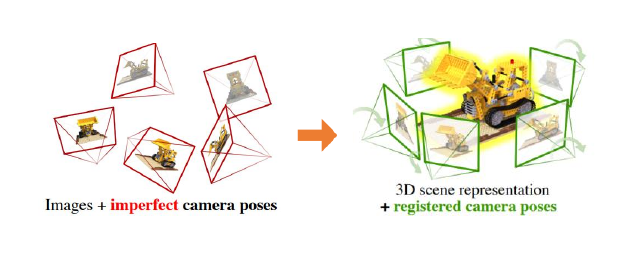}
                \label{UnKnownMap}
		\end{minipage}
	}
	\caption{An illustration of NeRF for Robotic Localization. Fig. \ref{KnownMap} \citep{yen2021inerf} makes a trained NeRF as the map, and Fig. \ref{UnKnownMap} \citep{lin2021barf} optimizes the camera pose and the model properties jointly.}
	\label{Localization}
\end{figure}

\subsubsection{Navigation} \label{navigation}
The core components of navigation include localization and path planning. Localization addresses the question of {the robot's current position}, while path planning addresses {how the robot reaches its destination}.

\emph{(a) Localization: }
Localization involves estimating the pose with 6 degrees of freedom (position and orientation) through the analysis of sensor data. 
Based on the presence or absence of a prior environment map, these localization approaches can be categorized into two classes: \emph{Known Map-based Localization} and \emph{Unknown Map-based Localization}, as shown in Fig. \ref{Localization}.

In the context of NeRF-based known map-based localization, the maps typically involve pretrained NeRF or extended NeRF models. iNeRF \citep{yen2021inerf} represents a milestone work as it is the first to regress camera poses using the implicit representation of NeRF. iNeRF introduces an inverse NeRF architecture and uses pixel-level photometric loss to optimize initial rendering poses based on the trained NeRF model. Subsequently, Direct-PoseNet \citep{chen2021direct} leverages a NeRF model to generate training data for Absolute Pose Regression (APR) networks. LENS \citep{moreau2022lens} positions multiple virtual cameras in high-density areas identified by the NeRF-W model \citep{martin2021nerf} to expand the training data space for APR models. 
{To enhance drone localization in city-scale environments, LATITUDE \citep{zhu2022latitude} first estimates coarse poses using an APR network trained with posed image data generated by the pre-trained Mega-NeRF \citep{Turki_2022_CVPR}, and subsequently refines these coarse poses using an inverse NeRF architecture.}
DFNet \citep{chen2022dfnet} optimizes an APR network to enhance robustness to illumination changes by minimizing the matching error between feature maps generated by histogram-assisted NeRF and those extracted by feature extractors. 

Another category of methods \citep{kuang2022ir,maggio2022loc,lin2023parallel} achieves global robot localization in implicit scene maps by combining the traditional Monte Carlo Localization \citep{dellaert1999monte}. These methods define pose estimation as a posterior probability estimation problem, modeling the posterior probability distribution as the distribution of weighted spatial particles. They iteratively update the particle weights and perform particle resampling based on the {discrepancy} between perception and the map until convergence to the correct pose. IR-MCL \citep{kuang2022ir} trains a neural occupancy field as the scene map and updates particle weights by comparing rendered 2D LiDAR scans with real LiDAR scan data. Loc-NeRF \citep{maggio2022loc} {directly learns} a general NeRF model as the map and calculates the particle weights using photometric differences. \cite{lin2023parallel} implement parallel processing of multiple Monte Carlo sampling processes based on the Instant-NGP model \citep{muller2022instant} to improve localization efficiency.
\cite{adamkiewicz2022vision} formulate the pose optimization problem as recursive Bayesian estimation based on iNeRF \citep{yen2021inerf}, outperforming iNeRF in rotation, translation, and velocity estimation while achieving lower variance.
 
When robots explore a new environment, the {lack} of reference maps poses a significant challenge {for} localization. In addition to {several methods} introduced in Section \ref{reconstruction} {that estimate} the robot's pose, some approaches estimate camera poses using NeRFs without {requiring explicit scene reconstruction}. 

NeRF$--$ \citep{wang2021nerf} jointly learns the representation of the environment and camera poses from 2D images. BARF \citep{lin2021barf} draws inspiration from classical 2D image alignment methods and extends the alignment concept to 3D space. SiNeRF \citep{xia2022sinerf} leverages the inherent smoothness of SIREN-MLP \citep{sitzmann2020implicit},  mitigating the risk of getting trapped in local optima. GARF \citep{shi2022garf} explores Gaussian activation functions, achieving higher pose estimation accuracy and improving network learning.
GNeRF \citep{meng2021gnerf} employs the NeRF model as a generator and trains it using a GAN-based approach. The pose-image pairs generated by the trained NeRF are used to train an inversion network that regresses to coarse poses. These coarse poses are further refined through photometric losses.
SCNeRF \citep{SCNeRF2021} jointly learns the scene model and camera parameters through geometric and photometric losses.
NoPe-NeRF \citep{bian2022nopenerf} incorporates additional constraints by learning undistorted depth maps. 
SPARF \citep{truong2023sparf} introduces {a} multi-view correspondence loss and {a} depth consistency loss. The {multi-view} correspondence loss enforces that corresponding pixels across {multiple views} {are back-projected} to the same 3D spatial point. 
The depth consistency loss ensures consistency between the depth of the trained viewpoint and the depth of unseen viewpoints, which are obtained by warping from the trained viewpoint.
PNeRFLoc \citep{zhao2024pnerfloc} is an integrated framework for visual localization that {employs} a point-based representation. The process begins with estimating the initial pose {via 2D–3D feature point matching}, followed by refining this pose using a rendering-centric optimization technique. In the pose estimation phase, PNeRFLoc introduces a feature adaptation module designed to reconcile the differences between the features utilized in visual localization and those employed in neural rendering. 

\begin{figure}[t]
	\centering
	\subfigure[NeRF-based path planning]{
		\begin{minipage}[t]{0.95\linewidth}
			\centering
			\includegraphics[width=1\linewidth]{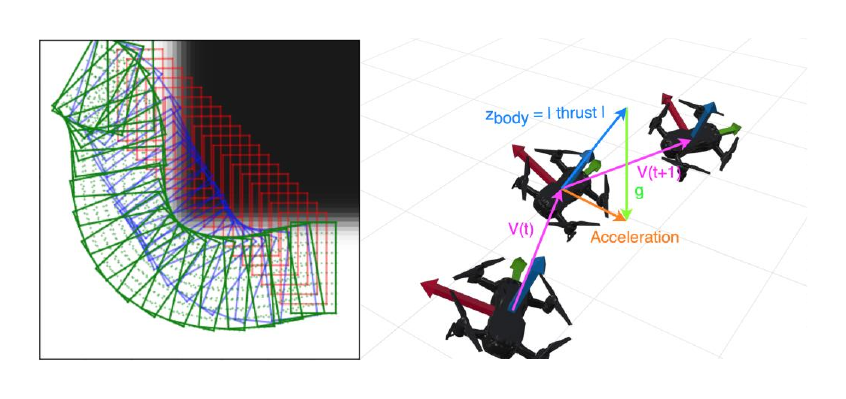}
                \label{NeRFBasedPP}
		\end{minipage}
	}
        \subfigure[Variant NeRF-based path planning]{
		\begin{minipage}[t]{0.95\linewidth}
			\centering
			\includegraphics[width=1\linewidth]{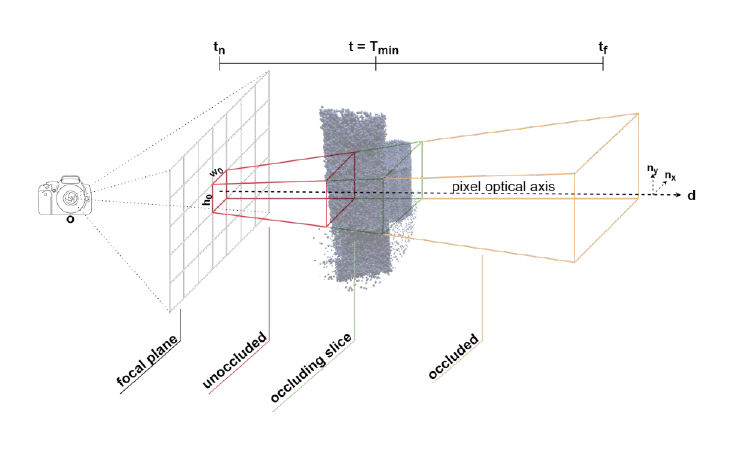}
                \label{VariantPP}
		\end{minipage}
	}
	\caption{An illustration of NeRF for Robotic Path Planning. Fig. \ref{NeRFBasedPP} \citep{adamkiewicz2022vision} shows planning a path avoiding the high-density area directly, and Fig. \ref{VariantPP} shows a variant \citep{chen2023catnips} that interprets density as the point density of a Poisson distribution.}
	\label{PathPlanning}
\end{figure}

\emph{(b) Path Planning: }
The geometry learned by the NeRF model {represents} space occupancy, enabling the direct integration of classical path-planning algorithms for navigation tasks in some {works} \citep{adamkiewicz2022vision,tong2022enforcing,byravan2023nerf2real,dai2024neural}. {In pursuit of} improved geometric interpretation {over} vanilla NeRF, some variants \citep{kurenkov2022nfomp,chen2023catnips,kwon2023renderable,shafiullah2022clip,marza2022multi} {have been explored} for navigation tasks. The basic idea of vanilla NeRF-based path planning and variants is illustrated in Fig. \ref{PathPlanning}.

NeRF-Navigation \citep{adamkiewicz2022vision} achieves safe navigation within a NeRF map by penalizing collision behavior between the point-cloud model of the robot body and the density field.
NFOMP \citep{kurenkov2022nfomp} learns an obstacle neural field for obstacle avoidance while optimizing the trajectory online. Furthermore, Lagrange multipliers are {introduced to handle} non-holonomic constraints.
\cite{tong2022enforcing} utilize future visual predictions provided by the learned NICE-SLAM model \citep{zhu2022nice} to implement robot safety control based on {visual-feedback} {Control Barrier Functions} (CBF). To realize the deployment of navigation strategies in real-world scenarios, a robot simulation system, NeRF2Real \citep{byravan2023nerf2real}, is introduced to train visual navigation and obstacle avoidance strategies leveraging NeRF as a bridge between simulation and real-world settings. 
\cite{dai2024neural} introduce {Neural Elevation Models} (NEMos) for complex terrain representation by training a NeRF and a height field jointly. The height field uses quantile regression \citep{koenker2001quantile} to extract terrain height information from images. Leveraging this height field, Dai et al. develop {an appropriate} cost function for path planning on the target terrain.

Unlike vanilla NeRF, some works extend the neural field to specially designed variant fields for path planning. 
CATNIPS \citep{chen2023catnips} reinterprets the density field as a collection of points in continuous space that follow the Poisson distribution (i.e., the Poisson Point Process), allowing for a rigorous {quantification} of the collision probability.
\cite{kwon2023renderable} introduce a visual navigation framework that includes mapping, localization, and target searching. In this work, RNR-Map is proposed to encode visual information. The features stored in the RNR-Map can be transformed into local NeRFs, and the corresponding encoder-decoder network {is} trained using an analysis-by-synthesis pipeline.

To fully exploit the semantic information, \cite{shafiullah2022clip} develop CLIP-Fields to capture both visual and semantic information. CLIP-Fields establish a mapping from spatial positions to semantic embedding vectors. Using learned CLIP-Fields, robots can achieve semantic navigation guided by language instructions. 
\cite{marza2022multi} {accomplish} multi-object navigation using {Reinforcement Learning} (RL) by learning the semantic and structural neural implicit representations online. Semantic information is used to identify object locations, while structural information is utilized to avoid obstacles.

\emph{(c) Conclusion for Navigation:} Research on NeRF-based robotic navigation has advanced toward more robust and generalizable localization and planning systems.

Early known-map methods employed pre-trained NeRFs for pose regression and APR data generation, later enhanced by Monte Carlo localization for robustness. Unknown-map approaches evolved from photometric optimization to geometric constraints \citep{SCNeRF2021}, depth priors \citep{bian2022nopenerf}, and multi-view consistency \citep{truong2023sparf}, thereby improving accuracy and stability for mobile robots.

Path planning research {has advanced from utilizing} NeRF density for collision avoidance to structured fields, including obstacle neural fields, Poisson point processes, and semantic fields, enabling more informed planning and task awareness. Recent works \citep{shafiullah2022clip,marza2022multi} integrate language and semantics for goal-directed navigation. These trends reflect a shift toward unified perception, mapping, and decision-making frameworks for adaptable robot navigation.

\subsubsection{Manipulation} \label{manipulation}
Manipulation typically involves the use of robotic arms or grippers to perform tasks, {effectively} replacing human hands. In the context of manipulation, accurately estimating the pose of the object is crucial {for determining} the final state of the robot, such as grasp poses. Between the initial and final states, a series of intermediate states can be generated by {various operational} methods.

\begin{figure}[t]
	\centering
	\subfigure[General object pose estimation]{
		\begin{minipage}[t]{0.95\linewidth}
			\centering
			\includegraphics[width=1\linewidth]{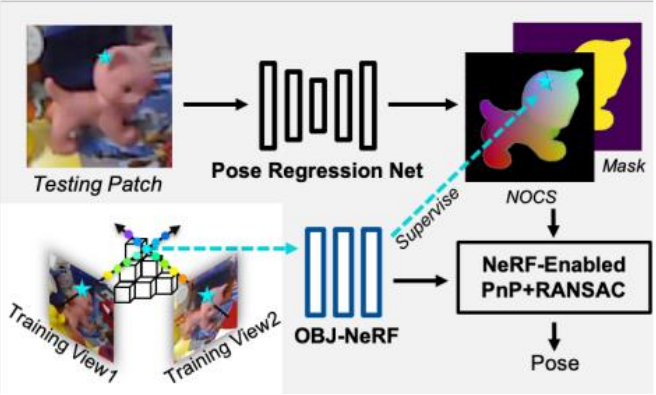}
                \label{ObjectPoseEstimation}
		\end{minipage}
	}
        \subfigure[Articulated object pose estimation]{
		\begin{minipage}[t]{0.95\linewidth}
			\centering
			\includegraphics[width=1\linewidth]{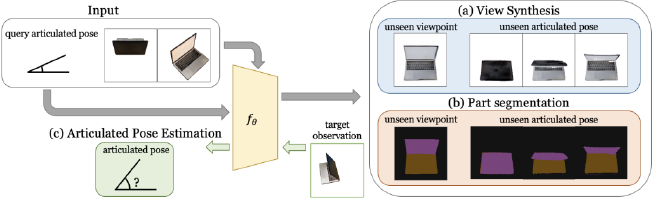}
                \label{ArticulatedOPE}
		\end{minipage}
	}
	\caption{An illustration of NeRF for Object Pose Estimation. Fig. \ref{ObjectPoseEstimation} \citep{li2023nerf} estimates the general object poses. In Fig. \ref{ArticulatedOPE} \citep{tseng2022cla}, the pose of the articulated object is estimated based on the specific connectivity properties.}
	\label{Object_Pose_Estimation}
\end{figure}

\emph{(a) Object Pose Estimation: }
Unlike robot localization, which estimates {the 6D pose of the robot} in the world, object 6D pose estimation {requires the robot to infer} the 6D pose of objects in the environment based on visual data. Moreover, we {distinguish} the pose estimation of {articulated objects} from the general object pose estimation due to the specific physical {structures}, as illustrated in Fig. \ref{Object_Pose_Estimation}.

ShAPO \citep{irshad2022shapo} learns implicit SDF geometry and texture fields from a CAD model dataset to serve as a prior database for supervising the learning of a single-shot detection and 3D prediction network.
TexPose \citep{chen2023texpose} generates a self-supervised dataset to train a 6D pose estimation network using synthetic data with perfect geometric labels and real data with realistic textures. NeRF is employed to embed realistic texture information into the model.
NeRF-Pose \citep{li2023nerf} follows the first-reconstruct-then-regress architecture and {starts} by constructing an OBJ-NeRF model, {after which object 6D poses are iteratively regressed through} a NeRF-Enabled PnP$+$RANSAC algorithm.
\cite{hu2023nerf} introduce NeRF-RPN, a universal framework for object detection that extracts features from implicit NeRF models. The entire NeRF-RPN process eliminates the need for time-consuming 3D-to-2D rendering and is applicable to various feature extraction networks and RPN models.
NeRF-Det \citep{xu2023nerf} proposes sharing geometric features between the NeRF branch and the 3D detection branch, leveraging NeRF's multi-view consistency to achieve more accurate detection results.
BundleSDF \citep{wen2023bundlesdf} constructs the neural object field {while simultaneously optimizing the pose graph online}, {enabling real-time estimation of object 6D poses and ensuring global consistency of the 3D representation}.
NeuralFeels \citep{suresh2024neuralfeels} integrates multi-modal visual and tactile dexterous hand perception, {interacts} with various objects using {proprioception-driven} techniques, and develops {an online neural field} to represent the geometry of objects. It also tracks the 6D pose of objects by refining a pose graph. In tasks involving in-hand manipulation, NeuralFeels demonstrates that tactile perception can, to some extent, resolve ambiguities present in visual perception.

Due to the specific physical properties of articulated objects, pose estimation can leverage these properties.
CLA-NeRF \citep{tseng2022cla} additionally estimates the segmentation of different articulated components. {By combining NeRF with articulated segmentation}, CLA-NeRF can forward-render images with novel articulated poses using an articulated deformation matrix and estimate the articulated pose from a given target image {through} inverse rendering.
NARF22 \citep{lewis2022narf22} learns various articulating parts and combines them based on a given configuration (i.e., articulating joint parameters). {Similarly,} NARF22 supports rendering images {with novel articulated poses} and estimating articulating configurations based on a given target image.

\begin{figure}[t]
	\centering
	\subfigure[NeRF-based operation]{
		\begin{minipage}[t]{0.95\linewidth}
			\centering
			\includegraphics[width=1\linewidth]{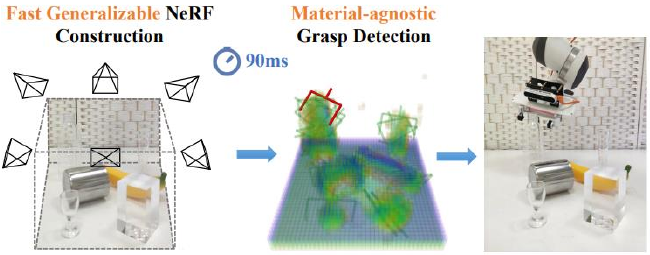}
                \label{NeRFBasedOp}
		\end{minipage}
	}
        \subfigure[Variant NeRF-based operation]{
		\begin{minipage}[t]{0.95\linewidth}
			\centering
			\includegraphics[width=1\linewidth]{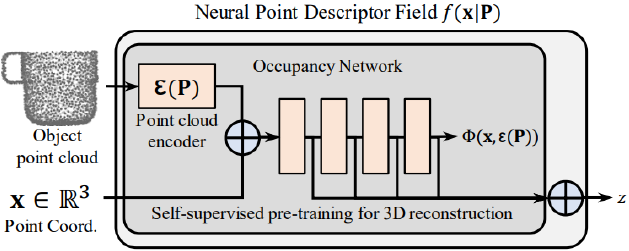}
                \label{VariantOp}
		\end{minipage}
	}
	\caption{An illustration of NeRF for Robotic Operation. subfigure \ref{NeRFBasedOp} \citep{dai2023graspnerf} illustrates a method that utilizes NeRF as a perceptual tool, subfigure \ref{VariantOp} \citep{simeonov2022neural} extends neural fields' boundaries to better serve operational tasks beyond radiance representation.}
	\label{Operation}
\end{figure}

\emph{(b) Object Operation: } \label{section-operation}
The 3D structural bias of NeRF contains richer scene information compared to 2D perception methods and can be directly applied to specific operational tasks when combined with certain operation planning methods \citep{hu2022template, chen2023perceiving, tang2023rgb, li20223d, driess2023learning, wang2022dynamical, lin2023mira, shen2022acid, ichnowski2021dex, dai2023graspnerf, kerr2022evo, zhong2023touching, higuera2023neural, driess2022reinforcement, shim2023snerl}. With continuous exploration, some concepts and methods from neural variants have extended the representation of vanilla NeRF, forming a more targeted expressions for operational tasks \citep{simeonov2022neural, simeonov2023se, chun2023local, yen2022nerf, blukis2023one, weng2023neural, khargonkar2023neuralgrasps, zhou2023nerf}, and thus achieving satisfactory performance. As illustrated in Fig. \ref{Operation}.

The most direct approach is to use NeRF to provide strong 3D scene priors for subsequent operation training.
\cite{hu2022template} learn a NeRF model of the target object without a known category to generate a large number of template images{, which are then used to train a detection network for manipulation.}
\cite{chen2023perceiving} propose continuously poking the detected object with a robotic arm to obtain complete visual perception for modeling an unknown target object. The constructed NeRF model is subsequently used to train other pose estimation networks for manipulation.
\cite{tang2023rgb} utilize the mesh representation built from a fast NeRF model to compute SDF. Based on the mesh model, a sampling-based {Model Predictive Control} (MPC) algorithm is employed to predict motion.
\cite{li20223d} train an encoder-decoder network to learn viewpoint-equivalent image states by employing time-contrastive loss and reconstruction loss. The viewpoint-equivalent image states {are then used to} train a motion prediction model{, which forecasts future states} relevant to actions. Finally, the predicted future states are integrated with MPC methods to learn visuomotor control strategies.
\cite{driess2023learning} encode the implicit representation of each object in the dynamic scene. A {Graph Neural Network} (GNN) is trained to predict the future states of the dynamic NeRF based on current encodings.
KP-NeRF \citep{wang2022dynamical} incorporates invariant relative positions between key points and query points as an additional condition to train a dynamic prediction model.
MIRA \citep{lin2023mira} employs orthographic ray casting instead of perspective ray casting to render novel views with invariant object size and appearance{, allowing for the prediction of operations by a learned action-value function.}
ACID \citep{shen2022acid} models the geometric occupancy of non-rigid objects implicitly based on images and predicts flow to represent dynamic deformations. Moreover, the correspondence between various deformation states is learned through contrastive learning. Finally, a model-based planning approach is trained to acquire a set of actions by minimizing the cost function.
\cite{blukis2023one} add a prediction head to estimate the score of sampled grasping poses in the grasping pose space. This {approach} involves predicting {feasible} grasping poses while rendering novel views of the object.

Moreover, NeRF demonstrates excellent performance in operating scenarios {where fine-grained 3D structures are crucial.}
Dex-NeRF \citep{ichnowski2021dex} leverages the volume density field of NeRF to capture globally consistent scene geometry, enabling grasp planning for transparent objects. 
GraspNeRF \citep{dai2023graspnerf} aggregates features and predicts the TSDF values. Then, a grasp detection network predicts the grasping poses of objects, including transparent and specular objects, based on the predicted TSDF values.
Evo-NeRF \citep{kerr2022evo} modifies Instant-NGP \citep{muller2022instant} to support collecting data {during NeRF model training, enabling adaptation} to continuous grasping operations. A radiance-adjusted grasp network is trained to calculate the grasp pose based on the rendered depth map of transparent objects.
NeRF-Supervision \citep{yen2022nerf} learns descriptors for thin and reflective objects from NeRF. The learned descriptors, which are useful for operation, represent correspondences between object surface points across frames.

Surprisingly, NeRF not only serves as a {tool for visual perception} but also {finds applications} in tactile perception.
\cite{zhong2023touching} train a {Generative Adversarial Network} (GAN) to generate tactile images that represent {touch interactions,} based on the images rendered by NeRF.
\cite{higuera2023neural} propose the {Neural Contact Field} (NCF) to predict the contact probability of the target object based on historical tactile perception data and the robot's end-effector position during operations.

At the same time, the strong 3D structure bias of NeRF has been shown to significantly enhance the performance of RL \citep{driess2022reinforcement}.
NeRF-RL \citep{driess2022reinforcement} treats the rendering of novel views as a proxy task, training an encoder and a NeRF decoder offline. During online RL policy learning, the latent space generated by the encoder serves as the state for action learning.
Furthermore, SNeRL \citep{shim2023snerl} enhances the supervision of the encoder not only with RGB information but also semantics. Additionally, the encoder is jointly supervised by a self-supervised teacher network.

Some extensions and techniques have been proposed in the neural fields to enhance the performance of operational tasks.
NDFs \citep{simeonov2022neural} learn a $SE(3)$-equivariant and class-equivariant neural descriptor from object point cloud models. Using few-shot imitation learning, robots can interact with previously unseen objects from the same category. Following this, the same team subsequently introduces R-NDFs \citep{simeonov2023se} and L-NDFs \citep{chun2023local}. The former extends NDFs to object rearrangement tasks, while the latter designs a more general neural descriptor for locally operable components, capturing similar operational priors across different object categories, and overcoming category boundaries.
\cite{weng2023neural} {propose} a neural grasp distance field {that estimates} the distance from a {given} pose to the nearest valid grasp pose{, with this distance being incorporated into the grasp cost}.
NeuralGrasps \citep{khargonkar2023neuralgrasps} introduces a novel implicit representation that establishes correlations between various robot grippers and even between robot grippers and human hands by learning similarity matrices.
SPARTN \citep{zhou2023nerf} introduces noise perturbations to the demonstration trajectories and generates perturbed trajectory-image pairs for offline data augmentation, thereby enhancing the success rate and robustness.
F3RM \citep{shen2023distilled} starts by acquiring robust priors through a visual language model and then applies distillation techniques to develop a feature field that integrates precise 3D geometry and semantics from the 2D foundation model. {This feature field representation allows for the extension to new open-set objects and the successful execution of specified language-guided operational tasks with only a limited number of few-shot operational demonstrations.}
\cite{kerr2024robot} propose Robot See Robot Do (RSRD), a two-phase framework for modeling objects and planning trajectories to replicate {the motion of target objects} from human demonstrations. {In the modeling phase, 4D-Differentiable Part Models (4D-DPM) are utilized,} guided by features from the pretrained DINO model \citep{caron2021emerging}. During the planning phase, RSRD selects optimal operation points and generates collision-free trajectories to {effectively} replicate {the motion of the target object}.

\emph{(c) Conclusion for Manipulation:} {Research on NeRF-based manipulation reveals a shift} from object-centric understanding and manipulation {towards the development of} integrated perception-action systems for robotic manipulation.

In {the domain of} object pose estimation, methods {have evolved} from {relying on} offline CAD priors to {enabling} real-time joint optimization using neural object fields{, while also integrating} tactile sensing to {address} visual ambiguities. {Furthermore, articulated object pose estimation has increasingly leveraged the structural connectivity of components to achieve more accurate pose inference.}

In object operation, NeRF has provided rich 3D structural priors to facilitate object modeling \citep{hu2022template}, motion planning \citep{tang2023rgb}, transparent object grasping \citep{ichnowski2021dex}, and tactile simulation \citep{zhong2023touching,higuera2023neural}. Recent advancements have introduced neural field variants that learn transferable descriptors, grasp distances, and cross-gripper correlations, enabling few-shot learning for open-set manipulation tasks. These {developments reflect a trend towards} unified neural representations that {enable} generalizable and efficient robotic manipulation across diverse scenarios.

\begin{table*}[th]
	\begin{center}
		\caption{The Evaluation Metrics Commonly Used in NeRFs Related to Robotic Tasks.}
		\label{metrics}
            \resizebox{\textwidth}{!}{
		\begin{tabularx}{\textwidth}{llp{10cm}}
                \hline
                \textbf{Tasks} &\textbf{Types} &\textbf{Metrics} \\
                \hline
                \multirow{14}{*}{Static Reconstruction} 
                    &\multirow{9}{*}{Accuracy} &Photometric Accuracy: \emph{Peak Signal-to-Noise} (\emph{PSNR} [dB])↑, \emph{Structural Similarity} (\emph{SSIM})↑ \citep{wang2004image}, \emph{Learned Perceptual Image Patch Similarity} (\emph{LPIPS})↓ \citep{zhang2018unreasonable}, Color \emph{L1}↓, Color \emph{Mean Squared Error} (\emph{MSE})↓ \\
                    
                    & &Geometry Accuracy: \emph{Chamfer Distance} (\emph{CD})↓, \emph{F-score}↑, \emph{Normal-Accuracy}↑, \emph{Normal-Consistency}↑ \citep{murez2020atlas}, Depth \emph{L1} [cm]↓ \citep{zhu2022nice} \\
                    & &Pose Accuracy: \emph{Root Mean Square Error of the Absolute Trajectory Error} (\emph{ATE RMSE} [cm])↓ \citep{sturm2012benchmark}
                    \\
                    \cmidrule(rl){2-3}
                    &\multirow{3}{*}{Completeness} &\emph{Precision} [\%]↑, \emph{Recall} [\%]↑, \emph{F-score} [\%]↑ \citep{murez2020atlas}, \emph{Completion} [cm]↓, \emph{Completion Ratio} [$<n$cm \%]↑ \citep{sucar2021imap}, \emph{Normal-Completion}↑ \citep{yu2022monosdf} 
                    \\
                    \cmidrule(rl){2-3}
                    &Efficiency &\emph{Running Time}↓, \emph{Memory Consumption}↓ \citep{sucar2021imap} 
                \\
                \hline
                \multirow{4}{*}{Dynamic Reconstruction} &\multirow{2}{*}{Difference}  &\emph{FLIP}↑ \citep{andersson2020flip}, \emph{Just-Objectionable-Difference} (\emph{JOD})↑ \citep{mantiuk2021fovvideovdp}\\
                \cmidrule(rl){2-3}
                &Consistency &\emph{(time) Optical Flow} \emph{(tOF)}↓, \emph{(time) LPIPS} \emph{(tLP)}↓ \citep{chu2018temporally}\\
                \hline
                
                \multirow{6}{*}{Segmentation} 
                    &\multirow{5}{*}{Accuracy} &\emph{Adjusted Rand Index} (\emph{ARI})↑ \citep{yu2021unsupervised} \\
                    & &\emph{mean Intersection-over-Union} (\emph{mIoU}) [\%]↑, \emph{Accuracy} [\%]↑ \citep{kobayashi2022decomposing}, \emph{mean Average Precision} (\emph{mAP})↑ \citep{tschernezki2022neural}\\ 
                    & &\emph{Panoptic Quality} (\emph{PQ})↑ \citep{kirillov2019panoptic} 
                    \\
                    \cmidrule(rl){2-3}
                    &Efficiency &\emph{Running Time}↓ \citep{cen2023segment} \\
                    
                \hline
                \multirow{4}{*}{Editing} 
                    &\multirow{3}{*}{Accuracy} &\emph{Fr\'{e}chet Inception Distance} (\emph{FID})↓ \citep{heusel2017gans}, \emph{Minimum Matching Distance} (\emph{MMD})↓, \emph{Coverage} (\emph{COV}) [\%]↑ \citep{tertikas2023partnerf}, \emph{Kernel Inception Distance} (\emph{KID})↓ \citep{binkowski2018demystifying}
                    \\
                    \cmidrule(rl){2-3}
                    &Efficiency &\emph{Editing Time}↓ \citep{liu2021editing} \\
                \hline
                
                \multirow{3}{*}{Navigation-Localization} 
                    &\multirow{3}{*}{Accuracy} &\emph{Absolute Trajectory Error} (\emph{ATE}): \emph{Rotation Error} [$\degree$]↓, \emph{Translation Error} [cm]↓, \emph{Outlier Ratio} [\%]↓ \citep{yen2021inerf} \\
                    & &\emph{Projected Ray Distance} (\emph{PRD})↓ \citep{SCNeRF2021} \\

                \hline
                \multirow{11}{*}{Navigation-Path Planning}
                    &\multirow{1}{*}{Accuracy} &\emph{Success Statistics}↑ \citep{kurenkov2022nfomp}  \\
                    \cmidrule(rl){2-3}
                    &\multirow{4}{*}{Efficiency} &\emph{Path Planning Time}↓, \emph{Path Length}↓ \citep{kurenkov2022nfomp}, \emph{Success Weighted by Path Length} (\emph{SPL})↑ \citep{anderson2018evaluation}, \emph{Progress Weighted by Path Length} (\emph{PPL})↑ \citep{wani2020multion}, \emph{Path Deviation}↓ \cite{chen2023catnips} \\
                    \cmidrule(rl){2-3}
                    &\multirow{2}{*}{Safety} &\emph{Signed Distance}, \emph{Maximum Inter-penetration Volume Per Trajectory} \citep{chen2023catnips}\\
                    \cmidrule(rl){2-3}
                    
                    &\multirow{2}{*}{Smoothness} &\emph{Maximum and Normalized Curvature}↓, \emph{Angle-over-Length} (\emph{AOL})↓ \citep{kurenkov2022nfomp} \\
                    \cmidrule(rl){2-3}
                    &Continuity &\emph{Cusps}↓ \citep{kurenkov2022nfomp} \\
                 
                \hline
                \multirow{9}{*}{Manipulation-Pose Estimation}
                   &\multirow{9}{*}{Accuracy} &\emph{Average Precision (AP)}: \emph{Rotation Error} [$\degree$]↓, \emph{Translation Error} [cm]↓, \emph{IoU}↑ \citep{irshad2022shapo} \\ 
                   & &\emph{Recall} [\%]↑ \citep{hu2023nerf}, 
                   \emph{Symmetric Average Euclidean Distance} \emph{ADD(-S)}↑ \citep{hinterstoisser2013model, tremblay2023diff}, 
                   \emph{Visible Surface Discrepancy} (\emph{VSD}) \citep{hodan2018bop, hodavn2016evaluation}, \emph{Maximum Symmetry-Aware Surface Distance} (\emph{MSSD}) \citep{drost2017introducing}, \emph{Maximum Symmetry-Aware Projection Distance} (\emph{MSPD}) \citep{li2023nerf} \\

                   & &\emph{Configuration Error}↓ \citep{lewis2022narf22} \\

                \hline
                \multirow{8}{*}{Manipulation-Object Operation}
                    &\multirow{5}{*}{Accuracy} &\emph{Success Rate} [\%]↑, \emph{Goal Reaching Error}↓ \citep{tang2023rgb}, 
                    \emph{Position Error}↓, \emph{Angle Error}↓ \citep{li20223d}, 
                    \emph{Average End Point Error} (\emph{AEPE})↓, \emph{Percentage Correct Keypoints} (\emph{PCK$@\delta$}) [<$\delta$ \%]↑ \citep{yen2022nerf}, 
                    \emph{Contact MSE}↓ \citep{higuera2023neural},
                    
                    \emph{Declutter Rate} (DR) [\%]↑ \citep{dai2023graspnerf} \\
                    \cmidrule(rl){2-3}
                    &Efficiency & \emph{Running Time}↓, \emph{Trajectory Used Ratio} [\%]↓ \citep{kerr2022evo} \\ 
                    \cmidrule(rl){2-3}
                    &Safety &\emph{Max Penetration} [cm]↓ \citep{tang2023rgb} \\
                \hline
 		\end{tabularx}
        }
	\end{center}
\end{table*}

\subsection{Metrics and Performance}
This section {presents} the evaluation metrics for NeRFs {in} robotic tasks, with Table \ref{metrics} {detailing the specific evaluation criteria. Additionally, the following subsections review the State-Of-The-Art (SOTA) advancements for each task.}

\subsubsection{Reconstruction}
The evaluation metrics for scene reconstruction {typically encompass} accuracy, completeness, and efficiency. 

With respect to accuracy metrics, there are further categorizations such as appearance, geometry, and pose. For appearance, rendering metrics typically evaluate the realism of novel views, such as \emph{Peak Signal-to-Noise Ratio} (\emph{PSNR} [dB]), \emph{Structural Similarity Index} (\emph{SSIM}) \citep{wang2004image}, and \emph{Learned Perceptual Image Patch Similarity} (\emph{LPIPS}) \citep{zhang2018unreasonable}. In addition, some metrics directly evaluate pixel differences, such as  \emph{Color L1} and \emph{Color Mean Squared Error} (\emph{MSE}). For evaluating geometric properties, 3D accuracy metrics such as \emph{Chamfer Distance} (\emph{CD}), \emph{F-score}, \emph{Normal Accuracy}, and \emph{Normal Consistency} \citep{murez2020atlas}, are commonly employed to assess discrepancies between the 3D ground truth model and the reconstructed model. Moreover, differences in 2D geometry can be assessed through depth maps, notably using \emph{Depth L1} \citep{zhu2022nice}. Pose-related metrics primarily evaluate the localization precision of SLAM methods, with widely used metrics including \emph{Absolute Trajectory Error Root Mean Squared Error} (\emph{ATE RMSE}) \citep{sturm2012benchmark}.

Completeness evaluation is usually {performed} by comparing the discrepancies between the predicted 3D models and the ground-truth models to determine whether the model {accurately encompasses all the content}. In the context of point cloud-based metrics, such as \emph{Precision}, \emph{Recall}, and \emph{F-score} \citep{murez2020atlas}, point count is commonly used. In addition, distances are evaluated using metrics such as \emph{Completion} [cm], \emph{Completion Ratio} [$<n$ cm \%] \citep{sucar2021imap}, and \emph{Normal-Completion} \citep{yu2022monosdf}.

Efficiency metrics primarily focus on computational efficiency and storage efficiency, typically assessed through \emph{Running Time} and \emph{Memory Consumption} \citep{sucar2021imap}.

In addition to the metrics mentioned above, dynamic reconstruction based on video data requires the assessment of video-related metrics, typically divided into two categories: those measuring video differences and those evaluating video consistency. The difference metrics include \emph{FLIP} \citep{andersson2020flip}, which quantifies the realism discrepancy between synthetic and real videos, and \emph{Just-Objectionable-Difference} (\emph{JOD}) \citep{mantiuk2021fovvideovdp}, which evaluates the visual differences between video frames. Consistency metrics over time involve \emph{tOF} and \emph{tLP} \citep{chu2018temporally}. \emph{tOF} compares the estimated optical flow between consecutive frames with the ground truth optical flow, while \emph{tLP} measures the difference between the rendered LPIPS and the ground truth LPIPS across consecutive frames.

GloRIE-SLAM \citep{zhang2024glorie} is the leading RGB-based technique for static scene reconstruction. Within the Replica dataset \citep{straub2019replica}, it achieves an average PSNR exceeding 30, an SSIM close to 0.95, and a rendering error of approximately 0.15 in LPIPS, while ensuring 85\% modeling completeness. Regarding tracking precision, GloRIE-SLAM attains an ATE RMSE of about 0.35. From an efficiency standpoint, GloRIE-SLAM requires 15 GB of memory at a rate of 0.2 FPS. The latest method in dynamic reconstruction, FlowIBR \citep{busching2024flowibr}, tested on the Nvidia Dynamic dataset \citep{yoon2020novel}, achieves a \emph{PSNR} over 30, an \emph{SSIM} of approximately 0.96, and a \emph{LPIPS} below 0.03, with a training duration of 1.5 hours.

\subsubsection{Segmentation}
Within the field of scene segmentation tasks, key evaluation metrics include accuracy, with components such as \emph{Adjusted Rand Index} (\emph{ARI}) \citep{yu2021unsupervised}, \emph{mean Average Precision} (\emph{mAP}) \citep{tschernezki2022neural}, \emph{mean intersection-over-union} (\emph{mIoU}) [\%], and \emph{Accuracy} [\%] \citep{kobayashi2022decomposing}. \emph{ARI} serves as a statistical metric in clustering analysis that evaluates the quality of unsupervised object segmentation. \emph{mAP} measures the precision of positive sample detection, assessing the effectiveness of target object segmentation. In 3D segmentation, \emph{mIoU} and accuracy describe the degree of overlap and correctness. Furthermore, \emph{Panoptic Quality} (\emph{PQ}) \citep{kirillov2019panoptic} is {relevant} to panoptic segmentation and {evaluates} the performance of predicted panoptic segmentation {across} all categories. {In addition, aside from precision}, the \emph{Running Time} \citep{cen2023segment} required to segment the intended objects is often included as an efficiency metric for the segmentation process.

GOV-NeSF \citep{wang2024gov}, SA3D \citep{cen2023segment}, and Panoptic lifting \citep{siddiqui2023panoptic} demonstrate excellent performance in semantic segmentation, instance segmentation, and panoptic segmentation, respectively.
GOV-NeSF, utilizing only 2D image data, achieves an mIoU of 52.2, an oAcc of 73.8, and an mAcc of 62.2 on the ScanNet dataset \citep{dai2017scannet}. 
SA3D achieves an average mIoU exceeding 88\% and an average mACC of 98\% on the NVOS dataset \citep{ren2022neural} 
and the SPIN-NeRF dataset \citep{mirzaei2023spin}, leveraging  NeRF's implicit representation, and records an mIoU exceeding 90\% with an mACC exceeding 98\% when employing TensorRF's tensor decomposition method. 
Panoptic Lifting \citep{siddiqui2023panoptic} achieves an average mIoU exceeding 65\%, a PQ of approximately 58, and a PSNR surpassing 28 on the HyperSim \citep{roberts2021hypersim}, Replica \citep{straub2019replica}, and ScanNet \citep{dai2017scannet} datasets.

\subsubsection{Editing}
Within the field of editing, the rendering metrics previously mentioned in static reconstruction are essential for evaluating the realism of modified images. In addition to these, the \emph{Fr\'{e}chet Inception Distance} (\emph{FID}) \citep{heusel2017gans} is employed to assess the quality of color and shape before and after editing. The \emph{Minimum Matching Distance} (\emph{MMD}) is used to measure the similarity between the generated and test shapes by computing the L2 Chamfer distance, and \emph{Coverage} (\emph{COV}) [\%] \citep{tertikas2023partnerf} is applied to determine the extent of shape variations in the generated forms. The \emph{Kernel Inception Distance} (\emph{KID}) \citep{binkowski2018demystifying} serves as a tool to assess the quality of generated images. For efficiency, \emph{Editing Time} \citep{liu2021editing} is used to measure the speed of editing.

DiffRF \citep{muller2023diffrf} demonstrates SOTA performance, achieving an FID of 15.95, a KID of 7.935, a COV of 58.93, and an MMD of 4.416 when evaluated on the PhotoShape Chairs dataset \citep{park2018photoshape}.

\subsubsection{Localization in Navigation}
The purpose of localization is to determine the position of the robot. To achieve this, the commonly used metric is \emph{Absolute Trajectory Error} (\emph{ATE}) \citep{yen2021inerf}, which evaluates the accuracy of the localization. Typically, ATE involves calculating \emph{Rotation Error} and \emph{Translation Error} by comparing the estimated trajectory with the ground trurh trajectory. It also includes the \emph{Outlier Ratio} [\%] to {represent the percentage of positions} exceeding a defined threshold. \emph{Projected Ray Distance} (\emph{PRD}) \citep{SCNeRF2021} measures a normalized distance by projecting points onto image planes, assessing alignment errors while excluding camera distortion effects.

PNeRFLoc \citep{zhao2024pnerfloc} demonstrates remarkable precision in indoor navigation. When tested on the Replica datasets \citep{straub2019replica}, PNeRFLoc achieves an average translation error of only 0.01 cm and a rotation error of 0.5$\degree$.

\subsubsection{Path Planning in Navigation}
The accuracy of localization has a significant impact on the precision of path planning, serving as a prerequisite for successful navigation. Additional key metrics for navigation focus on assessing \emph{Success Statistics} \citep{kurenkov2022nfomp}. In cases where the {robot's} execution phase is {disregarded}, success is determined by {the robot} obtaining a navigation path without any collisions \citep{kurenkov2022nfomp}. When execution is considered, success is characterized by the robot effectively reaching the target and transmitting an arrival notification \citep{anderson2018evaluation}.

When assessing efficiency, two key factors are considered: time and path efficiency. Time efficiency is commonly measured using the \emph{Path Planning Time} \citep{kurenkov2022nfomp}. In terms of path efficiency, it includes metrics such as \emph{Path Length} \citep{kurenkov2022nfomp}, \emph{Success Weighted by Path Length} (\emph{SPL}) \citep{anderson2018evaluation}, \emph{Progress Weighted by Path Length} (\emph{PPL}) \citep{wani2020multion}, and \emph{Path Deviation} \citep{chen2023catnips}. The path length quantifies the actual distance traveled by the robot. SPL and PPL evaluate path efficiency by comparing the ratio of ideal shortest paths to actual paths; SPL factors in success, while PPL focuses on navigation progress. While SPL and PPL are consistent for 1-ON navigation tasks, their calculation methods differ in multi-ON navigation tasks.
1-ON navigation tasks involve a single target, whereas multi-ON tasks involve a sequence of ordered targets.
For multi-ON navigation tasks,
SPL assigns success-based weights to the entire multi-target task \citep{anderson2018evaluation}, while PPL evaluates each sub-task separately and aggregates the results \citep{wani2020multion}. {Unlike SPL and PPL, path deviation measures the smallest discrepancy between the intended and the linear paths without referencing the ideal shortest path.}

In addition, safety, smoothness and jerkiness metrics are typically evaluated. Safety metrics provide a fundamental level of assurance by assessing the effectiveness of the planned route, ensuring safe execution, and minimizing collision risks. {Common metrics include} \emph{Signed Distance} and \emph{Maximum Inter-penetration Volume Per Trajectory} \citep{chen2023catnips}. {Smoothness metrics, on the other hand, serve as broader indicators,} such as \emph{Maximum and Normalized Curvature} and \emph{Angle-over-Length} (\emph{AOL}) \citep{kurenkov2022nfomp}. The former quantifies the curvature, while the latter evaluates the angle. Moreover, the continuity metric, \emph{Cusps} \citep{kurenkov2022nfomp}, measures the number of stops, turns and abrupt changes in robot direction, aiding in formulation of coherent strategies and minimizing unnecessary energy expenditure.

\cite{kwon2023renderable} demonstrate commendable performance in intricate indoor environments featuring multiple rooms, achieving an average navigation success rate of 65.7\% and an SPL greater than 40 on the NRNS dataset \citep{hahn2021no}.

\subsubsection{Pose Estimation in Manipulation}
Estimating the pose of objects serves as a critical perceptual goal during the execution of operational tasks, and its precision is measured using several metrics. \emph{Average Precision (AP)} \citep{irshad2022shapo} is the predominant metric, comprising two calculation approaches: one directly measures \emph{Rotation Error} [$\degree$] and \emph{Translation Error} [cm], while the other calculates \emph{IoU} with the ground truth. \emph{Recall} [\%] \citep{hu2023nerf} demonstrates the ability to identify the poses of all objects in a scene. \emph{ADD(-S)} \citep{hinterstoisser2013model} assesses the 6D pose error by calculating the Euclidean distance between the point-set in the estimated pose and the ground-truth. 
\emph{Visible Surface Discrepancy} (\emph{VSD}) \citep{hodan2018bop, hodavn2016evaluation} avoids potential occlusions by evaluating errors only at visible components. \emph{Maximum Symmetry-Aware Surface Distance} (\emph{MSSD}) \citep{drost2017introducing} and \emph{Maximum Symmetry-Aware Projection Distance} (\emph{MSPD}) \citep{li2023nerf} assess the estimated pose by determining the maximum distance and projection distance between the model surface points and the ground truth, respectively. Moreover, the pose of articulated objects is specifically evaluated using \emph{Configuration Error} \citep{lewis2022narf22}, considering unique connection methods.

NeuralFeels \citep{suresh2024neuralfeels} is a remarkable technique for estimating object poses, achieving accuracy on the scale of millimeters. By combining visual and tactile inputs, it achieves an average pose error of 5 mm in both simulated and real-world scenarios.

\subsubsection{Object Operation in Manipulation}
We classify the metrics associated with operations into three categories: accuracy, efficiency, and safety. Within accuracy metrics, the \emph{Success Rate} [\%] serves as the key indicator, quantifying the percentage of tasks successfully completed out of the total tasks. \emph{Goal Reaching Error} \citep{tang2023rgb} assesses the precision in reaching the target, calculated as the Euclidean distance between the target pose and the robot's final pose at the end of task execution. \emph{Position Error} and \emph{Angle Error} \citep{li20223d} determine the L2 distance for the position and orientation of the target operation. The \emph{Average End Point Error} (\emph{AEPE}) and \emph{Percentage Correct Keypoints} (\emph{PCK$@\delta$}) [<$\delta$ \%] \citep{yen2022nerf} assess the accuracy of keypoint correspondences across different views, helping to precisely identify operational points on the target object. The \emph{Contact MSE} \citep{higuera2023neural} calculates the mean squared error between the actual probability of contact and the predicted probability of contact, evaluating the precision of the prediction. Efficiency metrics comprise time efficiency, recorded as \emph{Running Time}, and execution efficiency, defined by the \emph{Trajectory Used Ratio} [\%] \citep{kerr2022evo}, which calculates the ratio of camera observing trajectory {within the entire motion trajectory, including both the observing and object-operation trajectories}.
{The} safety metric, \emph{Max Penetration} [cm] \citep{tang2023rgb}, estimates the deepest penetration distance of collision points in the object model during robot operation. 

As a method of applying field theory to robot operational tasks and achieving strong performance, F3RM \citep{shen2023distilled} has demonstrated its effectiveness in numerous object grasping and placement trials across different validation scenarios, achieving a success rate of 80\%, which is closely related to the 2D foundational model used. In language-driven tasks, F3RM attains a success rate of over 60\%.

\begin{figure*}[t]
	\vspace{-4mm}
	\centering
	\includegraphics[scale=0.552]{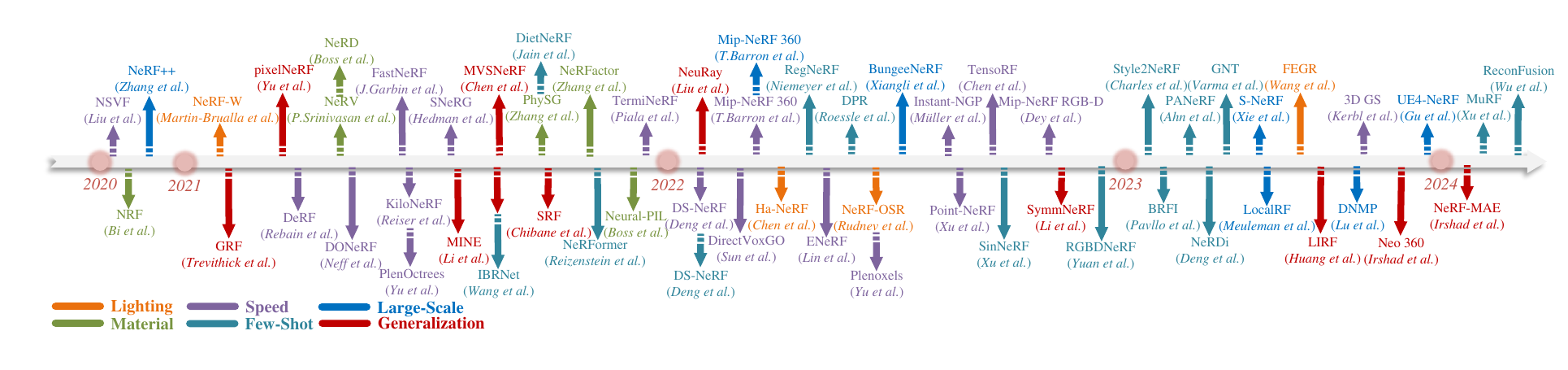}
	\vspace{-1mm}
	\caption{Chronological: Advances of NeRF related to robotic applications in  Section \ref{section4}.}
	\label{EnhancedNeRF}
\end{figure*}

\begin{figure}[t]
	\centering
		\includegraphics[width=1\linewidth]{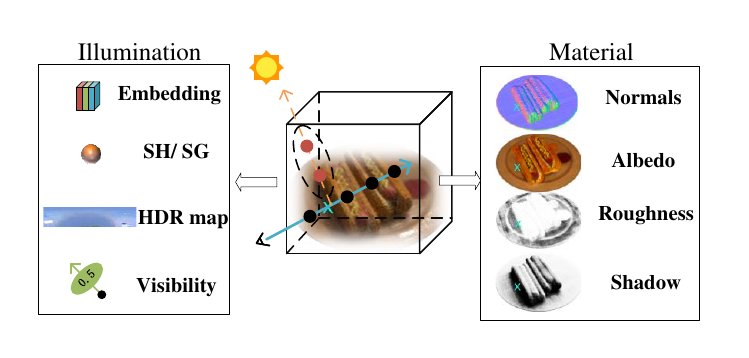}
	\caption{Realism: Quality Improvement on NeRF Representation. SH: Spherical Harmonics, SG: Spherical Gaussians, HDR: High-Dynamic Range. The images utilized in the ``HDR map" are sourced from \citep{wang2023neural}, ``Materials" from \citep{srinivasan2021nerv}. The hotdog image is sourced from the NeRF synthetic dataset, and the hotdog images below are similar.}
	\label{Reality}
\end{figure}

\section{{Advances} for {NeRFs} in Robotics} \label{section4}
Since \cite{mildenhall2020nerf} introduced NeRF, novel variants have improved realism, efficiency, and adaptability, all of which {have been successfully transferred} to the robotics domain.
The timeline of the collected works on enhancing NeRF properties related to robotic applications is presented in Fig. \ref{EnhancedNeRF}.

\subsection{Realism}
Realism is a crucial attribute of NeRF-based models. Vanilla NeRF interprets the imaging process as an integration of spatial particle radiance, avoiding the calculation of complex ray propagation and reflection. However, some flexibility is sacrificed, particularly when handling scenes with varying environmental lighting and materials, as shown in Fig. \ref{Reality}.

\subsubsection{Lighting}
In the editing section \ref{section-scene_editing}, these methods \citep{xu2022deforming, peng2022cagenerf, yuan2022nerf, yang2022neumesh} encounter challenges in handling lighting and shadows, which significantly impact the realism of edited scenes. This {highlights the importance of accurately representing lighting effects for realistic rendering}.

To enhance the capability of lighting representation,
NeRF-W \citep{martin2021nerf} introduces lighting embedding as an additional learnable condition to model the illumination. 
Ha-NeRF \citep{chen2022hallucinated} {further} trains a CNN encoding network to regress {the latent appearance vector for each image, which is then used as input to the NeRF model.} This approach ensures consistency in lighting while improving generalization to new scenes.
NeRF-OSR \citep{rudnev2022nerf} learns Spherical Harmonics (SH) coefficients to represent illumination from a set of unstructured images of outdoor scenes. {Additionally}, NeRF-OSR employs {separate networks for shadow and albedo, which} learn environmental shadows and object albedo, respectively.
For urban scenes, FEGR \citep{wang2023neural} learns the {Neural Intrinsic Field} (NIF) to model geometry, color, and material properties{, while} a {High Dynamic Range} (HDR) sky dome is learned for lighting. During rendering, FEGR \citep{wang2023neural} introduces a hybrid rendering, combining primary ray rendering based on the neural implicit model and secondary ray rendering based on an explicit mesh model derived from NeRF. The secondary ray rendering captures better lighting effects, such as highlights and shadows.

\subsubsection{Material}
The material properties, inherent to the object itself, typically encompass the reflective characteristics of surfaces within a scene, including diffuse and specular reflection. These properties determine {how light interacts with the surface, influencing the generation of reflections and shadows}. 

\cite{bi2020neural} extend NeRF to {Neural Reflectance Fields} (NRF), where the model not only learns the radiance and volume density for each ray-sampled point but also captures reflective properties, including diffuse albedo and specular roughness, which are typically represented by {the} {Bidirectional Reflectance Distribution Function} (BRDF).
NeRV \citep{srinivasan2021nerv} not only models a neural reflectance field to capture reflective properties but also learns a neural visibility field to regress the visibility of light sources at the sampled points. Visibility quantifies the propagation of light rays. Moreover, {directly inferring} the visibility field avoids the {computationally} expensive process of {integrating volumetric density} between light sources and sampled points.
Similarly, \cite{boss2021neural} utilize illumination embedding to represent lighting and propose a Pre-Integrated Light (PIL) network to decode lighting embeddings. This approach directly regresses lighting based on reflection properties at each point, replacing the integration process with a querying process.
PhySG \citep{zhang2021physg} uses {Signed Distance Functions} (SDF) to represent environmental geometry, {Spherical Gaussians} (SGs) for environmental illumination, and BRDF for object material. All parameters are jointly optimized based on photometric losses.
Similarly, NeRD \citep{boss2021nerd} models an explicit decomposition model, synchronously optimizing the shape, reflectance parameters represented by {Spatially Varying} BRDF (SVBRDF) and illumination represented by spherical Gaussians.
For unknown lighting conditions, NeRFactor \citep{zhang2021nerfactor} pre-trains additional prediction networks to reduce noise in normals and light visibility, typically calculated from density. NeRFactor \citep{zhang2021nerfactor} models illumination using an HDR light probe image and learns the reflection properties at surface points, including BRDF that absorbs reflection priors from real datasets and albedo for shadows.

\subsubsection{Conclusion for Realism}
The realism in NeRF-based models has progressed along two primary aspects: lighting and material modeling.

{In terms of lighting}, research has evolved from {the use of} global learnable embeddings to {the representation of} complex illumination {through} spherical harmonics or Gaussians, and further to hybrid rendering {that combines} implicit fields with explicit mesh-based secondary rays. These advancements significantly {enhance} the handling of {dynamic lighting conditions and shadows}.

{In the area of} material modeling, early {approaches primarily} focused on learning BRDF parameters \citep{bi2020neural}, {before expanding to include} reflectance fields \citep{srinivasan2021nerv}, light visibility fields \citep{boss2021neural}, and joint optimization of geometry, reflectance, and illumination \citep{zhang2021nerfactor}. {More recent techniques have incorporated} real-world priors and decomposed neural fields {to achieve enhanced} photorealism. {These advancements reflect a broader trend toward physically-informed and generalizable representations, which are crucial for realistic robotic perception, comprehensive scene understanding, and improved interactions within complex environments.}

\subsection{Efficiency}
In this survey, efforts to improve efficiency are categorized into two key aspects: \emph{speed} and \emph{few-shot}.
The former {focuses on enhancing} run-time efficiency, while the latter {aims at improving} data utilization efficiency.
\begin{figure}[t]
	\centering
		\includegraphics[width=0.9\linewidth]{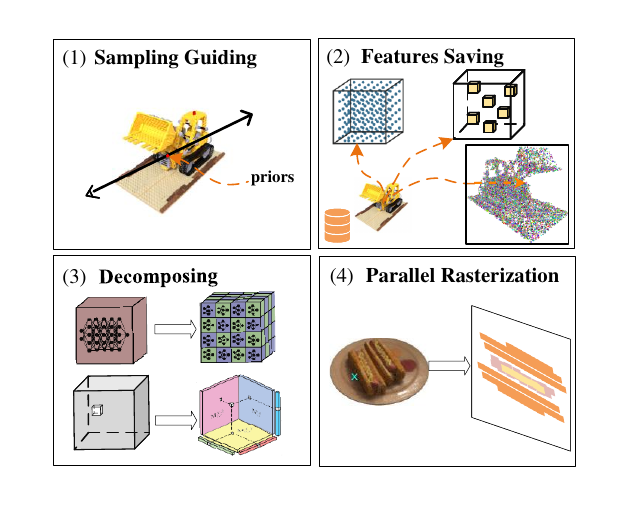}
	\caption{Speed: Speed Improvement on NeRF Representation. Some pictures of subfigure $(2)$ are extracted from \citep{yu2021plenoctrees} and \citep{xu2022point}, while some pictures of subfigure $(3)$ are taken from \citep{reiser2021kilonerf} and \citep{chen2022tensorf}.} 
	\label{Speed}
\end{figure}

\subsubsection{Speed}
The time-consuming multipoint querying process{, reliant on the MLP network, is a key factor limiting} the speed of vanilla NeRF. As shown in Fig. \ref{Speed}, various acceleration strategies are employed from different perspectives to optimize or replace the time-consuming querying process.

NeRF utilizes a coarse-to-fine sampling strategy, but the sampling process remains a bottleneck for efficiency. To address this, some methods \citep{barron2022mip, neff2021donerf, piala2021terminerf} introduce an additional sampling network to guide the sampling process. Other approaches \citep{dey2022mip, deng2022depth, neff2021donerf, lin2022efficient} leverage depth as a geometric prior to guide ray sampling on the surface. ENeRF \citep{lin2022efficient} {further enhances efficiency by utilizing the explicit geometry from Multiple View Geometry (MVS).}

{Although NeRF's implicit representation is storage-efficient, enhancing speed often comes at the cost of some storage.} To improve efficiency, attribute parameters are typically pre-stored in explicit structures, or tools based on explicit representations, such as CNNs, are employed.
\cite{sun2022direct} combine an explicit voxel grid representation with efficient interpolation to model scenes. Their approach involves interpolating first and then activating to compute the value of $\alpha$ in formula (\ref{3}), {which, as demonstrated by experiments, accelerates the acquisition of sharp surfaces. Additionally}, a coarse-to-fine {strategy} is {employed to bypass invalid regions and optimize computation in valid areas}.
Baking-NeRF \citep{hedman2021baking} stores view-independent diffuse colors compactly in a Sparse Neural Radiance Grid (SNeRG) for direct querying. 
NSVF \citep{liu2020neural} learns implicit voxel-bounded radiance fields, utilizing an explicit sparse voxel octree structure. 
\cite{yu2021plenoctrees} tabulate the density and SH coefficients of their NeRF-SH model, storing them in each leaf of a PlenOctree for direct querying. 
Subsequently, Plenoxels \citep{fridovich2022plenoxels} learns occupancy and SH coefficients for each vertex in sparse voxel grids explicitly, without relying on neural components.
Instant-NGP \citep{muller2022instant} constructs a hash table {with multiple} resolution layers, {enabling rapid feature querying}. 
Point-NeRF \citep{xu2022point} utilizes pre-trained CNNs to infer and generate a neural point cloud containing scene features. This neural radiance field, based on the neural point cloud, achieves {impressive results with minimal fine-tuning for specific scenes.}

{Another approach to improving efficiency is through decomposition, where the global, complex, or high-dimensional representation is broken down into local, simpler, or lower-dimensional components.}
DeRF \citep{rebain2021derf} and KiloNeRF \citep{reiser2021kilonerf} utilize multiple smaller neural networks to replace a single large network, with each network representing a small part of the scene. 
FastNeRF \citep{garbin2021fastnerf} computes the inner product of the decomposed position and direction functions to obtain the final RGB values. 
TensoRF \citep{chen2022tensorf} employs tensor decomposition to break down the 4D scene tensor representation into the element-wise multiplication of several compact low-rank tensor components. 

Lastly, substantial efficiency gains are achieved through advancements in acceleration techniques and rendering methods. \cite{kerbl20233d} use a set of 3D Gaussians as the core units for scene representation, leading to more realistic rendering outcomes. Sorting techniques and GPU acceleration are employed to {balance realism with enhanced speed}. Additionally, a tile-based rasterizer replaces the time-consuming ray marching rendering process.

\begin{figure}[t]
	\centering
		\includegraphics[width=0.85\linewidth]{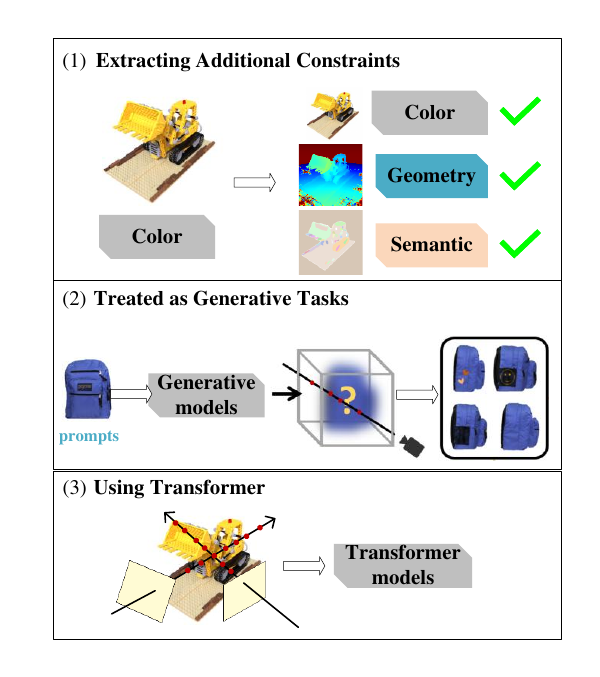}
	\caption{Few-Shot: Image Utilization Efficiency Improvement on NeRF Representation. First category of approaches extracts additional constraints (such as depth or semantics), as shown in subfigure $(1)$, part images of which are taken from \citep{xu2022sinnerf}. Second category transforms the task into a generative one, utilizing the limited views provided as prompts to guide the generation process, as shown in subfigure $(2)$, part images of which are taken from \citep{deng2023nerdi}. Last category uses Transformer models to correlate and aggregate features as shown in subfigure $(3)$.}
	\label{Few-Shot}
\end{figure}
\subsubsection{Few-Shot}
The challenge of rendering a novel view {with few shots stems from the limited information available}. In scenarios with only a few observations, the vanilla NeRF {either fails to} converge or overfits to a smooth solution \citep{jain2021putting}. To achieve an optimal model in a few-shot setting, additional constraints must be imposed, facilitating the extraction of more valuable prior knowledge, as illustrated in Fig.\ref{Few-Shot}.

When leveraging geometry,
RegNeRF \citep{niemeyer2022regnerf} applies both appearance and geometric regularization to patches rendered from unseen viewpoints. 
DS-NeRF \citep{deng2022depth} and \cite{roessle2022dense} use depth values generated during the Structure-from-Motion (SfM) process as guidance. Furthermore, Roessle et al. pretrain a depth completion network to densify the depth ground truth. 
When leveraging semantics,
DietNeRF \citep{jain2021putting} utilizes semantic priors provided from a pre-trained CLIP model to guide the learning process of the NeRF model. These semantic priors encourage high semantic similarity between different viewpoints of the same object. 
SinNeRF \citep{xu2022sinnerf} {combines} geometry and semantic information to generate a large amount of {pseudolabeled} data from a single reference frame for training. 
PANeRF \citep{ahn2022panerf} warps reference frames to 
create pseudoviews and integrates the CLIP model to ensure semantic consistency on both local and global scales. 
\cite{yuan2022neural} generate pseudo-training data from a coarse mesh constructed from sparse RGB-D observations.

Some approaches treat few-shot modeling as a generative task to achieve the desired results.
NeRDi \citep{deng2023nerdi} {leverages the generative power of} a language-guided diffusion model to transform the few-shot NeRF learning task into a generative process.
ReconFusion \citep{wu2024reconfusion} pre-trains a diffusion model to provide pseudo ground-truth supervision for unseen views during few-shot NeRF reconstruction.
Style2NeRF \citep{charles2022style2nerf} and \cite{pavllo2023shape} reframe the task of generating novel views from a single image as a 3D perception-based GAN inversion task.

Additionally, the Transformer's ability \citep{vaswani2017attention} to correlate and aggregate features significantly enhances the efficient utilization of image features from few-shot views. For instance, NerFormer \citep{reizenstein2021common} leverages transformers to aggregate image features from {the provided views along with features from sampled points along a ray}. Similarly, IBRNet \citep{wang2021ibrnet} proposes a ray transformer that aggregates the density features of sampled points along a ray. GNT \citep{varma2022attention} not only aggregates features via a transformer but also learns to directly render pixel colors using the ray transformer.
MuRF \citep{xu2024murf} employs a multi-view Transformer to extract image features from few-shot views, constructs a target-view-aligned volume representation, and generates a radiance field through a CNN applied to this volume.

\subsubsection{Conclusion for Efficiency}
Research {aimed at enhancing} the efficiency of NeRF-based models has {shifted} from optimizing network architectures to rethinking scene representations for {faster performance}, and from {relying} on dense observations to leveraging generative learning for few-shot {scenarios}.

Early research focused on accelerating the querying process through sampling strategies \citep{barron2022mip}, voxel grids \citep{fridovich2022plenoxels}, and multi-resolution hash encodings \citep{muller2022instant}, while {subsequent} methods introduced sparse neural fields and compact decompositions{, enhancing both rendering speed and memory efficiency}.

Simultaneously, research on few-shot NeRF has evolved from leveraging geometric and semantic priors to stabilize learning with limited observations, to reframing the task as a generative problem using diffusion models and GAN-based approaches. {Collectively}, these trends {highlight an increasing focus on} balancing performance, data efficiency, and computational practicality, enabling the deployment of NeRF-based perception in real-time, resource-constrained robotic applications.

\subsection{Adaptability}
The {suboptimal} performance of vanilla NeRF in large-scale and unseen scenes {limits} its adaptability {in robotic deployments}. {Enhancing its performance in these scenarios would significantly broaden its applicability across diverse environmental contexts}.

\begin{figure}[t]
	\centering
		\includegraphics[width=1\linewidth]{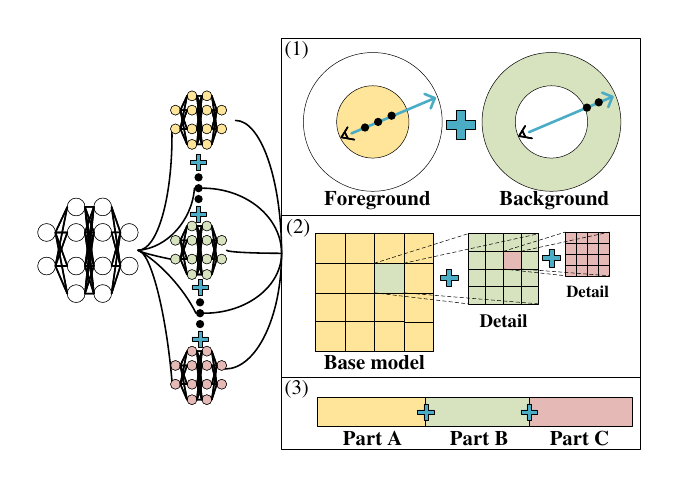}
	\caption{Large-scale: Adaptability of NeRF to Large-Scale Scenes. Multiple models are employed to model different parts of a large-scale scene according to different rules.}
	\label{Large-scale}
\end{figure}
\subsubsection{Large-Scale}
In large-scale scenes, only a {limited number of} viewpoints {capture} small areas of co-visible observations{, and details of distant objects are often insufficiently captured in unbounded environments}. {To address this, different scene regions} are modeled separately according to {distinct} rules, as {illustrated} in Fig. \ref{Large-scale}. {This approach prevents a single model from needing to reconcile diverse scene parts, ensuring smoother results}.

{To better parameterize distant regions}, 
NeRF++ \citep{zhang2020nerf++} introduces an inverted sphere parameterization and {constructs a separate NeRF model} for distant elements.
Similarly, Mip-NeRF 360 \citep{barron2022mip} {refines the cone sampling boundaries} of Mip-NeRF \citep{barron2021mip}, {consolidating} Gaussian samplings outside the {predefined spherical domain} into the sphere. 
{Building} on Mip-NeRF 360 and Mip-NeRF, S-NeRF \citep{xie2023s} {further integrates sparse LiDAR signals} and {generates} a confidence map to {guide} the learning process. 
Differently, Mega-NeRF \citep{Turki_2022_CVPR} {shifts from a unit sphere to an ellipsoidal domain, offering a more efficient bounding region.}
{To overcome the limitations of an individual neural network's capacity,} BungeeNeRF \citep{xiangli2022bungeenerf} introduces a progressive neural network framework, where additional residual blocks are {progressively incorporated} as more scene details are captured. 
LocalRF \citep{meuleman2023localrf} introduces a time-sliding window strategy for local NeRFs modeling. As the camera moves, {new content is continuously captured and modeled by adding local NeRFs}. Connections between adjacent NeRFs are established based on their co-visible regions. 
Similarly, UE4-NeRF \citep{gu2023ue4} divides large scenes into different blocks, {with a NeRF model constructed for each block. It also} integrates the Unreal Engine 4 (UE4) mesh rasterization pipeline, enabling real-time rendering. 
\cite{lu2023urban} propose a novel neural mesh representation element called {the Deformable Neural Mesh Primitive} (DNMP). By modeling the radiance field based on DNMP, {the approach facilitates scaling to large scenes with efficient rasterization-based rendering, while ensuring high-quality results}. 

\subsubsection{Generalization}
\begin{figure}[t]
	\centering
		\includegraphics[width=0.8\linewidth]{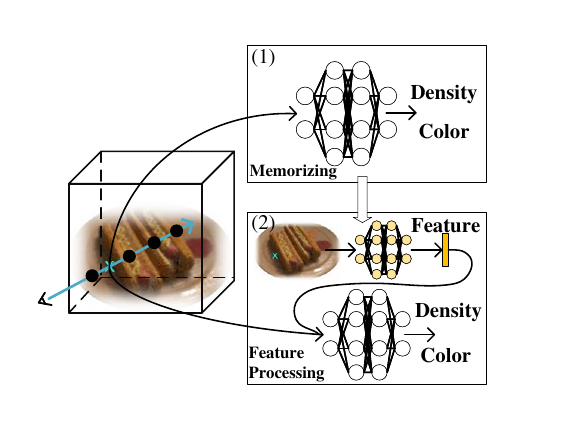}
	\caption{Generalization: Adaptability of NeRF to Novel Scenes. The core of generalization is training a network to learn a general capability for processing scene features, replacing the learning of memorizing scenes.}
	\label{Generalization}
\end{figure}

Vanilla NeRF implicitly memorizes a scene, {which leads to overfitting to that specific scene and poor performance in unknown scenarios}. To achieve {better} generalization, the network needs to {learn how to handle scene features in a more flexible way}, rather than {relying} solely on memorization. {This concept is illustrated} in Fig. \ref{Generalization}.

PixelNeRF \citep{yu2021pixelnerf} and GRF \citep{trevithick2021grf} {incorporate} extracted pixel-level features as additional input{, enabling the network to learn general feature processing capabilities instead of memorizing specific scenes}.
IBRNet \citep{wang2021ibrnet} further employs a ray transformer to correlate features of spatial points along the same ray, improving geometric accuracy.
MINE \citep{li2021mine} trains a general encoder-decoder network, decoding the encoded features of the source images plane by plane and regressing the color and volume density based on the multi-plane image structure of the camera's frustum. 
SRF \citep{chibane2021stereo}, inspired by classical Multiview Stereo (MVS) methods, trains a radiance field decoder to infer color and geometry based on extracted features with high inter-image similarity.
\cite{huang2023local} propose a local implicit ray function (LIRF) based on cone sampling, {which accounts for view visibility}. This method interpolates local region features from the queried image, corresponding to the eight vertices of the cone in which the sampled point lies.
SymmNeRF \citep{li2022symmnerf} incorporates a hypernetwork that learns to regress the NeRF weight parameters from the global image features. The NeRF model then utilizes both the feature of the sampled position and its corresponding symmetrical counterpart to refine the representation details.
MVSNeRF \citep{chen2021mvsnerf} employs a generalized MVS-like framework. First, it reconstructs a neural encoding volume using standard MVS techniques. Then, MVSNeRF trains a rendering network to infer color and density based on features extracted from the encoding volume.
NeuRay \citep{liu2022neural} predicts the visibility of features extracted using MVS-like methods, quantifying occlusion between different views. This enables a more efficient use of the extracted features.
NeO 360 \citep{irshad2023neo} extends the tri-planar representation to generate $360^{\circ}$ novel views of outdoor driving scenes from sparse RGB images, while also ensuring generalization. {Additionally, it introduces a panoramic driving dataset for $360^{\circ}$ scenes.}
NeRF-MAE \citep{irshad2024nerf} enhances NeRF's self-supervised learning by training a pyramid-structured transformer auto-encoder to encode NeRF's feature grids for the masked grid completion task. The encoded embeddings are then decoded by task-specific decoders, enabling adaptation to various downstream 3D tasks.

\subsubsection{Conclusion for Adaptability}
Recent advancements in NeRF adaptability research highlight a shift towards scalable and generalizable models designed for deployment in dynamic real-world environments.

To overcome the limitations of early NeRF models in large-scale scenes, subsequent research introduced strategies such as block division \citep{xiangli2022bungeenerf}, progressive networks \citep{xie2023s}, and local sliding windows \citep{meuleman2023localrf}, enabling scene coverage that exceeds the capacity of individual models.

{Simultaneously, research on generalization has investigated several approaches, including the integration of auxiliary features to reduce overfitting}, as well as the learning of global scene representations through tri-planar mappings and hypernetworks. Recent methods have expanded to include self-supervised learning and masked completion techniques to further enhance generalization. Collectively, these developments aim to provide NeRF models with the flexibility and robustness needed for robotic applications in diverse and previously unseen environments.

\section{Discussion} \label{section5}
In this section, we {outline several key} challenges and {discuss promising} research directions {inspired by} these {issues within the community.}

\subsection{Map Fusion}
{Robots typically move, and their surrounding environment changes as their location and time progress. Consequently, the robot needs to continuously update its map to reflect these changes. Moreover, in} large-scale environments, it is often more efficient to {deploy} multiple robots to {collaboratively build a 3D map}. Therefore, map fusion {becomes a critical challenge for applying NeRFs to robotic 3D mapping}. 

Here, we define two types of fusion: temporal fusion and spatial fusion. Temporal fusion addresses changes occurring within the same scene over time, including natural environmental variations and changes caused by robot interactions, such as illumination {shifts} at different times {or object displacement due to robot activity}. Spatial fusion involves merging NeRF scene maps in large-scale environments, {enabling a single} robot to adapt flexible spatial ranges or {facilitating the combination of multiple NeRF maps generated by multiple robots}. 

Temporal fusion focuses on accurately {localizing scene} changes, such as those {addressed in dynamic scene modeling} \citep{yuan2021star,ost2021neural,gao2021dynamic,li2021neural,xian2021space,du2021neural,gafni2021dynamic,thies2016face2face,park2021nerfies}, {by} updating only {the modified regions and integrating current observations with historical maps. Since the content of a scene typically remains relatively stable over short time intervals, repeatedly performing global reconstruction is inefficient and unnecessary.}

Spatial fusion focuses on the {accurate alignment of two or more scene maps. Achieving precise and seamless registration may involve combinations of 2D-2D, 2D-3D, or 3D-3D correspondences, and in some cases, temporal alignment is also required.} Furthermore, interruptions in a robot’s exploration, such as those caused by system failures, may prevent it from resuming its previous state upon returning to the environment. In such scenarios, multi-scale fusion of historical information becomes essential to ensure consistent and robust mapping.
{In the context of map fusion, we also consider the challenge} of information sharing among multiple robots {during} exploration of unfamiliar environments. {Deploying} multiple robots is one of the most {straightforward and effective strategies for accelerating the exploration and mapping of novel environments}. 
{Several recent studies have explored solutions to this challenge}. \cite{zhao2024distributed} {propose a distributed learning framework that enables multiple robots to share the weights of their individually trained NeRFs for collaborative environment mapping}. \cite{yu2025hammer} propose HAMMER, which incorporates a robot alignment module to estimate the relative poses between aligned and unaligned robots, {facilitating} multi-robot data alignment for joint map optimization. \cite{zhao2025ramen} further address issues related to communication loss in multi-robot systems by proposing an asynchronous multi-agent neural implicit mapping {approach that promotes consensus mapping under uncertainty}. Additionally, \cite{patel2023dronerf} present DroNeRF, which {optimizes drone viewpoints through iterative planning to capture more informative observations and improve geometric detail acquisition}.
However, the challenge of effectively fusing separately reconstructed maps generated by different robots remains unresolved. A well-designed spatiotemporal NeRF map fusion method could provide accurate and semantically enriched priors, thereby enabling more robust and informed robot decision-making in complex environments.

\subsection{Robot Relocalization for Large-Scale Scenes}
Once a complete NeRF map is {constructed,} the robot {can localize itself by estimating its current pose using both the map and incoming observations, similar to the approach proposed in iNeRF \citep{yen2021inerf}}. However, this optimization-based method may {fail to converge at the scene level due to vanishing gradients.} To address this challenge, we present two possible research directions.

{First, we posit that a coarse-to-fine multi-scale structure can be effective for robust pose estimation. Analogous to human intuition, an approximate pose can first be inferred by locating a visually similar region at a coarser scale, which is then refined through fine-grained optimization at higher resolutions.
Second, we propose the use of auxiliary features as markers embedded in both the NeRF map and robot observations to guide the optimization process.}
Recently, \cite{avraham2022nerfels} {introduced Nerfels, which are 3D primitive patches anchored at keypoints in 3D space}. Each Nerfel is {associated with} a renderable implicit embedding {that functions as a marker}, enabling end-to-end optimization {for camera pose estimation}.

Furthermore, {effective relocalization should go beyond relying solely on appearance features and must be robust to scene changes by incorporating multi-modal information}, such as semantics and {data from} multiple sensors. For example,
\cite{partha2024robust} enhance the NeRF-based neural city map \citep{partha2023neural} by {integrating} depth and semantic features{, enabling the system} to match the current observations {against the enhanced neural map under varying visual and environmental conditions}.

\subsection{More Generalization Ability across Various Scenarios}
We have introduced several generalization approaches \citep{yu2021pixelnerf,wang2021ibrnet,liu2022neural,chen2021mvsnerf,li2021mine} that render novel views conditioned on features extracted using neural network encoders. However, {the generalization achieved by these methods is typically limited to scenes that closely resemble the training data, primarily due to constraints in the representational capacity of the encoding networks}. A {significant} research gap {remains in achieving robust} generalization {across diverse real-world scenarios, which often involve a wide range of properties}, such as different mechanical properties (e.g., rigid bodies, {deformable objects}, fluids), geometry structures (e.g., {square-shaped and cylindrical chairs}) and complex illuminations (e.g., {daytime versus nighttime environments}).

{We propose two promising directions to enhance generalization capabilities based on feature processing.}
{First, leveraging or fine-tuning large pre-trained feature models across diverse scenarios presents a favorable approach compared to training small feature networks from scratch.} Advances in network architecture have enabled the training of larger models with increased depth and width on vast datasets, allowing these models to capture high-level features that generalize well to complex, real-world environments.
{Second, complementing large models, the integration of precise physical mechanisms into smaller, resource-efficient networks offers an alternative avenue.} As exemplified by \cite{xie2023physgaussian}, incorporating well-understood physical priors can guide networks to extract meaningful features from distinct scene components and fuse diverse characteristics into NeRF representations. This physics-informed approach facilitates improved generalization in practical scenarios while maintaining computational efficiency.

\subsection{Rendering to Real}
The ability of NeRF to realistically {reconstruct scenes holds significant promise for generating training data and simulation environments for robotic learning}. NeRF2Real \citep{byravan2023nerf2real}, RialTo \citep{torne2024reconciling}, and RL-GSBridge \citep{wu2024rl} have {begun exploring this potential}. {Acquiring training data is particularly critical for scenarios that are difficult to capture in the real world}, such as abnormal {driving behavior in autonomous vehicles or extreme environments like deserts, deep oceans, or outer space, where human operation is challenging}. {Robots} inadequately trained on such corner cases {are prone to failure when deployed in unfamiliar or safety-critical situations, potentially causing severe consequences}. {Moreover, real-world training is costly and time-consuming, and traditional environment modeling often requires experienced professionals to create highly realistic simulations}, which {can be} inefficient. Therefore, {employing} NeRF-based {techniques to synthesize training data and enabling successful sim-to-real transfer is highly valuable}. {Nonetheless}, this approach faces notable challenges{, including limited} physical realism {in rendered scenes} and the scarcity of learnable data {representing rare or extreme conditions}.

The lack of physical realism {manifests as inaccurate} rendering of {fine-grained} variations in lighting and shadows {observed in real-world scenes}. {Meanwhile,} the scarcity of learnable data {for} corner cases and extreme environments {complicates the prediction of} dynamic changes {arising from complex physical} interactions. To overcome these challenges, one direction is to leverage extensive expertise from computer graphics and utilize virtual engine tools, which have the potential to enable a qualitative leap in simulation fidelity and robustness. 
In addition, NeRF-based approaches for few-shot scenarios
\citep{jain2021putting,xu2022sinnerf,niemeyer2022regnerf,humasknerf,yuan2022neural} that leverage more constraints have demonstrated promising results in addressing the challenges of corner cases, highlighting a valuable direction for future research.

We also anticipate {increased} exploration {of the integration} of generative models, such as GANs and diffusion models, which {have demonstrated strong} capabilities {in synthesizing high-quality} data under {conditional} guidance. {Moreover, large pretrained generative models exhibit impressive capabilities, including the ability to generate images or videos directly from textual prompts} \citep{singer2022make, betker2023improving, videoworldsimulators2024}. The {prospect} of combining NeRF with the {generative power of such models} to directly {synthesize controllable 3D environments is particularly compelling and opens up exciting opportunities for future research}.

\subsection{Robot Interaction with Multi-Modal Sensors}
In realistic environments, robots are {exposed to rich} multi-modal information, including color, geometry, semantics, {sound, and even smell and taste}. {These modalities are perceived through various sensory channels such as vision, hearing, touch, and olfaction}. NeRF and its extensions {primarily} focus on visual perception, {capturing radiance and geometry to represent scenes, with some recent efforts also incorporating semantic understanding} \citep{zhi2021place, vora2021nesf, cciccek20163d, zhi2021ilabel, blomqvist2022baking, liu2022unsupervised, zhu2023sni}. 

{In addition to visual perception,} preliminary explorations {have been conducted} into auditory and tactile modalities. 
For {example}, AD-NeRF \citep{guo2021ad} encodes audio signals from videos to {synthesize talking head animations, while} 
the Neural Acoustic Field (NAF) \citep{luo2022learning} implicitly models spatial sound propagation. NeRAF \citep{brunetto2024neraf} jointly reconstructs neural radiance and acoustic fields, enabling the rendering novel audio-visual data.
Furthermore, works such as \cite{zhong2023touching} and \cite{higuera2023neural} render tactile images to represent the state of a gripper during object contact.

In addition to {conventional visual, auditory}, and tactile modalities, spatial perception {using} LiDAR signals {plays a critical role in robotic sensing} \citep{huang2023neural,deng2023nerf,zhang2024nerf,tao2024lidar,sun2024lidarf}.
{To improve cost-efficiency}, low-resolution ranging sensors, such as infrared and ultrasonic devices, are {often adopted as alternatives to} expensive LiDAR or depth cameras for depth perception \citep{schmid2024virus}. Infrared sensing, in particular, is widely employed for robot perception and scene reconstruction in visually degraded environments \citep{ye2024thermal,xu2024leveraging,lin2024thermalnerf}.

{The findings suggest that multi-modal research grounded in NeRFs is a promising direction for further exploration}.
This potential can be qualitatively understood: scenes that pose challenges to visual perception alone may become more tractable when augmented with other sensory modalities. 
For instance, visually guided tasks such as pouring water into a container can suffer from significant errors due to occlusions by the robotic arm or the use of opaque materials. 
In contrast, {auditory cues, such as changes in sound pitch corresponding to varying water levels, can provide reliable supplementary information}. 
{As such}, integrating multi-modal scene perception and understanding is an emerging and important research direction \citep{li2022see}. {The goal is for different sensory modalities to enhance, complement, and cross-validate one another, ultimately enabling robots to operate more robustly in complex and dynamic real-world environments}.

Fortunately, several publicly available multi-modal robot-related datasets support exploration in this direction. 
For example, \cite{clarke2023realimpact} introduce the REALIMPACT dataset, which contains 150,000 recordings of impact sound fields from 50 common real-world objects, annotated with impact locations, microphone positions, contact force curves, material types, and RGB-D images. 
\cite{fang2023rh20t} present the RH20T dataset, {which comprises} over 110,000 real-world robot manipulation {instances, including multi-sensory} data such as vision, force, audio, and action information. Additionally, \cite{liu2024maniwav} propose the ManiWAV dataset, collected using an "ear-in-hand" device, which captures human-demonstrated manipulation data with synchronized audio-visual feedback and corresponding manipulation policies.

In summary, multi-modal perception not only complements missing {or ambiguous} environmental information but also enables more flexible sensing {strategies} in extreme {or challenging conditions}. Exploring additional sensor {modalities and developing advanced} information fusion {methods are key to enhancing robotic adaptability in complex real-world environments}.

\section{Conclusion} \label{section6}
NeRF {introduces new opportunities} for robotics {by providing a powerful framework for} understanding and interacting with complex environments. It offers flexible and {high-fidelity} 3D scene representation, {along with learning-based approaches that benefit a range of robotic tasks, including} reconstruction, scene segmentation and editing, navigation, and manipulation. {While its potential to improve} realism, data efficiency, and adaptability {has been increasingly recognized, much remains to be explored to fully realize the synergy between NeRFs and robotics}. {Nevertheless}, integrating NeRF into robotic systems {presents significant} challenges, such as spatiotemporal map fusion, robust relocalization at the scene level, generalization across diverse environments, {bridging the gap between virtual rendering and real-world deployment, and incorporating multi-modal sensor interactions}. {These open challenges also point to numerous promising research opportunities} in this rapidly {advancing} field.

From {the perspective of} technical evolution, the field has {followed} a clear trajectory of advancement. 
In scene understanding, {early} works focus on static scene reconstruction {using} volumetric NeRFs{, which provide} dense mappings but {face challenges in} scalability and geometric {accuracy}. 
These limitations {spur the development of} hierarchical multi-MLP architectures, voxel-based grids, and hybrid volumetric-TSDF representations to {enhance} memory efficiency, scalability, and {reconstruction fidelity in} large-scale environments. For dynamic scenes, time-conditioned NeRFs evolve into deformation-based fields and flow-based methods, significantly improving temporal consistency and dynamic scene understanding, which is critical for long-term autonomous robot operation.
{Moreover}, NeRF {has expanded} beyond {purely} photometric modeling to {support} multimodal perception, {incorporating} semantic, instance-level, and panoptic segmentation. {Initial reliance} on large-scale supervised datasets has been {progressively alleviated by approaches utilizing sparse annotations}, self-supervised learning, and open-vocabulary models, {paving the way for more} flexible and generalizable perception frameworks {in robotics}. 

In {terms of} robotic interaction, NeRF has {advanced} from passive {scene} modeling to active {deployment} in real-time localization, planning, and manipulation tasks. Localization {techniques have transitioned} from pose regression on pre-trained NeRF maps to joint optimization of camera poses and neural scene {representations}, reducing dependence on {static} pre-built maps. Path planning has {evolved} from {basic} density-based avoidance to probabilistic modeling and {semantic-aware} navigation policies. In the domain of manipulation, NeRFs empower fine-grained modeling and tracking of objects under occlusions and articulations, and enable the fusion of visual and tactile sensing to facilitate robust grasping and in-hand manipulation.

The evolution of NeRF methods in robotics reflects a broader methodological shift in robot perception and interaction: real-time performance and memory efficiency, hybrid representation, adaptability and robustness, and multi-task integration.
(1) 
Early explorations prioritized dense scene encoding and accurate novel view synthesis, yet were constrained by scalability and computational inefficiency.
The focus subsequently moved to balancing representational richness with real-time performance and memory efficiency, which led to hierarchical, modular, and hybrid design
philosophies.
(2) 
A hybrid representation, combining the expressive capabilities of neural implicit fields with the structured reliability of explicit models, is advancing scalable, efficient, and general-purpose robotic systems.
(3) 
As robotic tasks face dynamic conditions and partial observability, NeRF-based approaches have increasingly incorporated temporal consistency, multi-modal priors, and learned uncertainty to enhance adaptability and robustness. 
(4) 
Moreover, multi-task integration has encouraged the development of unified models that fuse localization, mapping, semantic understanding, and decision-making within a shared representation space.

Fundamentally, robots {are often tasked with} solving the 3D inverse problem, which involves inferring physical {properties and} events in 3D space {based on} observations {from} sensors such as cameras, LiDAR, and tactile sensors.
NeRFs introduce a {transformative} paradigm for addressing this challenge in robotics. They leverage a differentiable, physics-based rendering pipeline to compare synthesized sensor observations with real-world measurements, using gradient-based optimization to infer a compact and consistent 3D representation of the environment.
This "effects-to-cause" reasoning framework {closely parallels} the human cognitive process of deducing underlying physical properties from surface observations. {As a result,} this research paradigm is expected to profoundly inspire future research directions in robotic perception and reasoning.

\section{Funding}
This work was supported in part by the Natural Science Foundation of China under Grants U1613218 and 61722309.

\section{Declaration of conflicting interests}
The author(s) declared no potential conflicts of interest with respect
to the research, authorship, and/or publication of this article.

\bibliographystyle{SageH}
\bibliography{bare_jrnl.bib}

\begin{thebibliography}{291}
\providecommand{\natexlab}[1]{#1}
\providecommand{\url}[1]{\texttt{#1}}
\providecommand{\urlprefix}{URL }
\expandafter\ifx\csname urlstyle\endcsname\relax
  \providecommand{\doi}[1]{DOI:\discretionary{}{}{}#1}\else
  \providecommand{\doi}{DOI:\discretionary{}{}{}\begingroup
  \urlstyle{rm}\Url}\fi

\bibitem[{Adamkiewicz et~al.(2022)Adamkiewicz, Chen, Caccavale, Gardner,
  Culbertson, Bohg and Schwager}]{adamkiewicz2022vision}
Adamkiewicz M, Chen T, Caccavale A, Gardner R, Culbertson P, Bohg J and
  Schwager M (2022) Vision-only robot navigation in a neural radiance world.
\newblock \emph{RA-L} : 4606--4613.

\bibitem[{Ahn et~al.(2022)Ahn, Jang, Park, Kim and Kang}]{ahn2022panerf}
Ahn YC, Jang S, Park S, Kim JY and Kang N (2022) Panerf: Pseudo-view
  augmentation for improved neural radiance fields based on few-shot inputs.
\newblock \emph{arXiv preprint arXiv:2211.12758} .

\bibitem[{Anderson et~al.(2018)Anderson, Chang, Chaplot, Dosovitskiy, Gupta,
  Koltun, Kosecka, Malik, Mottaghi, Savva et~al.}]{anderson2018evaluation}
Anderson P, Chang A, Chaplot DS, Dosovitskiy A, Gupta S, Koltun V, Kosecka J,
  Malik J, Mottaghi R, Savva M et~al. (2018) On evaluation of embodied
  navigation agents.
\newblock \emph{arXiv preprint arXiv:1807.06757} .

\bibitem[{Andersson et~al.(2020)Andersson, Nilsson, Akenine-M{\"o}ller,
  Oskarsson, {\AA}str{\"o}m and Fairchild}]{andersson2020flip}
Andersson P, Nilsson J, Akenine-M{\"o}ller T, Oskarsson M, {\AA}str{\"o}m K and
  Fairchild MD (2020) Flip: A difference evaluator for alternating images.
\newblock \emph{PACMCGIT} 3(2): 15--1.

\bibitem[{Avraham et~al.(2022)Avraham, Straub, Shen, Yang, Germain, Sweeney,
  Balntas, Novotny, DeTone and Newcombe}]{avraham2022nerfels}
Avraham G, Straub J, Shen T, Yang TY, Germain H, Sweeney C, Balntas V, Novotny
  D, DeTone D and Newcombe R (2022) Nerfels: Renderable neural codes for
  improved camera pose estimation.
\newblock In: \emph{CVPR}. pp. 5061--5070.

\bibitem[{Azinovi{\'c} et~al.(2022)Azinovi{\'c}, Martin-Brualla, Goldman,
  Nie{\ss}ner and Thies}]{azinovic2022neural}
Azinovi{\'c} D, Martin-Brualla R, Goldman DB, Nie{\ss}ner M and Thies J (2022)
  Neural rgb-d surface reconstruction.
\newblock In: \emph{CVPR}. pp. 6290--6301.

\bibitem[{Bao et~al.(2023)Bao, Zhang, Yang, Fan, Yang, Bao, Zhang and
  Cui}]{bao2023sine}
Bao C, Zhang Y, Yang B, Fan T, Yang Z, Bao H, Zhang G and Cui Z (2023) Sine:
  Semantic-driven image-based nerf editing with prior-guided editing field.
\newblock In: \emph{CVPR}. pp. 20919--20929.

\bibitem[{Barron et~al.(2021)Barron, Mildenhall, Tancik, Hedman, Martin-Brualla
  and Srinivasan}]{barron2021mip}
Barron JT, Mildenhall B, Tancik M, Hedman P, Martin-Brualla R and Srinivasan PP
  (2021) Mip-nerf: A multiscale representation for anti-aliasing neural
  radiance fields.
\newblock In: \emph{ICCV}. pp. 5855--5864.

\bibitem[{Barron et~al.(2022)Barron, Mildenhall, Verbin, Srinivasan and
  Hedman}]{barron2022mip}
Barron JT, Mildenhall B, Verbin D, Srinivasan PP and Hedman P (2022) Mip-nerf
  360: Unbounded anti-aliased neural radiance fields.
\newblock In: \emph{CVPR}. pp. 5470--5479.

\bibitem[{Betker et~al.(2023)Betker, Goh, Jing, Brooks, Wang, Li, Ouyang,
  Zhuang, Lee, Guo et~al.}]{betker2023improving}
Betker J, Goh G, Jing L, Brooks T, Wang J, Li L, Ouyang L, Zhuang J, Lee J, Guo
  Y et~al. (2023) Improving image generation with better captions.
\newblock \emph{Computer Science} : 8.

\bibitem[{Bi et~al.(2020)Bi, Xu, Srinivasan, Mildenhall, Sunkavalli,
  Ha{\v{s}}an, Hold-Geoffroy, Kriegman and Ramamoorthi}]{bi2020neural}
Bi S, Xu Z, Srinivasan P, Mildenhall B, Sunkavalli K, Ha{\v{s}}an M,
  Hold-Geoffroy Y, Kriegman D and Ramamoorthi R (2020) Neural reflectance
  fields for appearance acquisition.
\newblock \emph{arXiv preprint arXiv:2008.03824} .

\bibitem[{Bian et~al.(2023)Bian, Wang, Li, Bian and
  Prisacariu}]{bian2022nopenerf}
Bian W, Wang Z, Li K, Bian J and Prisacariu VA (2023) Nope-nerf: Optimising
  neural radiance field with no pose prior.

\bibitem[{Bi{\'n}kowski et~al.(2018)Bi{\'n}kowski, Sutherland, Arbel and
  Gretton}]{binkowski2018demystifying}
Bi{\'n}kowski M, Sutherland DJ, Arbel M and Gretton A (2018) Demystifying mmd
  gans.
\newblock \emph{ICLR} .

\bibitem[{Blomqvist et~al.(2023)Blomqvist, Ott, Chung and
  Siegwart}]{blomqvist2022baking}
Blomqvist K, Ott L, Chung JJ and Siegwart R (2023) Baking in the feature:
  Accelerating volumetric segmentation by rendering feature maps.
\newblock In: \emph{IROS}. IEEE, pp. 7629--7634.

\bibitem[{Blukis et~al.(2023)Blukis, Yoon, Lee, Tremblay, Wen, Kweon, Fox and
  Birchfield}]{blukis2023one}
Blukis V, Yoon KJ, Lee T, Tremblay J, Wen B, Kweon IS, Fox D and Birchfield S
  (2023) One-shot neural fields for 3d object understanding.
\newblock In: \emph{CVPRW}.

\bibitem[{Boss et~al.(2021{\natexlab{a}})Boss, Braun, Jampani, Barron, Liu and
  Lensch}]{boss2021nerd}
Boss M, Braun R, Jampani V, Barron JT, Liu C and Lensch H (2021{\natexlab{a}})
  Nerd: Neural reflectance decomposition from image collections.
\newblock In: \emph{ICCV}. pp. 12684--12694.

\bibitem[{Boss et~al.(2021{\natexlab{b}})Boss, Jampani, Braun, Liu, Barron and
  Lensch}]{boss2021neural}
Boss M, Jampani V, Braun R, Liu C, Barron J and Lensch H (2021{\natexlab{b}})
  Neural-pil: Neural pre-integrated lighting for reflectance decomposition.
\newblock \emph{NeurIPS} : 10691--10704.

\bibitem[{Brooks et~al.(2024)Brooks, Peebles, Holmes, DePue, Guo, Jing,
  Schnurr, Taylor, Luhman, Luhman, Ng, Wang and
  Ramesh}]{videoworldsimulators2024}
Brooks T, Peebles B, Holmes C, DePue W, Guo Y, Jing L, Schnurr D, Taylor J,
  Luhman T, Luhman E, Ng C, Wang R and Ramesh A (2024) Video generation models
  as world simulators
  \urlprefix\url{https://openai.com/research/video-generation-models-as-world-simulators}.

\bibitem[{Brunetto et~al.(2025)Brunetto, Hornauer and
  Moutarde}]{brunetto2024neraf}
Brunetto A, Hornauer S and Moutarde F (2025) Neraf: 3d scene infused neural
  radiance and acoustic fields.
\newblock \emph{ICLR} .

\bibitem[{B{\"u}sching et~al.(2024)B{\"u}sching, Bengtson, Nilsson and
  Bj{\"o}rkman}]{busching2024flowibr}
B{\"u}sching M, Bengtson J, Nilsson D and Bj{\"o}rkman M (2024) Flowibr:
  Leveraging pre-training for efficient neural image-based rendering of dynamic
  scenes.
\newblock In: \emph{CVPR}. pp. 8016--8026.

\bibitem[{Bylow et~al.(2013)Bylow, Sturm, Kerl, Kahl and
  Cremers}]{bylow2013real}
Bylow E, Sturm J, Kerl C, Kahl F and Cremers D (2013) Real-time camera tracking
  and 3d reconstruction using signed distance functions.
\newblock In: \emph{RSS}.

\bibitem[{Byravan et~al.(2023)Byravan, Humplik, Hasenclever, Brussee, Nori,
  Haarnoja, Moran, Bohez, Sadeghi, Vujatovic et~al.}]{byravan2023nerf2real}
Byravan A, Humplik J, Hasenclever L, Brussee A, Nori F, Haarnoja T, Moran B,
  Bohez S, Sadeghi F, Vujatovic B et~al. (2023) Nerf2real: Sim2real transfer of
  vision-guided bipedal motion skills using neural radiance fields.
\newblock In: \emph{ICRA}. pp. 9362--9369.

\bibitem[{Cao and Johnson(2023)}]{cao2023hexplane}
Cao A and Johnson J (2023) Hexplane: A fast representation for dynamic scenes.
\newblock In: \emph{CVPR}. pp. 130--141.

\bibitem[{Caron et~al.(2021)Caron, Touvron, Misra, J{\'e}gou, Mairal,
  Bojanowski and Joulin}]{caron2021emerging}
Caron M, Touvron H, Misra I, J{\'e}gou H, Mairal J, Bojanowski P and Joulin A
  (2021) Emerging properties in self-supervised vision transformers.
\newblock In: \emph{ICCV}. pp. 9650--9660.

\bibitem[{Cen et~al.(2023)Cen, Zhou, Fang, Shen, Xie, Jiang, Zhang, Tian
  et~al.}]{cen2023segment}
Cen J, Zhou Z, Fang J, Shen W, Xie L, Jiang D, Zhang X, Tian Q et~al. (2023)
  Segment anything in 3d with nerfs.
\newblock \emph{NeurIPS} : 25971--25990.

\bibitem[{Charles et~al.(2022)Charles, Abbeloos, Reino and
  Cipolla}]{charles2022style2nerf}
Charles J, Abbeloos W, Reino DO and Cipolla R (2022) Style2nerf: An
  unsupervised one-shot nerf for semantic 3d reconstruction.
\newblock In: \emph{BMVC}. p. 104.

\bibitem[{Chen et~al.(2022{\natexlab{a}})Chen, Xu, Geiger, Yu and
  Su}]{chen2022tensorf}
Chen A, Xu Z, Geiger A, Yu J and Su H (2022{\natexlab{a}}) Tensorf: Tensorial
  radiance fields.
\newblock In: \emph{ECCV}. pp. 333--350.

\bibitem[{Chen et~al.(2021{\natexlab{a}})Chen, Xu, Zhao, Zhang, Xiang, Yu and
  Su}]{chen2021mvsnerf}
Chen A, Xu Z, Zhao F, Zhang X, Xiang F, Yu J and Su H (2021{\natexlab{a}})
  Mvsnerf: Fast generalizable radiance field reconstruction from multi-view
  stereo.
\newblock In: \emph{ICCV}. pp. 14124--14133.

\bibitem[{Chen et~al.(2023{\natexlab{a}})Chen, Manhardt, Navab and
  Busam}]{chen2023texpose}
Chen H, Manhardt F, Navab N and Busam B (2023{\natexlab{a}}) Texpose: Neural
  texture learning for self-supervised 6d object pose estimation.
\newblock In: \emph{CVPR}. pp. 4841--4852.

\bibitem[{Chen et~al.(2023{\natexlab{b}})Chen, Lyu and
  Wang}]{chen2023neuraleditor}
Chen JK, Lyu J and Wang YX (2023{\natexlab{b}}) Neuraleditor: Editing neural
  radiance fields via manipulating point clouds.
\newblock In: \emph{CVPR}. pp. 12439--12448.

\bibitem[{Chen et~al.(2023{\natexlab{c}})Chen, Song, Bao and
  Zhou}]{chen2023perceiving}
Chen L, Song Y, Bao H and Zhou X (2023{\natexlab{c}}) Perceiving unseen 3d
  objects by poking the objects.
\newblock \emph{ICRA} .

\bibitem[{Chen et~al.(2022{\natexlab{b}})Chen, Li, Wang and
  Prisacariu}]{chen2022dfnet}
Chen S, Li X, Wang Z and Prisacariu V (2022{\natexlab{b}}) {DFN}et: {E}nhance
  absolute pose regression with direct feature matching.
\newblock In: \emph{ECCV}.

\bibitem[{Chen et~al.(2021{\natexlab{b}})Chen, Wang and
  Prisacariu}]{chen2021direct}
Chen S, Wang Z and Prisacariu V (2021{\natexlab{b}}) Direct-posenet: absolute
  pose regression with photometric consistency.
\newblock In: \emph{3DV}. pp. 1175--1185.

\bibitem[{Chen et~al.(2024{\natexlab{a}})Chen, Culbertson and
  Schwager}]{chen2023catnips}
Chen T, Culbertson P and Schwager M (2024{\natexlab{a}}) Catnips: Collision
  avoidance through neural implicit probabilistic scenes.
\newblock \emph{T-RO} .

\bibitem[{Chen et~al.(2022{\natexlab{c}})Chen, Zhang, Li, Chen, Feng, Wang and
  Wang}]{chen2022hallucinated}
Chen X, Zhang Q, Li X, Chen Y, Feng Y, Wang X and Wang J (2022{\natexlab{c}})
  Hallucinated neural radiance fields in the wild.
\newblock In: \emph{CVPR}. pp. 12943--12952.

\bibitem[{Chen et~al.(2024{\natexlab{b}})Chen, Yuan, Li, Liu, Wang, Xie, Wen
  and Yu}]{chen2022upst}
Chen Y, Yuan Q, Li Z, Liu Y, Wang W, Xie C, Wen X and Yu Q (2024{\natexlab{b}})
  Upst-nerf: Universal photorealistic style transfer of neural radiance fields
  for 3d scene.
\newblock \emph{T-VCG} .

\bibitem[{Cheng et~al.(2020)Cheng, Collins, Zhu, Liu, Huang, Adam and
  Chen}]{cheng2020panoptic}
Cheng B, Collins MD, Zhu Y, Liu T, Huang TS, Adam H and Chen LC (2020)
  Panoptic-deeplab: A simple, strong, and fast baseline for bottom-up panoptic
  segmentation.
\newblock In: \emph{CVPR}. pp. 12475--12485.

\bibitem[{Chibane et~al.(2021)Chibane, Bansal, Lazova and
  Pons-Moll}]{chibane2021stereo}
Chibane J, Bansal A, Lazova V and Pons-Moll G (2021) Stereo radiance fields
  (srf): Learning view synthesis for sparse views of novel scenes.
\newblock In: \emph{CVPR}. pp. 7911--7920.

\bibitem[{Chu et~al.(2018)Chu, Xie, Leal-Taix{\'e} and
  Thuerey}]{chu2018temporally}
Chu M, Xie Y, Leal-Taix{\'e} L and Thuerey N (2018) Temporally coherent gans
  for video super-resolution (tecogan).
\newblock \emph{ACM TOG} 1(2): 3.

\bibitem[{Chun et~al.(2023)Chun, Du, Simeonov, Lozano-Perez and
  Kaelbling}]{chun2023local}
Chun E, Du Y, Simeonov A, Lozano-Perez T and Kaelbling L (2023) Local neural
  descriptor fields: Locally conditioned object representations for
  manipulation.
\newblock \emph{ICRA} .

\bibitem[{Chung et~al.(2023)Chung, Tseng, Hsu, Shi, Hua, Yeh, Chen, Chen and
  Hsu}]{chung2023orbeez}
Chung CM, Tseng YC, Hsu YC, Shi XQ, Hua YH, Yeh JF, Chen WC, Chen YT and Hsu WH
  (2023) Orbeez-slam: A real-time monocular visual slam with orb features and
  nerf-realized mapping.
\newblock In: \emph{ICRA}. pp. 9400--9406.

\bibitem[{{\c{C}}i{\c{c}}ek et~al.(2016){\c{C}}i{\c{c}}ek, Abdulkadir,
  Lienkamp, Brox and Ronneberger}]{cciccek20163d}
{\c{C}}i{\c{c}}ek {\"O}, Abdulkadir A, Lienkamp SS, Brox T and Ronneberger O
  (2016) 3d u-net: learning dense volumetric segmentation from sparse
  annotation.
\newblock In: \emph{MICCAI}. pp. 424--432.

\bibitem[{Clarke et~al.(2023)Clarke, Gao, Wang, Rau, Xu, Wang, James and
  Wu}]{clarke2023realimpact}
Clarke S, Gao R, Wang M, Rau M, Xu J, Wang JH, James DL and Wu J (2023)
  Realimpact: A dataset of impact sound fields for real objects.
\newblock In: \emph{CVPR}. pp. 1516--1525.

\bibitem[{Dai et~al.(2017)Dai, Chang, Savva, Halber, Funkhouser and
  Nie{\ss}ner}]{dai2017scannet}
Dai A, Chang AX, Savva M, Halber M, Funkhouser T and Nie{\ss}ner M (2017)
  Scannet: Richly-annotated 3d reconstructions of indoor scenes.
\newblock In: \emph{CVPR}. pp. 5828--5839.

\bibitem[{Dai et~al.(2024)Dai, Gupta and Gao}]{dai2024neural}
Dai A, Gupta S and Gao G (2024) Neural elevation models for terrain mapping and
  path planning.
\newblock \emph{ICRA} .

\bibitem[{Dai et~al.(2023)Dai, Zhu, Geng, Ruan, Zhang and
  Wang}]{dai2023graspnerf}
Dai Q, Zhu Y, Geng Y, Ruan C, Zhang J and Wang H (2023) Graspnerf:
  multiview-based 6-dof grasp detection for transparent and specular objects
  using generalizable nerf.
\newblock In: \emph{ICRA}. pp. 1757--1763.

\bibitem[{Dellaert et~al.(1999)Dellaert, Fox, Burgard and
  Thrun}]{dellaert1999monte}
Dellaert F, Fox D, Burgard W and Thrun S (1999) Monte carlo localization for
  mobile robots.
\newblock In: \emph{ICRA}. pp. 1322--1328.

\bibitem[{Dellaert and Yen-Chen(2020)}]{dellaert2020neural}
Dellaert F and Yen-Chen L (2020) Neural volume rendering: Nerf and beyond.
\newblock \emph{arXiv preprint arXiv:2101.05204} .

\bibitem[{Deng et~al.(2023{\natexlab{a}})Deng, Jiang, Qi, Yan, Zhou, Guibas,
  Anguelov et~al.}]{deng2023nerdi}
Deng C, Jiang C, Qi CR, Yan X, Zhou Y, Guibas L, Anguelov D et~al.
  (2023{\natexlab{a}}) Nerdi: Single-view nerf synthesis with language-guided
  diffusion as general image priors.
\newblock In: \emph{CVPR}. pp. 20637--20647.

\bibitem[{Deng et~al.(2023{\natexlab{b}})Deng, Wu, Chen, Xia, Sun, Liu, Yu and
  Pei}]{deng2023nerf}
Deng J, Wu Q, Chen X, Xia S, Sun Z, Liu G, Yu W and Pei L (2023{\natexlab{b}})
  Nerf-loam: Neural implicit representation for large-scale incremental lidar
  odometry and mapping.
\newblock In: \emph{ICCV}.

\bibitem[{Deng et~al.(2022)Deng, Liu, Zhu and Ramanan}]{deng2022depth}
Deng K, Liu A, Zhu JY and Ramanan D (2022) Depth-supervised nerf: Fewer views
  and faster training for free.
\newblock In: \emph{CVPR}. pp. 12882--12891.

\bibitem[{Dey et~al.(2022)Dey, Ahmine and Comport}]{dey2022mip}
Dey A, Ahmine Y and Comport AI (2022) Mip-nerf rgb-d: Depth assisted fast
  neural radiance fields.
\newblock \emph{WSCG} .

\bibitem[{Driess et~al.(2023)Driess, Huang, Li, Tedrake and
  Toussaint}]{driess2023learning}
Driess D, Huang Z, Li Y, Tedrake R and Toussaint M (2023) Learning multi-object
  dynamics with compositional neural radiance fields.
\newblock In: \emph{CoRL}. pp. 1755--1768.

\bibitem[{Driess et~al.(2022)Driess, Schubert, Florence, Li and
  Toussaint}]{driess2022reinforcement}
Driess D, Schubert I, Florence P, Li Y and Toussaint M (2022) Reinforcement
  learning with neural radiance fields.
\newblock \emph{NeurIPS} .

\bibitem[{Drost et~al.(2017)Drost, Ulrich, Bergmann, Hartinger and
  Steger}]{drost2017introducing}
Drost B, Ulrich M, Bergmann P, Hartinger P and Steger C (2017) Introducing
  mvtec itodd-a dataset for 3d object recognition in industry.
\newblock In: \emph{ICCV workshops}. pp. 2200--2208.

\bibitem[{Du et~al.(2021)Du, Zhang, Yu, Tenenbaum and Wu}]{du2021neural}
Du Y, Zhang Y, Yu HX, Tenenbaum JB and Wu J (2021) Neural radiance flow for 4d
  view synthesis and video processing.
\newblock In: \emph{ICCV}. pp. 14304--14314.

\bibitem[{El~Banani et~al.(2021)El~Banani, Gao and
  Johnson}]{el2021unsupervisedr}
El~Banani M, Gao L and Johnson J (2021) Unsupervisedr\&r: Unsupervised point
  cloud registration via differentiable rendering.
\newblock In: \emph{CVPR}. pp. 7129--7139.

\bibitem[{Elia et~al.(2023)Elia, Leonard, Antonio, Matthias, Vladlen and
  Davide}]{Kaufmann2023drone}
Elia K, Leonard B, Antonio L, Matthias M, Vladlen K and Davide S (2023)
  Champion-level drone racing using deep reinforcement learning.
\newblock \emph{Nature} : 982–987.

\bibitem[{Fang et~al.(2023)Fang, Fang, Tang, Liu, Wang, Wang, Zhu and
  Lu}]{fang2023rh20t}
Fang HS, Fang H, Tang Z, Liu J, Wang C, Wang J, Zhu H and Lu C (2023) Rh20t: A
  comprehensive robotic dataset for learning diverse skills in one-shot.
\newblock \emph{RSS} .

\bibitem[{Fang et~al.(2022)Fang, Yi, Wang, Xie, Zhang, Liu, Nie{\ss}ner and
  Tian}]{fang2022fast}
Fang J, Yi T, Wang X, Xie L, Zhang X, Liu W, Nie{\ss}ner M and Tian Q (2022)
  Fast dynamic radiance fields with time-aware neural voxels.
\newblock In: \emph{SIGGRAPH}. pp. 1--9.

\bibitem[{Fridovich-Keil et~al.(2023)Fridovich-Keil, Meanti, Warburg, Recht and
  Kanazawa}]{fridovich2023k}
Fridovich-Keil S, Meanti G, Warburg FR, Recht B and Kanazawa A (2023) K-planes:
  Explicit radiance fields in space, time, and appearance.
\newblock In: \emph{CVPR}. pp. 12479--12488.

\bibitem[{Fridovich-Keil et~al.(2022)Fridovich-Keil, Yu, Tancik, Chen, Recht
  and Kanazawa}]{fridovich2022plenoxels}
Fridovich-Keil S, Yu A, Tancik M, Chen Q, Recht B and Kanazawa A (2022)
  Plenoxels: Radiance fields without neural networks.
\newblock In: \emph{CVPR}. pp. 5501--5510.

\bibitem[{Fu et~al.(2022)Fu, Zhang, Chen, Lu, Zhu, Zhou, Geiger and
  Liao}]{fu2022panoptic}
Fu X, Zhang S, Chen T, Lu Y, Zhu L, Zhou X, Geiger A and Liao Y (2022) Panoptic
  nerf: 3d-to-2d label transfer for panoptic urban scene segmentation.
\newblock \emph{3DV} .

\bibitem[{Gafni et~al.(2021)Gafni, Thies, Zollhofer and
  Nie{\ss}ner}]{gafni2021dynamic}
Gafni G, Thies J, Zollhofer M and Nie{\ss}ner M (2021) Dynamic neural radiance
  fields for monocular 4d facial avatar reconstruction.
\newblock In: \emph{CVPR}. pp. 8649--8658.

\bibitem[{Gao et~al.(2021)Gao, Saraf, Kopf and Huang}]{gao2021dynamic}
Gao C, Saraf A, Kopf J and Huang JB (2021) Dynamic view synthesis from dynamic
  monocular video.
\newblock In: \emph{ICCV}. pp. 5712--5721.

\bibitem[{Gao et~al.(2022)Gao, Gao, He, Lu, Xu and Li}]{gao2022nerf}
Gao K, Gao Y, He H, Lu D, Xu L and Li J (2022) Nerf: Neural radiance field in
  3d vision, a comprehensive review.
\newblock \emph{TPAMI} .

\bibitem[{Garbin et~al.(2021)Garbin, Kowalski, Johnson, Shotton and
  Valentin}]{garbin2021fastnerf}
Garbin SJ, Kowalski M, Johnson M, Shotton J and Valentin J (2021) Fastnerf:
  High-fidelity neural rendering at 200fps.
\newblock In: \emph{ICCV}. pp. 14346--14355.

\bibitem[{Gu et~al.(2023)Gu, Jiang, Li, Lu, Zhu, Shah, Zhang and
  Bennamoun}]{gu2023ue4}
Gu J, Jiang M, Li H, Lu X, Zhu G, Shah SAA, Zhang L and Bennamoun M (2023)
  Ue4-nerf: Neural radiance field for real-time rendering of large-scale scene.
\newblock \emph{NeurIPS} .

\bibitem[{Guo et~al.(2022)Guo, Peng, Lin, Wang, Zhang, Bao and
  Zhou}]{guo2022neural}
Guo H, Peng S, Lin H, Wang Q, Zhang G, Bao H and Zhou X (2022) Neural 3d scene
  reconstruction with the manhattan-world assumption.
\newblock In: \emph{CVPR}. pp. 5511--5520.

\bibitem[{Guo et~al.(2021)Guo, Chen, Liang, Liu, Bao and Zhang}]{guo2021ad}
Guo Y, Chen K, Liang S, Liu YJ, Bao H and Zhang J (2021) Ad-nerf: Audio driven
  neural radiance fields for talking head synthesis.
\newblock In: \emph{ICCV}. pp. 5784--5794.

\bibitem[{Hahn et~al.(2021)Hahn, Chaplot, Tulsiani, Mukadam, Rehg and
  Gupta}]{hahn2021no}
Hahn M, Chaplot DS, Tulsiani S, Mukadam M, Rehg JM and Gupta A (2021) No rl, no
  simulation: Learning to navigate without navigating.
\newblock \emph{Advances in Neural Information Processing Systems} 34:
  26661--26673.

\bibitem[{Hedman et~al.(2021)Hedman, Srinivasan, Mildenhall, Barron and
  Debevec}]{hedman2021baking}
Hedman P, Srinivasan PP, Mildenhall B, Barron JT and Debevec P (2021) Baking
  neural radiance fields for real-time view synthesis.
\newblock In: \emph{ICCV}. pp. 5875--5884.

\bibitem[{Heusel et~al.(2017)Heusel, Ramsauer, Unterthiner, Nessler and
  Hochreiter}]{heusel2017gans}
Heusel M, Ramsauer H, Unterthiner T, Nessler B and Hochreiter S (2017) Gans
  trained by a two time-scale update rule converge to a local nash equilibrium.
\newblock \emph{Advances in neural information processing systems} 30.

\bibitem[{Higuera et~al.(2023)Higuera, Dong, Boots and
  Mukadam}]{higuera2023neural}
Higuera C, Dong S, Boots B and Mukadam M (2023) Neural contact fields: Tracking
  extrinsic contact with tactile sensing.
\newblock In: \emph{ICRA}. pp. 12576--12582.

\bibitem[{Hinterstoisser et~al.(2013)Hinterstoisser, Lepetit, Ilic, Holzer,
  Bradski, Konolige and Navab}]{hinterstoisser2013model}
Hinterstoisser S, Lepetit V, Ilic S, Holzer S, Bradski G, Konolige K and Navab
  N (2013) Model based training, detection and pose estimation of texture-less
  3d objects in heavily cluttered scenes.
\newblock In: \emph{ACCV}. Springer, pp. 548--562.

\bibitem[{Hoda{\v{n}} et~al.(2016)Hoda{\v{n}}, Matas and
  Obdr{\v{z}}{\'a}lek}]{hodavn2016evaluation}
Hoda{\v{n}} T, Matas J and Obdr{\v{z}}{\'a}lek {\v{S}} (2016) On evaluation of
  6d object pose estimation.
\newblock In: \emph{ECCV}. Springer, pp. 606--619.

\bibitem[{Hodan et~al.(2018)Hodan, Michel, Brachmann, Kehl, GlentBuch, Kraft,
  Drost, Vidal, Ihrke, Zabulis et~al.}]{hodan2018bop}
Hodan T, Michel F, Brachmann E, Kehl W, GlentBuch A, Kraft D, Drost B, Vidal J,
  Ihrke S, Zabulis X et~al. (2018) Bop: Benchmark for 6d object pose
  estimation.
\newblock In: \emph{ECCV}. pp. 19--34.

\bibitem[{Hu et~al.(2023)Hu, Huang, Liu, Tai and Tang}]{hu2023nerf}
Hu B, Huang J, Liu Y, Tai YW and Tang CK (2023) Nerf-rpn: A general framework
  for object detection in nerfs.
\newblock In: \emph{CVPR}. pp. 23528--23538.

\bibitem[{Hu et~al.(2022{\natexlab{a}})Hu, Yu, Lanqing, Hu, Lee, Li
  et~al.}]{humasknerf}
Hu S, Yu L, Lanqing H, Hu T, Lee GH, Li Z et~al. (2022{\natexlab{a}}) Masknerf:
  Masked neural radiance fields for sparse view synthesis .

\bibitem[{Hu et~al.(2022{\natexlab{b}})Hu, Tan, Zhou, Woon and
  Lv}]{hu2022template}
Hu Z, Tan R, Zhou Y, Woon J and Lv C (2022{\natexlab{b}}) Template-based
  category-agnostic instance detection for robotic manipulation.
\newblock \emph{RA-L} : 12451--12458.

\bibitem[{Huang et~al.(2023{\natexlab{a}})Huang, Gojcic, Wang, Williams,
  Kasten, Fidler, Schindler and Litany}]{huang2023neural}
Huang S, Gojcic Z, Wang Z, Williams F, Kasten Y, Fidler S, Schindler K and
  Litany O (2023{\natexlab{a}}) Neural lidar fields for novel view synthesis.
\newblock In: \emph{ICCV}. pp. 18236--18246.

\bibitem[{Huang et~al.(2023{\natexlab{b}})Huang, Zhang, Feng, Li, Wang and
  Wang}]{huang2023local}
Huang X, Zhang Q, Feng Y, Li X, Wang X and Wang Q (2023{\natexlab{b}}) Local
  implicit ray function for generalizable radiance field representation.
\newblock In: \emph{CVPR}. pp. 97--107.

\bibitem[{Huang et~al.(2022)Huang, He, Yuan, Lai and
  Gao}]{huang2022stylizednerf}
Huang YH, He Y, Yuan YJ, Lai YK and Gao L (2022) Stylizednerf: consistent 3d
  scene stylization as stylized nerf via 2d-3d mutual learning.
\newblock In: \emph{CVPR}. pp. 18342--18352.

\bibitem[{Ichnowski et~al.(2021)Ichnowski, Avigal, Kerr and
  Goldberg}]{ichnowski2021dex}
Ichnowski J, Avigal Y, Kerr J and Goldberg K (2021) Dex-nerf: Using a neural
  radiance field to grasp transparent objects.
\newblock \emph{CoRL} .

\bibitem[{Irshad et~al.(2024)Irshad, Zakahrov, Guizilini, Gaidon, Kira and
  Ambrus}]{irshad2024nerf}
Irshad MZ, Zakahrov S, Guizilini V, Gaidon A, Kira Z and Ambrus R (2024)
  Nerf-mae: Masked autoencoders for self supervised 3d representation learning
  for neural radiance fields.
\newblock In: \emph{ECCV}.

\bibitem[{Irshad et~al.(2022)Irshad, Zakharov, Ambrus, Kollar, Kira and
  Gaidon}]{irshad2022shapo}
Irshad MZ, Zakharov S, Ambrus R, Kollar T, Kira Z and Gaidon A (2022) Shapo:
  Implicit representations for multi object shape appearance and pose
  optimization.
\newblock In: \emph{ECCV}.

\bibitem[{Irshad et~al.(2023)Irshad, Zakharov, Liu, Guizilini, Kollar, Gaidon,
  Kira and Ambrus}]{irshad2023neo}
Irshad MZ, Zakharov S, Liu K, Guizilini V, Kollar T, Gaidon A, Kira Z and
  Ambrus R (2023) Neo 360: Neural fields for sparse view synthesis of outdoor
  scenes.
\newblock In: \emph{ICCV}. pp. 9187--9198.

\bibitem[{Jain et~al.(2021)Jain, Tancik and Abbeel}]{jain2021putting}
Jain A, Tancik M and Abbeel P (2021) Putting nerf on a diet: Semantically
  consistent few-shot view synthesis.
\newblock In: \emph{ICCV}. pp. 5885--5894.

\bibitem[{Jang and Agapito(2021)}]{jang2021codenerf}
Jang W and Agapito L (2021) Codenerf: Disentangled neural radiance fields for
  object categories.
\newblock In: \emph{ICCV}. pp. 12949--12958.

\bibitem[{Jeong et~al.(2021)Jeong, Ahn, Choy, Anandkumar, Cho and
  Park}]{SCNeRF2021}
Jeong Y, Ahn S, Choy C, Anandkumar A, Cho M and Park J (2021) Self-calibrating
  neural radiance fields.
\newblock In: \emph{ICCV}.

\bibitem[{Johari et~al.(2023)Johari, Carta and Fleuret}]{johari2023eslam}
Johari MM, Carta C and Fleuret F (2023) Eslam: Efficient dense slam system
  based on hybrid representation of signed distance fields.
\newblock In: \emph{CVPR}. pp. 17408--17419.

\bibitem[{Kajiya and Von~Herzen(1984)}]{kajiya1984ray}
Kajiya JT and Von~Herzen BP (1984) Ray tracing volume densities.
\newblock \emph{ACM SIGGRAPH} : 165--174.

\bibitem[{K{\'a}roly et~al.(2020)K{\'a}roly, Galambos, Kuti and
  Rudas}]{karoly2020deep}
K{\'a}roly AI, Galambos P, Kuti J and Rudas IJ (2020) Deep learning in
  robotics: Survey on model structures and training strategies.
\newblock \emph{SMCS} : 266--279.

\bibitem[{Kerbl et~al.(2023)Kerbl, Kopanas, Leimk{\"u}hler and
  Drettakis}]{kerbl20233d}
Kerbl B, Kopanas G, Leimk{\"u}hler T and Drettakis G (2023) 3d gaussian
  splatting for real-time radiance field rendering.
\newblock \emph{ACM TOG} : 1--14.

\bibitem[{Kerr et~al.(2022)Kerr, Fu, Huang, Avigal, Tancik, Ichnowski, Kanazawa
  and Goldberg}]{kerr2022evo}
Kerr J, Fu L, Huang H, Avigal Y, Tancik M, Ichnowski J, Kanazawa A and Goldberg
  K (2022) Evo-nerf: Evolving nerf for sequential robot grasping of transparent
  objects.
\newblock In: \emph{CoRL}.

\bibitem[{Kerr et~al.(2024)Kerr, Kim, Wu, Yi, Wang, Goldberg and
  Kanazawa}]{kerr2024robot}
Kerr J, Kim CM, Wu M, Yi B, Wang Q, Goldberg K and Kanazawa A (2024) Robot see
  robot do: Imitating articulated object manipulation with monocular 4d
  reconstruction.
\newblock \emph{CoRL} .

\bibitem[{Khargonkar et~al.(2023)Khargonkar, Song, Xu, Prabhakaran and
  Xiang}]{khargonkar2023neuralgrasps}
Khargonkar N, Song N, Xu Z, Prabhakaran B and Xiang Y (2023) Neuralgrasps:
  Learning implicit representations for grasps of multiple robotic hands.
\newblock In: \emph{CoRL}. pp. 516--526.

\bibitem[{Kirillov et~al.(2019)Kirillov, He, Girshick, Rother and
  Doll{\'a}r}]{kirillov2019panoptic}
Kirillov A, He K, Girshick R, Rother C and Doll{\'a}r P (2019) Panoptic
  segmentation.
\newblock In: \emph{CVPR}. pp. 9404--9413.

\bibitem[{Kirillov et~al.(2023)Kirillov, Mintun, Ravi, Mao, Rolland, Gustafson,
  Xiao, Whitehead, Berg, Lo et~al.}]{kirillov2023segany}
Kirillov A, Mintun E, Ravi N, Mao H, Rolland C, Gustafson L, Xiao T, Whitehead
  S, Berg AC, Lo WY et~al. (2023) Segment anything.
\newblock In: \emph{ICCV}. pp. 4015--4026.

\bibitem[{Kobayashi et~al.(2022)Kobayashi, Matsumoto and
  Sitzmann}]{kobayashi2022decomposing}
Kobayashi S, Matsumoto E and Sitzmann V (2022) Decomposing nerf for editing via
  feature field distillation.
\newblock \emph{NeurIPS} : 23311--23330.

\bibitem[{Koenker and Hallock(2001)}]{koenker2001quantile}
Koenker R and Hallock KF (2001) Quantile regression.
\newblock \emph{JEP} .

\bibitem[{Kruzhkov et~al.(2022)Kruzhkov, Savinykh, Karpyshev, Kurenkov, Yudin,
  Potapov and Tsetserukou}]{kruzhkov2022meslam}
Kruzhkov E, Savinykh A, Karpyshev P, Kurenkov M, Yudin E, Potapov A and
  Tsetserukou D (2022) Meslam: Memory efficient slam based on neural fields.
\newblock In: \emph{SMC}. pp. 430--435.

\bibitem[{Kuang et~al.(2022)Kuang, Chen, Guadagnino, Zimmerman, Behley and
  Stachniss}]{kuang2022ir}
Kuang H, Chen X, Guadagnino T, Zimmerman N, Behley J and Stachniss C (2022)
  Ir-mcl: Implicit representation-based online global localization.
\newblock \emph{RA-L} .

\bibitem[{Kundu et~al.(2022)Kundu, Genova, Yin, Fathi, Pantofaru, Guibas,
  Tagliasacchi, Dellaert and Funkhouser}]{kundu2022panoptic}
Kundu A, Genova K, Yin X, Fathi A, Pantofaru C, Guibas LJ, Tagliasacchi A,
  Dellaert F and Funkhouser T (2022) Panoptic neural fields: A semantic
  object-aware neural scene representation.
\newblock In: \emph{CVPR}. pp. 12871--12881.

\bibitem[{Kurenkov et~al.(2022)Kurenkov, Potapov, Savinykh, Yudin, Kruzhkov,
  Karpyshev and Tsetserukou}]{kurenkov2022nfomp}
Kurenkov M, Potapov A, Savinykh A, Yudin E, Kruzhkov E, Karpyshev P and
  Tsetserukou D (2022) Nfomp: Neural field for optimal motion planner of
  differential drive robots with nonholonomic constraints.
\newblock \emph{RA-L} : 10991--10998.

\bibitem[{Kwon et~al.(2023)Kwon, Park and Oh}]{kwon2023renderable}
Kwon O, Park J and Oh S (2023) Renderable neural radiance map for visual
  navigation.
\newblock In: \emph{CVPR}. pp. 9099--9108.

\bibitem[{Lee et~al.(2020)Lee, Hwangbo, Wellhausen, Koltun and
  Hutter}]{lee2020learning}
Lee J, Hwangbo J, Wellhausen L, Koltun V and Hutter M (2020) Learning
  quadrupedal locomotion over challenging terrain.
\newblock \emph{Science robotics} : eabc5986.

\bibitem[{Lee et~al.(2022)Lee, Chen, Wang, Liniger, Kumar and
  Yu}]{lee2022uncertainty}
Lee S, Chen L, Wang J, Liniger A, Kumar S and Yu F (2022) Uncertainty guided
  policy for active robotic 3d reconstruction using neural radiance fields.
\newblock \emph{RA-L} .

\bibitem[{Lewis et~al.(2022)Lewis, Pavlasek and Jenkins}]{lewis2022narf22}
Lewis S, Pavlasek J and Jenkins OC (2022) Narf22: Neural articulated radiance
  fields for configuration-aware rendering.
\newblock In: \emph{IROS}. pp. 770--777.

\bibitem[{Li et~al.(2022{\natexlab{a}})Li, Weinberger, Belongie, Koltun and
  Ranftl}]{li2022language}
Li B, Weinberger KQ, Belongie S, Koltun V and Ranftl R (2022{\natexlab{a}})
  Language-driven semantic segmentation.
\newblock \emph{arXiv preprint arXiv:2201.03546} .

\bibitem[{Li et~al.(2022{\natexlab{b}})Li, Weinberger, Belongie, Koltun and
  Ranftl}]{li2022languagedriven}
Li B, Weinberger KQ, Belongie S, Koltun V and Ranftl R (2022{\natexlab{b}})
  Language-driven semantic segmentation.
\newblock In: \emph{ICLR}.

\bibitem[{Li et~al.(2023{\natexlab{a}})Li, Vutukur, Yu, Shugurov, Busam, Yang
  and Ilic}]{li2023nerf}
Li F, Vutukur SR, Yu H, Shugurov I, Busam B, Yang S and Ilic S
  (2023{\natexlab{a}}) Nerf-pose: A first-reconstruct-then-regress approach for
  weakly-supervised 6d object pose estimation.
\newblock In: \emph{ICCV}. pp. 2123--2133.

\bibitem[{Li et~al.(2022{\natexlab{c}})Li, Zhang, Zhu, Wang, Lee, Xu, Adelson,
  Fei-Fei, Gao and Wu}]{li2022see}
Li H, Zhang Y, Zhu J, Wang S, Lee MA, Xu H, Adelson E, Fei-Fei L, Gao R and Wu
  J (2022{\natexlab{c}}) See, hear, and feel: Smart sensory fusion for robotic
  manipulation.
\newblock \emph{CoRL} .

\bibitem[{Li et~al.(2021{\natexlab{a}})Li, Feng, She, Ding, Wang and
  Lee}]{li2021mine}
Li J, Feng Z, She Q, Ding H, Wang C and Lee GH (2021{\natexlab{a}}) Mine:
  Towards continuous depth mpi with nerf for novel view synthesis.
\newblock In: \emph{ICCV}. pp. 12578--12588.

\bibitem[{Li et~al.(2022{\natexlab{d}})Li, Tang, Prisacariu and
  Torr}]{li2022bnv}
Li K, Tang Y, Prisacariu VA and Torr PH (2022{\natexlab{d}}) Bnv-fusion: Dense
  3d reconstruction using bi-level neural volume fusion.
\newblock In: \emph{CVPR}. pp. 6166--6175.

\bibitem[{Li et~al.(2022{\natexlab{e}})Li, Slavcheva, Zollhoefer, Green,
  Lassner, Kim, Schmidt, Lovegrove, Goesele, Newcombe et~al.}]{li2022neural}
Li T, Slavcheva M, Zollhoefer M, Green S, Lassner C, Kim C, Schmidt T,
  Lovegrove S, Goesele M, Newcombe R et~al. (2022{\natexlab{e}}) Neural 3d
  video synthesis from multi-view video.
\newblock In: \emph{CVPR}. pp. 5521--5531.

\bibitem[{Li et~al.(2022{\natexlab{f}})Li, Hong, Wang, Cao, Xian and
  Lin}]{li2022symmnerf}
Li X, Hong C, Wang Y, Cao Z, Xian K and Lin G (2022{\natexlab{f}}) Symmnerf:
  Learning to explore symmetry prior for single-view view synthesis.
\newblock In: \emph{ACCV}. pp. 1726--1742.

\bibitem[{Li et~al.(2022{\natexlab{g}})Li, Li, Sitzmann, Agrawal and
  Torralba}]{li20223d}
Li Y, Li S, Sitzmann V, Agrawal P and Torralba A (2022{\natexlab{g}}) 3d neural
  scene representations for visuomotor control.
\newblock In: \emph{CoRL}. pp. 112--123.

\bibitem[{Li et~al.(2023{\natexlab{b}})Li, Lin, Forsyth, Huang and
  Wang}]{li2023climatenerf}
Li Y, Lin ZH, Forsyth D, Huang JB and Wang S (2023{\natexlab{b}}) Climatenerf:
  Extreme weather synthesis in neural radiance field.
\newblock In: \emph{ICCV}. pp. 3227--3238.

\bibitem[{Li et~al.(2021{\natexlab{b}})Li, Niklaus, Snavely and
  Wang}]{li2021neural}
Li Z, Niklaus S, Snavely N and Wang O (2021{\natexlab{b}}) Neural scene flow
  fields for space-time view synthesis of dynamic scenes.
\newblock In: \emph{CVPR}. pp. 6498--6508.

\bibitem[{Liang et~al.(2022)Liang, Liu, Wu, Tai and Tang}]{liang2022onerf}
Liang S, Liu Y, Wu S, Tai YW and Tang CK (2022) Onerf: Unsupervised 3d object
  segmentation from multiple views.
\newblock \emph{NeurIPS} .

\bibitem[{Liao et~al.(2022)Liao, Xie and Geiger}]{liao2022kitti}
Liao Y, Xie J and Geiger A (2022) Kitti-360: A novel dataset and benchmarks for
  urban scene understanding in 2d and 3d.
\newblock \emph{TPAMI} : 3292--3310.

\bibitem[{Lin et~al.(2021)Lin, Ma, Torralba and Lucey}]{lin2021barf}
Lin CH, Ma WC, Torralba A and Lucey S (2021) Barf: Bundle-adjusting neural
  radiance fields.
\newblock In: \emph{ICCV}. pp. 5741--5751.

\bibitem[{Lin et~al.(2022)Lin, Peng, Xu, Yan, Shuai, Bao and
  Zhou}]{lin2022efficient}
Lin H, Peng S, Xu Z, Yan Y, Shuai Q, Bao H and Zhou X (2022) Efficient neural
  radiance fields for interactive free-viewpoint video.
\newblock In: \emph{SIGGRAPH Asia}. pp. 1--9.

\bibitem[{Lin et~al.(2023{\natexlab{a}})Lin, M{\"u}ller, Tremblay, Wen, Tyree,
  Evans, Vela and Birchfield}]{lin2023parallel}
Lin Y, M{\"u}ller T, Tremblay J, Wen B, Tyree S, Evans A, Vela PA and
  Birchfield S (2023{\natexlab{a}}) Parallel inversion of neural radiance
  fields for robust pose estimation.
\newblock In: \emph{ICRA}. pp. 9377--9384.

\bibitem[{Lin et~al.(2023{\natexlab{b}})Lin, Florence, Zeng, Barron, Du, Ma,
  Simeonov, Garcia and Isola}]{lin2023mira}
Lin YC, Florence P, Zeng A, Barron JT, Du Y, Ma WC, Simeonov A, Garcia AR and
  Isola P (2023{\natexlab{b}}) Mira: Mental imagery for robotic affordances.
\newblock In: \emph{CoRL}. pp. 1916--1927.

\bibitem[{Lin et~al.(2024)Lin, Pan, Fridovich-Keil and
  Wetzstein}]{lin2024thermalnerf}
Lin YY, Pan XY, Fridovich-Keil S and Wetzstein G (2024) Thermalnerf: Thermal
  radiance fields.
\newblock In: \emph{ICCP}.

\bibitem[{Lisus and Holmes(2023)}]{lisus2023towards}
Lisus D and Holmes C (2023) Towards open world nerf-based slam.
\newblock \emph{CRV} .

\bibitem[{Liu et~al.(2022{\natexlab{a}})Liu, Shen, Chen et~al.}]{liu2022nerf}
Liu HK, Shen I, Chen BY et~al. (2022{\natexlab{a}}) Nerf-in: Free-form nerf
  inpainting with rgb-d priors.
\newblock \emph{CG\&A} : 100--109.

\bibitem[{Liu et~al.(2020)Liu, Gu, Zaw~Lin, Chua and Theobalt}]{liu2020neural}
Liu L, Gu J, Zaw~Lin K, Chua TS and Theobalt C (2020) Neural sparse voxel
  fields.
\newblock \emph{NeurIPS} : 15651--15663.

\bibitem[{Liu et~al.(2021)Liu, Zhang, Zhang, Zhang, Zhu and
  Russell}]{liu2021editing}
Liu S, Zhang X, Zhang Z, Zhang R, Zhu JY and Russell B (2021) Editing
  conditional radiance fields.
\newblock In: \emph{ICCV}. pp. 5773--5783.

\bibitem[{Liu et~al.(2022{\natexlab{b}})Liu, Peng, Liu, Wang, Wang, Theobalt,
  Zhou and Wang}]{liu2022neural}
Liu Y, Peng S, Liu L, Wang Q, Wang P, Theobalt C, Zhou X and Wang W
  (2022{\natexlab{b}}) Neural rays for occlusion-aware image-based rendering.
\newblock In: \emph{CVPR}. pp. 7824--7833.

\bibitem[{Liu et~al.(2023{\natexlab{a}})Liu, Gao, Meuleman, Tseng, Saraf, Kim,
  Chuang, Kopf and Huang}]{liu2023robust}
Liu YL, Gao C, Meuleman A, Tseng HY, Saraf A, Kim C, Chuang YY, Kopf J and
  Huang JB (2023{\natexlab{a}}) Robust dynamic radiance fields.
\newblock \emph{CVPR} .

\bibitem[{Liu et~al.(2024)Liu, Chi, Cousineau, Kuppuswamy, Burchfiel and
  Song}]{liu2024maniwav}
Liu Z, Chi C, Cousineau E, Kuppuswamy N, Burchfiel B and Song S (2024) Maniwav:
  Learning robot manipulation from in-the-wild audio-visual data.
\newblock In: \emph{CoRL}.

\bibitem[{Liu et~al.(2023{\natexlab{b}})Liu, Milano, Frey, Hutter, Siegwart,
  Blum and Cadena}]{liu2022unsupervised}
Liu Z, Milano F, Frey J, Hutter M, Siegwart R, Blum H and Cadena C
  (2023{\natexlab{b}}) Unsupervised continual semantic adaptation through
  neural rendering.
\newblock \emph{CVPR} .

\bibitem[{Lu et~al.(2023)Lu, Xu, Chen, Li, Lin and Jiang}]{lu2023urban}
Lu F, Xu Y, Chen G, Li H, Lin KY and Jiang C (2023) Urban radiance field
  representation with deformable neural mesh primitives.
\newblock In: \emph{ICCV}. pp. 465--476.

\bibitem[{Luo et~al.(2022)Luo, Du, Tarr, Tenenbaum, Torralba and
  Gan}]{luo2022learning}
Luo A, Du Y, Tarr M, Tenenbaum J, Torralba A and Gan C (2022) Learning neural
  acoustic fields.
\newblock \emph{NeurIPS} 35: 3165--3177.

\bibitem[{Maggio et~al.(2023)Maggio, Abate, Shi, Mario and
  Carlone}]{maggio2022loc}
Maggio D, Abate M, Shi J, Mario C and Carlone L (2023) Loc-nerf: Monte carlo
  localization using neural radiance fields.
\newblock \emph{ICRA} .

\bibitem[{Mantiuk et~al.(2021)Mantiuk, Denes, Chapiro, Kaplanyan, Rufo, Bachy,
  Lian and Patney}]{mantiuk2021fovvideovdp}
Mantiuk RK, Denes G, Chapiro A, Kaplanyan A, Rufo G, Bachy R, Lian T and Patney
  A (2021) Fovvideovdp: A visible difference predictor for wide field-of-view
  video.
\newblock \emph{ACM TOG} 40(4): 1--19.

\bibitem[{Martin-Brualla et~al.(2021)Martin-Brualla, Radwan, Sajjadi, Barron,
  Dosovitskiy and Duckworth}]{martin2021nerf}
Martin-Brualla R, Radwan N, Sajjadi MS, Barron JT, Dosovitskiy A and Duckworth
  D (2021) Nerf in the wild: Neural radiance fields for unconstrained photo
  collections.
\newblock In: \emph{CVPR}. pp. 7210--7219.

\bibitem[{Marza et~al.(2024)Marza, Matignon, Simonin, Batra, Wolf and
  Chaplot}]{marza2023autonerf}
Marza P, Matignon L, Simonin O, Batra D, Wolf C and Chaplot DS (2024) Autonerf:
  Training implicit scene representations with autonomous agents.
\newblock \emph{ICLR} .

\bibitem[{Marza et~al.(2023)Marza, Matignon, Simonin and Wolf}]{marza2022multi}
Marza P, Matignon L, Simonin O and Wolf C (2023) Multi-object navigation with
  dynamically learned neural implicit representations.
\newblock In: \emph{ICCV}. pp. 11004--11015.

\bibitem[{Meng et~al.(2021)Meng, Chen, Luo, Wu, Su, Xu, He and
  Yu}]{meng2021gnerf}
Meng Q, Chen A, Luo H, Wu M, Su H, Xu L, He X and Yu J (2021) Gnerf: Gan-based
  neural radiance field without posed camera.
\newblock In: \emph{ICCV}. pp. 6351--6361.

\bibitem[{Menolotto et~al.(2020)Menolotto, Komaris, Tedesco, O’Flynn and
  Walsh}]{menolotto2020motion}
Menolotto M, Komaris DS, Tedesco S, O’Flynn B and Walsh M (2020) Motion
  capture technology in industrial applications: A systematic review.
\newblock \emph{Sensors} 20(19): 5687.

\bibitem[{Meuleman et~al.(2023)Meuleman, Liu, Gao, Huang, Kim, Kim and
  Kopf}]{meuleman2023localrf}
Meuleman A, Liu YL, Gao C, Huang JB, Kim C, Kim MH and Kopf J (2023)
  Progressively optimized local radiance fields for robust view synthesis.
\newblock In: \emph{CVPR}.

\bibitem[{Mildenhall et~al.(2020)Mildenhall, Srinivasan, Tancik, Barron,
  Ramamoorthi and Ng}]{mildenhall2020nerf}
Mildenhall B, Srinivasan PP, Tancik M, Barron JT, Ramamoorthi R and Ng R (2020)
  Nerf: Representing scenes as neural radiance fields for view synthesis.
\newblock In: \emph{ECCV}.

\bibitem[{Ming et~al.(2022)Ming, Ye and Calway}]{ming2022idf}
Ming Y, Ye W and Calway A (2022) idf-slam: End-to-end rgb-d slam with neural
  implicit mapping and deep feature tracking.
\newblock \emph{arXiv preprint arXiv:2209.07919} .

\bibitem[{Mirzaei et~al.(2023)Mirzaei, Aumentado-Armstrong, Derpanis, Kelly,
  Brubaker, Gilitschenski and Levinshtein}]{mirzaei2023spin}
Mirzaei A, Aumentado-Armstrong T, Derpanis KG, Kelly J, Brubaker MA,
  Gilitschenski I and Levinshtein A (2023) Spin-nerf: Multiview segmentation
  and perceptual inpainting with neural radiance fields.
\newblock In: \emph{CVPR}. pp. 20669--20679.

\bibitem[{Mirzaei et~al.(2022)Mirzaei, Kant, Kelly and
  Gilitschenski}]{mirzaei2022laterf}
Mirzaei A, Kant Y, Kelly J and Gilitschenski I (2022) Laterf: Label and text
  driven object radiance fields.
\newblock In: \emph{ECCV}. pp. 20--36.

\bibitem[{Moreau et~al.(2022)Moreau, Piasco, Tsishkou, Stanciulescu and
  de~La~Fortelle}]{moreau2022lens}
Moreau A, Piasco N, Tsishkou D, Stanciulescu B and de~La~Fortelle A (2022)
  Lens: Localization enhanced by nerf synthesis.
\newblock In: \emph{CoRL}. pp. 1347--1356.

\bibitem[{M{\"u}ller et~al.(2023)M{\"u}ller, Siddiqui, Porzi, Bulo,
  Kontschieder and Nie{\ss}ner}]{muller2023diffrf}
M{\"u}ller N, Siddiqui Y, Porzi L, Bulo SR, Kontschieder P and Nie{\ss}ner M
  (2023) Diffrf: Rendering-guided 3d radiance field diffusion.
\newblock In: \emph{CVPR}. pp. 4328--4338.

\bibitem[{M{\"u}ller et~al.(2022)M{\"u}ller, Evans, Schied and
  Keller}]{muller2022instant}
M{\"u}ller T, Evans A, Schied C and Keller A (2022) Instant neural graphics
  primitives with a multiresolution hash encoding.
\newblock \emph{ACM TOG} : 1--15.

\bibitem[{Murez et~al.(2020)Murez, Van~As, Bartolozzi, Sinha, Badrinarayanan
  and Rabinovich}]{murez2020atlas}
Murez Z, Van~As T, Bartolozzi J, Sinha A, Badrinarayanan V and Rabinovich A
  (2020) Atlas: End-to-end 3d scene reconstruction from posed images.
\newblock In: \emph{Computer Vision--ECCV 2020: 16th European Conference,
  Glasgow, UK, August 23--28, 2020, Proceedings, Part VII 16}. Springer, pp.
  414--431.

\bibitem[{Neff et~al.(2021)Neff, Stadlbauer, Parger, Kurz, Mueller, Chaitanya,
  Kaplanyan and Steinberger}]{neff2021donerf}
Neff T, Stadlbauer P, Parger M, Kurz A, Mueller JH, Chaitanya CRA, Kaplanyan A
  and Steinberger M (2021) Donerf: Towards real-time rendering of compact
  neural radiance fields using depth oracle networks.
\newblock In: \emph{CGF}. pp. 45--59.

\bibitem[{Newcombe et~al.(2011)Newcombe, Izadi, Hilliges, Molyneaux, Kim,
  Davison, Kohi, Shotton, Hodges and Fitzgibbon}]{newcombe2011kinectfusion}
Newcombe RA, Izadi S, Hilliges O, Molyneaux D, Kim D, Davison AJ, Kohi P,
  Shotton J, Hodges S and Fitzgibbon A (2011) Kinectfusion: Real-time dense
  surface mapping and tracking.
\newblock In: \emph{ISMAR}. pp. 127--136.

\bibitem[{Niemeyer et~al.(2022)Niemeyer, Barron, Mildenhall, Sajjadi, Geiger
  and Radwan}]{niemeyer2022regnerf}
Niemeyer M, Barron JT, Mildenhall B, Sajjadi MS, Geiger A and Radwan N (2022)
  Regnerf: Regularizing neural radiance fields for view synthesis from sparse
  inputs.
\newblock In: \emph{CVPR}. pp. 5480--5490.

\bibitem[{Oechsle et~al.(2021)Oechsle, Peng and Geiger}]{oechsle2021unisurf}
Oechsle M, Peng S and Geiger A (2021) Unisurf: Unifying neural implicit
  surfaces and radiance fields for multi-view reconstruction.
\newblock In: \emph{ICCV}. pp. 5589--5599.

\bibitem[{Or-El et~al.(2022)Or-El, Luo, Shan, Shechtman, Park and
  Kemelmacher-Shlizerman}]{or2022stylesdf}
Or-El R, Luo X, Shan M, Shechtman E, Park JJ and Kemelmacher-Shlizerman I
  (2022) Stylesdf: High-resolution 3d-consistent image and geometry generation.
\newblock In: \emph{CVPR}. pp. 13503--13513.

\bibitem[{Ost et~al.(2021)Ost, Mannan, Thuerey, Knodt and
  Heide}]{ost2021neural}
Ost J, Mannan F, Thuerey N, Knodt J and Heide F (2021) Neural scene graphs for
  dynamic scenes.
\newblock In: \emph{CVPR}. pp. 2856--2865.

\bibitem[{Park et~al.(2018)Park, Rematas, Farhadi and
  Seitz}]{park2018photoshape}
Park K, Rematas K, Farhadi A and Seitz SM (2018) Photoshape: Photorealistic
  materials for large-scale shape collections.
\newblock \emph{SIGGRAPH Asia} .

\bibitem[{Park et~al.(2021{\natexlab{a}})Park, Sinha, Barron, Bouaziz, Goldman,
  Seitz and Martin-Brualla}]{park2021nerfies}
Park K, Sinha U, Barron JT, Bouaziz S, Goldman DB, Seitz SM and Martin-Brualla
  R (2021{\natexlab{a}}) Nerfies: Deformable neural radiance fields.
\newblock In: \emph{ICCV}. pp. 5865--5874.

\bibitem[{Park et~al.(2021{\natexlab{b}})Park, Sinha, Hedman, Barron, Bouaziz,
  Goldman, Martin-Brualla and Seitz}]{park2021hypernerf}
Park K, Sinha U, Hedman P, Barron JT, Bouaziz S, Goldman DB, Martin-Brualla R
  and Seitz SM (2021{\natexlab{b}}) Hypernerf: A higher-dimensional
  representation for topologically varying neural radiance fields.
\newblock \emph{ACM TOG} .

\bibitem[{Partha et~al.(2023)Partha, Gupta and Gao}]{partha2023neural}
Partha M, Gupta S and Gao G (2023) Neural city maps: A case for 3d urban
  environment representations based on radiance fields.
\newblock In: \emph{ION GNSS+ 2023}. pp. 1953--1973.

\bibitem[{Partha et~al.(2024)Partha, Neamati, Gupta and Gao}]{partha2024robust}
Partha M, Neamati D, Gupta S and Gao G (2024) Robust 3d map-matching with
  visual environment features for neural city maps.
\newblock In: \emph{ION GNSS+ 2024}. pp. 2080--2095.

\bibitem[{Patel et~al.(2023)Patel, Pham and Bera}]{patel2023dronerf}
Patel D, Pham P and Bera A (2023) Dronerf: Real-time multi-agent drone pose
  optimization for computing neural radiance fields.
\newblock In: \emph{IROS}.

\bibitem[{Pavllo et~al.(2023)Pavllo, Tan, Rakotosaona and
  Tombari}]{pavllo2023shape}
Pavllo D, Tan DJ, Rakotosaona MJ and Tombari F (2023) Shape, pose, and
  appearance from a single image via bootstrapped radiance field inversion.
\newblock In: \emph{CVPR}. pp. 4391--4401.

\bibitem[{Peng et~al.(2022)Peng, Yan, Liu, Cheng, Guan, Pan, Zhai and
  Yang}]{peng2022cagenerf}
Peng Y, Yan Y, Liu S, Cheng Y, Guan S, Pan B, Zhai G and Yang X (2022)
  Cagenerf: Cage-based neural radiance field for generalized 3d deformation and
  animation.
\newblock \emph{NeurIPS} : 31402--31415.

\bibitem[{Piala and Clark(2021)}]{piala2021terminerf}
Piala M and Clark R (2021) Terminerf: Ray termination prediction for efficient
  neural rendering.
\newblock In: \emph{3DV}. pp. 1106--1114.

\bibitem[{Pumarola et~al.(2021)Pumarola, Corona, Pons-Moll and
  Moreno-Noguer}]{pumarola2021d}
Pumarola A, Corona E, Pons-Moll G and Moreno-Noguer F (2021) D-nerf: Neural
  radiance fields for dynamic scenes.
\newblock In: \emph{CVPR}. pp. 10318--10327.

\bibitem[{Rabby and Zhang(2023)}]{rabby2023beyondpixels}
Rabby A and Zhang C (2023) Beyondpixels: A comprehensive review of the
  evolution of neural radiance fields.
\newblock \emph{JACM} .

\bibitem[{Radford et~al.(2021)Radford, Kim, Hallacy, Ramesh, Goh, Agarwal,
  Sastry, Askell, Mishkin, Clark et~al.}]{radford2021learning}
Radford A, Kim JW, Hallacy C, Ramesh A, Goh G, Agarwal S, Sastry G, Askell A,
  Mishkin P, Clark J et~al. (2021) Learning transferable visual models from
  natural language supervision.
\newblock In: \emph{ICML}. pp. 8748--8763.

\bibitem[{Ran et~al.(2023)Ran, Zeng, He, Chen, Li, Chen, Lee and
  Ye}]{ran2023neurar}
Ran Y, Zeng J, He S, Chen J, Li L, Chen Y, Lee G and Ye Q (2023) Neurar: Neural
  uncertainty for autonomous 3d reconstruction with implicit neural
  representations.
\newblock \emph{RA-L} : 1125--1132.

\bibitem[{Rebain et~al.(2021)Rebain, Jiang, Yazdani, Li, Yi and
  Tagliasacchi}]{rebain2021derf}
Rebain D, Jiang W, Yazdani S, Li K, Yi KM and Tagliasacchi A (2021) Derf:
  Decomposed radiance fields.
\newblock In: \emph{CVPR}. pp. 14153--14161.

\bibitem[{Reiser et~al.(2021)Reiser, Peng, Liao and
  Geiger}]{reiser2021kilonerf}
Reiser C, Peng S, Liao Y and Geiger A (2021) Kilonerf: Speeding up neural
  radiance fields with thousands of tiny mlps.
\newblock In: \emph{ICCV}. pp. 14335--14345.

\bibitem[{Reizenstein et~al.(2021)Reizenstein, Shapovalov, Henzler, Sbordone,
  Labatut and Novotny}]{reizenstein2021common}
Reizenstein J, Shapovalov R, Henzler P, Sbordone L, Labatut P and Novotny D
  (2021) Common objects in 3d: Large-scale learning and evaluation of real-life
  3d category reconstruction.
\newblock In: \emph{ICCV}. pp. 10901--10911.

\bibitem[{Rematas et~al.(2022)Rematas, Liu, Srinivasan, Barron, Tagliasacchi,
  Funkhouser and Ferrari}]{rematas2022urban}
Rematas K, Liu A, Srinivasan PP, Barron JT, Tagliasacchi A, Funkhouser T and
  Ferrari V (2022) Urban radiance fields.
\newblock In: \emph{CVPR}. pp. 12932--12942.

\bibitem[{Ren et~al.(2022)Ren, Agarwala, Russell, Schwing and
  Wang}]{ren2022neural}
Ren Z, Agarwala A, Russell B, Schwing AG and Wang O (2022) Neural volumetric
  object selection.
\newblock In: \emph{CVPR}. pp. 6133--6142.

\bibitem[{Roberts et~al.(2021)Roberts, Ramapuram, Ranjan, Kumar, Bautista,
  Paczan, Webb and Susskind}]{roberts2021hypersim}
Roberts M, Ramapuram J, Ranjan A, Kumar A, Bautista MA, Paczan N, Webb R and
  Susskind JM (2021) Hypersim: A photorealistic synthetic dataset for holistic
  indoor scene understanding.
\newblock In: \emph{ICCV}. pp. 10912--10922.

\bibitem[{Roessle et~al.(2022)Roessle, Barron, Mildenhall, Srinivasan and
  Nie{\ss}ner}]{roessle2022dense}
Roessle B, Barron JT, Mildenhall B, Srinivasan PP and Nie{\ss}ner M (2022)
  Dense depth priors for neural radiance fields from sparse input views.
\newblock In: \emph{CVPR}. pp. 12892--12901.

\bibitem[{Rosinol et~al.(2023)Rosinol, Leonard and Carlone}]{rosinol2022nerf}
Rosinol A, Leonard JJ and Carlone L (2023) Nerf-slam: Real-time dense monocular
  slam with neural radiance fields.
\newblock In: \emph{IROS}. IEEE, pp. 3437--3444.

\bibitem[{Rudnev et~al.(2022)Rudnev, Elgharib, Smith, Liu, Golyanik and
  Theobalt}]{rudnev2022nerf}
Rudnev V, Elgharib M, Smith W, Liu L, Golyanik V and Theobalt C (2022) Nerf for
  outdoor scene relighting.
\newblock In: \emph{ECCV}. pp. 615--631.

\bibitem[{Schmid et~al.(2024)Schmid, Von~Einem, Cadena, Siegwart, Hruby and
  Tschopp}]{schmid2024virus}
Schmid N, Von~Einem C, Cadena C, Siegwart R, Hruby L and Tschopp F (2024)
  Virus-nerf-vision, infrared and ultrasonic based neural radiance fields.
\newblock In: \emph{IROS}.

\bibitem[{Schonberger and Frahm(2016)}]{schonberger2016structure}
Schonberger JL and Frahm JM (2016) Structure-from-motion revisited.
\newblock In: \emph{CVPR}. pp. 4104--4113.

\bibitem[{Shafiullah et~al.(2023)Shafiullah, Paxton, Pinto, Chintala and
  Szlam}]{shafiullah2022clip}
Shafiullah NMM, Paxton C, Pinto L, Chintala S and Szlam A (2023) Clip-fields:
  Weakly supervised semantic fields for robotic memory.
\newblock \emph{RSS} .

\bibitem[{Shen et~al.(2022)Shen, Jiang, Choy, Guibas, Savarese, Anandkumar and
  Zhu}]{shen2022acid}
Shen B, Jiang Z, Choy C, Guibas LJ, Savarese S, Anandkumar A and Zhu Y (2022)
  Acid: Action-conditional implicit visual dynamics for deformable object
  manipulation.
\newblock \emph{RSS} .

\bibitem[{Shen et~al.(2023)Shen, Yang, Yu, Wong, Kaelbling and
  Isola}]{shen2023distilled}
Shen W, Yang G, Yu A, Wong J, Kaelbling LP and Isola P (2023) Distilled feature
  fields enable few-shot language-guided manipulation.
\newblock \emph{PMLR} .

\bibitem[{Shi et~al.(2022)Shi, Rong, Ni, Chen and Zhang}]{shi2022garf}
Shi Y, Rong D, Ni B, Chen C and Zhang W (2022) Garf: Geometry-aware generalized
  neural radiance field.
\newblock \emph{arXiv preprint arXiv:2212.02280} .

\bibitem[{Shim et~al.(2023)Shim, Lee and Kim}]{shim2023snerl}
Shim D, Lee S and Kim HJ (2023) Snerl: Semantic-aware neural radiance fields
  for reinforcement learning.
\newblock \emph{ICML} .

\bibitem[{Siddiqui et~al.(2023)Siddiqui, Porzi, Bul{\'o}, M{\"u}ller,
  Nie{\ss}ner, Dai and Kontschieder}]{siddiqui2023panoptic}
Siddiqui Y, Porzi L, Bul{\'o} SR, M{\"u}ller N, Nie{\ss}ner M, Dai A and
  Kontschieder P (2023) Panoptic lifting for 3d scene understanding with neural
  fields.
\newblock In: \emph{CVPR}. pp. 9043--9052.

\bibitem[{Simeonov et~al.(2023)Simeonov, Du, Lin, Garcia, Kaelbling,
  Lozano-P{\'e}rez and Agrawal}]{simeonov2023se}
Simeonov A, Du Y, Lin YC, Garcia AR, Kaelbling LP, Lozano-P{\'e}rez T and
  Agrawal P (2023) Se (3)-equivariant relational rearrangement with neural
  descriptor fields.
\newblock In: \emph{CoRL}. pp. 835--846.

\bibitem[{Simeonov et~al.(2022)Simeonov, Du, Tagliasacchi, Tenenbaum,
  Rodriguez, Agrawal and Sitzmann}]{simeonov2022neural}
Simeonov A, Du Y, Tagliasacchi A, Tenenbaum JB, Rodriguez A, Agrawal P and
  Sitzmann V (2022) Neural descriptor fields: Se (3)-equivariant object
  representations for manipulation.
\newblock In: \emph{ICRA}. pp. 6394--6400.

\bibitem[{Singer et~al.(2023)Singer, Polyak, Hayes, Yin, An, Zhang, Hu, Yang,
  Ashual, Gafni et~al.}]{singer2022make}
Singer U, Polyak A, Hayes T, Yin X, An J, Zhang S, Hu Q, Yang H, Ashual O,
  Gafni O et~al. (2023) Make-a-video: Text-to-video generation without
  text-video data.
\newblock \emph{ICLR} .

\bibitem[{Sitzmann et~al.(2020)Sitzmann, Martel, Bergman, Lindell and
  Wetzstein}]{sitzmann2020implicit}
Sitzmann V, Martel J, Bergman A, Lindell D and Wetzstein G (2020) Implicit
  neural representations with periodic activation functions.
\newblock \emph{NeurIPS} : 7462--7473.

\bibitem[{Sorkine and Alexa(2007)}]{sorkine2007rigid}
Sorkine O and Alexa M (2007) As-rigid-as-possible surface modeling.
\newblock In: \emph{SGP}. pp. 109--116.

\bibitem[{Srinivasan et~al.(2021)Srinivasan, Deng, Zhang, Tancik, Mildenhall
  and Barron}]{srinivasan2021nerv}
Srinivasan PP, Deng B, Zhang X, Tancik M, Mildenhall B and Barron JT (2021)
  Nerv: Neural reflectance and visibility fields for relighting and view
  synthesis.
\newblock In: \emph{CVPR}. pp. 7495--7504.

\bibitem[{Straub et~al.(2019)Straub, Whelan, Ma, Chen, Wijmans, Green, Engel,
  Mur-Artal, Ren, Verma et~al.}]{straub2019replica}
Straub J, Whelan T, Ma L, Chen Y, Wijmans E, Green S, Engel JJ, Mur-Artal R,
  Ren C, Verma S et~al. (2019) The replica dataset: A digital replica of indoor
  spaces.
\newblock \emph{arXiv preprint arXiv:1906.05797} .

\bibitem[{Sturm et~al.(2012)Sturm, Engelhard, Endres, Burgard and
  Cremers}]{sturm2012benchmark}
Sturm J, Engelhard N, Endres F, Burgard W and Cremers D (2012) A benchmark for
  the evaluation of rgb-d slam systems.
\newblock In: \emph{2012 IEEE/RSJ international conference on intelligent
  robots and systems}. IEEE, pp. 573--580.

\bibitem[{Sucar et~al.(2021)Sucar, Liu, Ortiz and Davison}]{sucar2021imap}
Sucar E, Liu S, Ortiz J and Davison AJ (2021) imap: Implicit mapping and
  positioning in real-time.
\newblock In: \emph{ICCV}. pp. 6229--6238.

\bibitem[{Sun et~al.(2022{\natexlab{a}})Sun, Sun and Chen}]{sun2022direct}
Sun C, Sun M and Chen HT (2022{\natexlab{a}}) Direct voxel grid optimization:
  Super-fast convergence for radiance fields reconstruction.
\newblock In: \emph{CVPR}. pp. 5459--5469.

\bibitem[{Sun et~al.(2022{\natexlab{b}})Sun, Chen, Wang, Li, Averbuch-Elor,
  Zhou and Snavely}]{sun2022neural}
Sun J, Chen X, Wang Q, Li Z, Averbuch-Elor H, Zhou X and Snavely N
  (2022{\natexlab{b}}) Neural 3d reconstruction in the wild.
\newblock In: \emph{ACM SIGGRAPH}. pp. 1--9.

\bibitem[{Sun et~al.(2024)Sun, Zhuang, Jiang, Liu, Xie and
  Chandraker}]{sun2024lidarf}
Sun S, Zhuang B, Jiang Z, Liu B, Xie X and Chandraker M (2024) Lidarf: Delving
  into lidar for neural radiance field on street scenes.
\newblock In: \emph{CVPR}. pp. 19563--19572.

\bibitem[{Suresh et~al.(2024)Suresh, Qi, Wu, Fan, Pineda, Lambeta, Malik,
  Kalakrishnan, Calandra, Kaess et~al.}]{suresh2024neuralfeels}
Suresh S, Qi H, Wu T, Fan T, Pineda L, Lambeta M, Malik J, Kalakrishnan M,
  Calandra R, Kaess M et~al. (2024) Neuralfeels with neural fields:
  Visuotactile perception for in-hand manipulation.
\newblock \emph{Science Robotics} .

\bibitem[{Tancik et~al.(2022)Tancik, Casser, Yan, Pradhan, Mildenhall,
  Srinivasan, Barron and Kretzschmar}]{tancik2022block}
Tancik M, Casser V, Yan X, Pradhan S, Mildenhall B, Srinivasan PP, Barron JT
  and Kretzschmar H (2022) Block-nerf: Scalable large scene neural view
  synthesis.
\newblock In: \emph{CVPR}. pp. 8248--8258.

\bibitem[{Tancik et~al.(2023)Tancik, Weber, Ng, Li, Yi, Wang, Kristoffersen,
  Austin, Salahi, Ahuja et~al.}]{tancik2023nerfstudio}
Tancik M, Weber E, Ng E, Li R, Yi B, Wang T, Kristoffersen A, Austin J, Salahi
  K, Ahuja A et~al. (2023) Nerfstudio: A modular framework for neural radiance
  field development.
\newblock In: \emph{ACM SIGGRAPH}. pp. 1--12.

\bibitem[{Tang et~al.(2023)Tang, Sundaralingam, Tremblay, Wen, Yuan, Tyree,
  Loop, Schwing and Birchfield}]{tang2023rgb}
Tang Z, Sundaralingam B, Tremblay J, Wen B, Yuan Y, Tyree S, Loop C, Schwing A
  and Birchfield S (2023) Rgb-only reconstruction of tabletop scenes for
  collision-free manipulator control.
\newblock In: \emph{ICRA}. pp. 1778--1785.

\bibitem[{Tao et~al.(2024)Tao, Gao, Wang, Lao, Chen, Zhao, Hao, Liang, Salzmann
  and Yu}]{tao2024lidar}
Tao T, Gao L, Wang G, Lao Y, Chen P, Zhao H, Hao D, Liang X, Salzmann M and Yu
  K (2024) Lidar-nerf: Novel lidar view synthesis via neural radiance fields.
\newblock In: \emph{ACM MM}. pp. 390--398.

\bibitem[{Tertikas et~al.(2023)Tertikas, Despoina, Pan, Park, Uy, Emiris,
  Avrithis and Guibas}]{tertikas2023partnerf}
Tertikas K, Despoina P, Pan B, Park JJ, Uy MA, Emiris I, Avrithis Y and Guibas
  L (2023) Partnerf: Generating part-aware editable 3d shapes without 3d
  supervision.
\newblock \emph{CVPR} .

\bibitem[{Tewari et~al.(2020)Tewari, Fried, Thies, Sitzmann, Lombardi,
  Sunkavalli, Martin-Brualla, Simon, Saragih, Nie{\ss}ner
  et~al.}]{tewari2020state}
Tewari A, Fried O, Thies J, Sitzmann V, Lombardi S, Sunkavalli K,
  Martin-Brualla R, Simon T, Saragih J, Nie{\ss}ner M et~al. (2020) State of
  the art on neural rendering.
\newblock In: \emph{CGF}.

\bibitem[{Tewari et~al.(2022)Tewari, Thies, Mildenhall, Srinivasan, Tretschk,
  Yifan, Lassner, Sitzmann, Martin-Brualla, Lombardi
  et~al.}]{tewari2022advances}
Tewari A, Thies J, Mildenhall B, Srinivasan P, Tretschk E, Yifan W, Lassner C,
  Sitzmann V, Martin-Brualla R, Lombardi S et~al. (2022) Advances in neural
  rendering.
\newblock In: \emph{CGF}.

\bibitem[{Thies et~al.(2016)Thies, Zollhofer, Stamminger, Theobalt and
  Nie{\ss}ner}]{thies2016face2face}
Thies J, Zollhofer M, Stamminger M, Theobalt C and Nie{\ss}ner M (2016)
  Face2face: Real-time face capture and reenactment of rgb videos.
\newblock In: \emph{CVPR}. pp. 2387--2395.

\bibitem[{Tong et~al.(2022)Tong, Dawson and Fan}]{tong2022enforcing}
Tong M, Dawson C and Fan C (2022) Enforcing safety for vision-based controllers
  via control barrier functions and neural radiance fields.
\newblock \emph{ICRA} .

\bibitem[{Torne et~al.(2024)Torne, Simeonov, Li, Chan, Chen, Gupta and
  Agrawal}]{torne2024reconciling}
Torne M, Simeonov A, Li Z, Chan A, Chen T, Gupta A and Agrawal P (2024)
  Reconciling reality through simulation: A real-to-sim-to-real approach for
  robust manipulation.
\newblock \emph{arXiv preprint arXiv:2403.03949} .

\bibitem[{Tremblay et~al.(2023)Tremblay, Wen, Blukis, Sundaralingam, Tyree and
  Birchfield}]{tremblay2023diff}
Tremblay J, Wen B, Blukis V, Sundaralingam B, Tyree S and Birchfield S (2023)
  Diff-dope: Differentiable deep object pose estimation.
\newblock \emph{arXiv preprint arXiv:2310.00463} .

\bibitem[{Tretschk et~al.(2021)Tretschk, Tewari, Golyanik, Zollh{\"o}fer,
  Lassner and Theobalt}]{tretschk2021non}
Tretschk E, Tewari A, Golyanik V, Zollh{\"o}fer M, Lassner C and Theobalt C
  (2021) Non-rigid neural radiance fields: Reconstruction and novel view
  synthesis of a dynamic scene from monocular video.
\newblock In: \emph{ICCV}. pp. 12959--12970.

\bibitem[{Trevithick and Yang(2021)}]{trevithick2021grf}
Trevithick A and Yang B (2021) Grf: Learning a general radiance field for 3d
  representation and rendering.
\newblock In: \emph{ICCV}. pp. 15182--15192.

\bibitem[{Truong et~al.(2023)Truong, Rakotosaona, Manhardt and
  Tombari}]{truong2023sparf}
Truong P, Rakotosaona MJ, Manhardt F and Tombari F (2023) Sparf: Neural
  radiance fields from sparse and noisy poses.
\newblock In: \emph{CVPR}. pp. 4190--4200.

\bibitem[{Tschernezki et~al.(2022)Tschernezki, Laina, Larlus and
  Vedaldi}]{tschernezki2022neural}
Tschernezki V, Laina I, Larlus D and Vedaldi A (2022) Neural feature fusion
  fields: 3d distillation of self-supervised 2d image representations.
\newblock \emph{3DV} .

\bibitem[{Tschernezki et~al.(2021)Tschernezki, Larlus and
  Vedaldi}]{tschernezki2021neuraldiff}
Tschernezki V, Larlus D and Vedaldi A (2021) Neuraldiff: Segmenting 3d objects
  that move in egocentric videos.
\newblock In: \emph{3DV}. pp. 910--919.

\bibitem[{Tseng et~al.(2022)Tseng, Liao, Yen-Chen and Sun}]{tseng2022cla}
Tseng WC, Liao HJ, Yen-Chen L and Sun M (2022) Cla-nerf: Category-level
  articulated neural radiance field.
\newblock In: \emph{ICRA}. pp. 8454--8460.

\bibitem[{Turki et~al.(2022)Turki, Ramanan and
  Satyanarayanan}]{Turki_2022_CVPR}
Turki H, Ramanan D and Satyanarayanan M (2022) Mega-nerf: Scalable construction
  of large-scale nerfs for virtual fly-throughs.
\newblock In: \emph{CVPR}.

\bibitem[{Turki et~al.(2023)Turki, Zhang, Ferroni and Ramanan}]{turki2023suds}
Turki H, Zhang JY, Ferroni F and Ramanan D (2023) Suds: Scalable urban dynamic
  scenes.
\newblock In: \emph{CVPR}. pp. 12375--12385.

\bibitem[{Varma et~al.(2023)Varma, Wang, Chen, Chen, Venugopalan and
  Wang}]{varma2022attention}
Varma M, Wang P, Chen X, Chen T, Venugopalan S and Wang Z (2023) Is attention
  all that nerf needs?
\newblock In: \emph{ICLR}.

\bibitem[{Vaswani et~al.(2017)Vaswani, Shazeer, Parmar, Uszkoreit, Jones,
  Gomez, Kaiser and Polosukhin}]{vaswani2017attention}
Vaswani A, Shazeer N, Parmar N, Uszkoreit J, Jones L, Gomez AN, Kaiser {\L} and
  Polosukhin I (2017) Attention is all you need.
\newblock \emph{NeurIPS} 30.

\bibitem[{Vora et~al.(2022)Vora, Radwan, Greff, Meyer, Genova, Sajjadi, Pot,
  Tagliasacchi and Duckworth}]{vora2021nesf}
Vora S, Radwan N, Greff K, Meyer H, Genova K, Sajjadi MS, Pot E, Tagliasacchi A
  and Duckworth D (2022) Nesf: Neural semantic fields for generalizable
  semantic segmentation of 3d scenes.
\newblock \emph{TMLR} .

\bibitem[{Wang et~al.(2022{\natexlab{a}})Wang, Chai, He, Chen and
  Liao}]{wang2022clip}
Wang C, Chai M, He M, Chen D and Liao J (2022{\natexlab{a}}) Clip-nerf:
  Text-and-image driven manipulation of neural radiance fields.
\newblock In: \emph{CVPR}. pp. 3835--3844.

\bibitem[{Wang et~al.(2023{\natexlab{a}})Wang, Jiang, Chai, He, Chen and
  Liao}]{wang2022nerf}
Wang C, Jiang R, Chai M, He M, Chen D and Liao J (2023{\natexlab{a}}) Nerf-art:
  Text-driven neural radiance fields stylization.
\newblock \emph{TVCG} .

\bibitem[{Wang et~al.(2021{\natexlab{a}})Wang, Liu, Liu, Theobalt, Komura and
  Wang}]{wang2021neus}
Wang P, Liu L, Liu Y, Theobalt C, Komura T and Wang W (2021{\natexlab{a}})
  Neus: Learning neural implicit surfaces by volume rendering for multi-view
  reconstruction.
\newblock \emph{(NeurIPS} .

\bibitem[{Wang et~al.(2021{\natexlab{b}})Wang, Wang, Genova, Srinivasan, Zhou,
  Barron, Martin-Brualla, Snavely and Funkhouser}]{wang2021ibrnet}
Wang Q, Wang Z, Genova K, Srinivasan PP, Zhou H, Barron JT, Martin-Brualla R,
  Snavely N and Funkhouser T (2021{\natexlab{b}}) Ibrnet: Learning multi-view
  image-based rendering.
\newblock In: \emph{CVPR}. pp. 4690--4699.

\bibitem[{Wang et~al.(2022{\natexlab{b}})Wang, Morgan, Dollar and
  Hager}]{wang2022dynamical}
Wang W, Morgan AS, Dollar AM and Hager GD (2022{\natexlab{b}}) Dynamical scene
  representation and control with keypoint-conditioned neural radiance field.
\newblock In: \emph{CASE}. pp. 1138--1143.

\bibitem[{Wang et~al.(2024)Wang, Chen and Lee}]{wang2024gov}
Wang Y, Chen H and Lee GH (2024) Gov-nesf: Generalizable open-vocabulary neural
  semantic fields.
\newblock In: \emph{CVPR}. pp. 20443--20453.

\bibitem[{Wang et~al.(2004)Wang, Bovik, Sheikh and Simoncelli}]{wang2004image}
Wang Z, Bovik AC, Sheikh HR and Simoncelli EP (2004) Image quality assessment:
  from error visibility to structural similarity.
\newblock \emph{IEEE transactions on image processing} 13(4): 600--612.

\bibitem[{Wang et~al.(2023{\natexlab{b}})Wang, Shen, Gao, Huang, Munkberg,
  Hasselgren, Gojcic, Chen and Fidler}]{wang2023neural}
Wang Z, Shen T, Gao J, Huang S, Munkberg J, Hasselgren J, Gojcic Z, Chen W and
  Fidler S (2023{\natexlab{b}}) Neural fields meet explicit geometric
  representations for inverse rendering of urban scenes.
\newblock In: \emph{CVPR}. pp. 8370--8380.

\bibitem[{Wang et~al.(2021{\natexlab{c}})Wang, Wu, Xie, Chen and
  Prisacariu}]{wang2021nerf}
Wang Z, Wu S, Xie W, Chen M and Prisacariu VA (2021{\natexlab{c}}) Ne{RF}$--$:
  Neural radiance fields without known camera parameters.
\newblock \emph{arXiv preprint arXiv:2102.07064} .

\bibitem[{Wani et~al.(2020)Wani, Patel, Jain, Chang and
  Savva}]{wani2020multion}
Wani S, Patel S, Jain U, Chang A and Savva M (2020) Multion: Benchmarking
  semantic map memory using multi-object navigation.
\newblock \emph{NeurIPS} 33: 9700--9712.

\bibitem[{Weder et~al.(2023)Weder, Garcia-Hernando, Monszpart, Pollefeys,
  Brostow, Firman and Vicente}]{weder2023removing}
Weder S, Garcia-Hernando G, Monszpart {\'{A}}, Pollefeys M, Brostow G, Firman M
  and Vicente S (2023) Removing objects from {NeRFs}.
\newblock In: \emph{CVPR}.

\bibitem[{Wen et~al.(2023)Wen, Tremblay, Blukis, Tyree, M{\"u}ller, Evans, Fox,
  Kautz and Birchfield}]{wen2023bundlesdf}
Wen B, Tremblay J, Blukis V, Tyree S, M{\"u}ller T, Evans A, Fox D, Kautz J and
  Birchfield S (2023) Bundlesdf: Neural 6-dof tracking and 3d reconstruction of
  unknown objects.
\newblock In: \emph{CVPR}. pp. 606--617.

\bibitem[{Weng et~al.(2023)Weng, Held, Meier and Mukadam}]{weng2023neural}
Weng T, Held D, Meier F and Mukadam M (2023) Neural grasp distance fields for
  robot manipulation.
\newblock In: \emph{ICRA}. pp. 1814--1821.

\bibitem[{Wu et~al.(2024)Wu, Mildenhall, Henzler, Park, Gao, Watson,
  Srinivasan, Verbin, Barron, Poole et~al.}]{wu2024reconfusion}
Wu R, Mildenhall B, Henzler P, Park K, Gao R, Watson D, Srinivasan PP, Verbin
  D, Barron JT, Poole B et~al. (2024) Reconfusion: 3d reconstruction with
  diffusion priors.
\newblock In: \emph{CVPR}. pp. 21551--21561.

\bibitem[{Wu et~al.(2022)Wu, Zhong, Tagliasacchi, Cole and Oztireli}]{wu2022d}
Wu T, Zhong F, Tagliasacchi A, Cole F and Oztireli C (2022) D\^{} 2nerf:
  Self-supervised decoupling of dynamic and static objects from a monocular
  video.
\newblock \emph{NeurIPS} : 32653--32666.

\bibitem[{Wu et~al.(2025)Wu, Pan, Wu, Wang, Miao, Xu and Wang}]{wu2024rl}
Wu Y, Pan L, Wu W, Wang G, Miao Y, Xu F and Wang H (2025) Rl-gsbridge: 3d
  gaussian splatting based real2sim2real method for robotic manipulation
  learning.
\newblock \emph{ICRA} .

\bibitem[{Xia et~al.(2022)Xia, Tang, Timofte and Gool}]{xia2022sinerf}
Xia Y, Tang H, Timofte R and Gool LV (2022) Sinerf: Sinusoidal neural radiance
  fields for joint pose estimation and scene reconstruction.
\newblock In: \emph{BMVC}.

\bibitem[{Xian et~al.(2021)Xian, Huang, Kopf and Kim}]{xian2021space}
Xian W, Huang JB, Kopf J and Kim C (2021) Space-time neural irradiance fields
  for free-viewpoint video.
\newblock In: \emph{CVPR}. pp. 9421--9431.

\bibitem[{Xiangli et~al.(2022)Xiangli, Xu, Pan, Zhao, Rao, Theobalt, Dai and
  Lin}]{xiangli2022bungeenerf}
Xiangli Y, Xu L, Pan X, Zhao N, Rao A, Theobalt C, Dai B and Lin D (2022)
  Bungeenerf: Progressive neural radiance field for extreme multi-scale scene
  rendering.
\newblock In: \emph{ECCV}. pp. 106--122.

\bibitem[{Xie et~al.(2024)Xie, Zong, Qiu, Li, Feng, Yang and
  Jiang}]{xie2023physgaussian}
Xie T, Zong Z, Qiu Y, Li X, Feng Y, Yang Y and Jiang C (2024) Physgaussian:
  Physics-integrated 3d gaussians for generative dynamics.
\newblock \emph{CVPR} .

\bibitem[{Xie et~al.(2022)Xie, Takikawa, Saito, Litany, Yan, Khan, Tombari,
  Tompkin, Sitzmann and Sridhar}]{xie2022neural}
Xie Y, Takikawa T, Saito S, Litany O, Yan S, Khan N, Tombari F, Tompkin J,
  Sitzmann V and Sridhar S (2022) Neural fields in visual computing and beyond.
\newblock In: \emph{CGF}. pp. 641--676.

\bibitem[{Xie et~al.(2023)Xie, Zhang, Li, Zhang and Zhang}]{xie2023s}
Xie Z, Zhang J, Li W, Zhang F and Zhang L (2023) S-nerf: Neural radiance fields
  for street views.
\newblock \emph{ICLR} .

\bibitem[{Xu et~al.(2023)Xu, Wu, Hou, Tsai, Li, Wang, Zhan, He, Vajda, Keutzer
  et~al.}]{xu2023nerf}
Xu C, Wu B, Hou J, Tsai S, Li R, Wang J, Zhan W, He Z, Vajda P, Keutzer K
  et~al. (2023) Nerf-det: Learning geometry-aware volumetric representation for
  multi-view 3d object detection.
\newblock In: \emph{CVPR}. pp. 23320--23330.

\bibitem[{Xu et~al.(2022{\natexlab{a}})Xu, Jiang, Wang, Fan, Shi and
  Wang}]{xu2022sinnerf}
Xu D, Jiang Y, Wang P, Fan Z, Shi H and Wang Z (2022{\natexlab{a}}) Sinnerf:
  Training neural radiance fields on complex scenes from a single image.
\newblock \emph{ECCV} .

\bibitem[{Xu et~al.(2024{\natexlab{a}})Xu, Chen, Chen, Sakaridis, Zhang,
  Pollefeys, Geiger and Yu}]{xu2024murf}
Xu H, Chen A, Chen Y, Sakaridis C, Zhang Y, Pollefeys M, Geiger A and Yu F
  (2024{\natexlab{a}}) Murf: multi-baseline radiance fields.
\newblock In: \emph{CVPR}. pp. 20041--20050.

\bibitem[{Xu et~al.(2024{\natexlab{b}})Xu, Liao, Kathirvel and
  Patel}]{xu2024leveraging}
Xu J, Liao M, Kathirvel RP and Patel VM (2024{\natexlab{b}}) Leveraging thermal
  modality to enhance reconstruction in low-light conditions.
\newblock In: \emph{ECCV}.

\bibitem[{Xu et~al.(2022{\natexlab{b}})Xu, Xu, Philip, Bi, Shu, Sunkavalli and
  Neumann}]{xu2022point}
Xu Q, Xu Z, Philip J, Bi S, Shu Z, Sunkavalli K and Neumann U
  (2022{\natexlab{b}}) Point-nerf: Point-based neural radiance fields.
\newblock In: \emph{CVPR}. pp. 5438--5448.

\bibitem[{Xu and Harada(2022)}]{xu2022deforming}
Xu T and Harada T (2022) Deforming radiance fields with cages.
\newblock In: \emph{ECCV}. pp. 159--175.

\bibitem[{Xue et~al.(2024)Xue, Zheng, Lu, Wei, Chen et~al.}]{xue2024geonlf}
Xue W, Zheng Z, Lu F, Wei H, Chen G et~al. (2024) Geonlf: Geometry guided
  pose-free neural lidar fields.
\newblock \emph{NeurIPS} .

\bibitem[{Yan et~al.(2023)Yan, Li and Lee}]{yan2023nerf}
Yan Z, Li C and Lee GH (2023) Nerf-ds: Neural radiance fields for dynamic
  specular objects.
\newblock In: \emph{CVPR}. pp. 8285--8295.

\bibitem[{Yang et~al.(2022{\natexlab{a}})Yang, Bao, Zeng, Bao, Zhang, Cui and
  Zhang}]{yang2022neumesh}
Yang B, Bao C, Zeng J, Bao H, Zhang Y, Cui Z and Zhang G (2022{\natexlab{a}})
  Neumesh: Learning disentangled neural mesh-based implicit field for geometry
  and texture editing.
\newblock In: \emph{ECCV}. pp. 597--614.

\bibitem[{Yang et~al.(2021)Yang, Zhang, Xu, Li, Zhou, Bao, Zhang and
  Cui}]{yang2021learning}
Yang B, Zhang Y, Xu Y, Li Y, Zhou H, Bao H, Zhang G and Cui Z (2021) Learning
  object-compositional neural radiance field for editable scene rendering.
\newblock In: \emph{ICCV}. pp. 13779--13788.

\bibitem[{Yang et~al.(2023)Yang, Ivanovic, Litany, Weng, Kim, Li, Che, Xu,
  Fidler, Pavone et~al.}]{yang2023emernerf}
Yang J, Ivanovic B, Litany O, Weng X, Kim SW, Li B, Che T, Xu D, Fidler S,
  Pavone M et~al. (2023) Emernerf: Emergent spatial-temporal scene
  decomposition via self-supervision.
\newblock \emph{arXiv preprint arXiv:2311.02077} .

\bibitem[{Yang et~al.(2022{\natexlab{b}})Yang, Li, Zhai, Ming, Liu and
  Zhang}]{yang2022vox}
Yang X, Li H, Zhai H, Ming Y, Liu Y and Zhang G (2022{\natexlab{b}})
  Vox-fusion: Dense tracking and mapping with voxel-based neural implicit
  representation.
\newblock In: \emph{ISMAR}. pp. 499--507.

\bibitem[{Yariv et~al.(2021)Yariv, Gu, Kasten and Lipman}]{yariv2021volume}
Yariv L, Gu J, Kasten Y and Lipman Y (2021) Volume rendering of neural implicit
  surfaces.
\newblock \emph{NeurIPS} : 4805--4815.

\bibitem[{Ye et~al.(2024)Ye, Wu, Deng, Liu, Liu, Xia, Pang, Yu and
  Pei}]{ye2024thermal}
Ye T, Wu Q, Deng J, Liu G, Liu L, Xia S, Pang L, Yu W and Pei L (2024)
  Thermal-nerf: Neural radiance fields from an infrared camera.
\newblock In: \emph{IROS}.

\bibitem[{Yen-Chen et~al.(2022)Yen-Chen, Florence, Barron, Lin, Rodriguez and
  Isola}]{yen2022nerf}
Yen-Chen L, Florence P, Barron JT, Lin TY, Rodriguez A and Isola P (2022)
  Nerf-supervision: Learning dense object descriptors from neural radiance
  fields.
\newblock \emph{ICRA} .

\bibitem[{Yen-Chen et~al.(2021)Yen-Chen, Florence, Barron, Rodriguez, Isola and
  Lin}]{yen2021inerf}
Yen-Chen L, Florence P, Barron JT, Rodriguez A, Isola P and Lin TY (2021)
  inerf: Inverting neural radiance fields for pose estimation.
\newblock In: \emph{IROS}. pp. 1323--1330.

\bibitem[{Yoon et~al.(2020)Yoon, Kim, Gallo, Park and Kautz}]{yoon2020novel}
Yoon JS, Kim K, Gallo O, Park HS and Kautz J (2020) Novel view synthesis of
  dynamic scenes with globally coherent depths from a monocular camera.
\newblock In: \emph{CVPR}. pp. 5336--5345.

\bibitem[{You and Hou(2024)}]{you2023decoupling}
You M and Hou J (2024) Decoupling dynamic monocular videos for dynamic view
  synthesis.
\newblock \emph{T-VCG} .

\bibitem[{Yu et~al.(2021{\natexlab{a}})Yu, Li, Tancik, Li, Ng and
  Kanazawa}]{yu2021plenoctrees}
Yu A, Li R, Tancik M, Li H, Ng R and Kanazawa A (2021{\natexlab{a}})
  Plenoctrees for real-time rendering of neural radiance fields.
\newblock In: \emph{ICCV}. pp. 5752--5761.

\bibitem[{Yu et~al.(2021{\natexlab{b}})Yu, Ye, Tancik and
  Kanazawa}]{yu2021pixelnerf}
Yu A, Ye V, Tancik M and Kanazawa A (2021{\natexlab{b}}) pixelnerf: Neural
  radiance fields from one or few images.
\newblock In: \emph{CVPR}. pp. 4578--4587.

\bibitem[{Yu et~al.(2022{\natexlab{a}})Yu, Guibas and Wu}]{yu2021unsupervised}
Yu HX, Guibas LJ and Wu J (2022{\natexlab{a}}) Unsupervised discovery of object
  radiance fields.
\newblock \emph{ICLR} .

\bibitem[{Yu et~al.(2025)Yu, Chen and Schwager}]{yu2025hammer}
Yu J, Chen T and Schwager M (2025) Hammer: Heterogeneous, multi-robot semantic
  gaussian splatting.
\newblock \emph{arXiv preprint arXiv:2501.14147} .

\bibitem[{Yu et~al.(2023)Yu, Low, Nagami and Schwager}]{yu2023nerfbridge}
Yu J, Low JE, Nagami K and Schwager M (2023) Nerfbridge: Bringing real-time,
  online neural radiance field training to robotics.
\newblock \emph{ICRA Workshop} .

\bibitem[{Yu et~al.(2022{\natexlab{b}})Yu, Peng, Niemeyer, Sattler and
  Geiger}]{yu2022monosdf}
Yu Z, Peng S, Niemeyer M, Sattler T and Geiger A (2022{\natexlab{b}}) Monosdf:
  Exploring monocular geometric cues for neural implicit surface
  reconstruction.
\newblock \emph{NeurIPS} : 25018--25032.

\bibitem[{Yuan et~al.(2021)Yuan, Lv, Schmidt and Lovegrove}]{yuan2021star}
Yuan W, Lv Z, Schmidt T and Lovegrove S (2021) Star: Self-supervised tracking
  and reconstruction of rigid objects in motion with neural rendering.
\newblock In: \emph{CVPR}. pp. 13144--13152.

\bibitem[{Yuan et~al.(2022{\natexlab{a}})Yuan, Lai, Huang, Kobbelt and
  Gao}]{yuan2022neural}
Yuan YJ, Lai YK, Huang YH, Kobbelt L and Gao L (2022{\natexlab{a}}) Neural
  radiance fields from sparse rgb-d images for high-quality view synthesis.
\newblock \emph{TPAMI} .

\bibitem[{Yuan et~al.(2022{\natexlab{b}})Yuan, Sun, Lai, Ma, Jia and
  Gao}]{yuan2022nerf}
Yuan YJ, Sun YT, Lai YK, Ma Y, Jia R and Gao L (2022{\natexlab{b}})
  Nerf-editing: geometry editing of neural radiance fields.
\newblock In: \emph{CVPR}. pp. 18353--18364.

\bibitem[{Zeng et~al.(2023)Zeng, Li, Ran, Li, Gao, Li, He, Chen and
  Ye}]{zeng2023efficient}
Zeng J, Li Y, Ran Y, Li S, Gao F, Li L, He S, Chen J and Ye Q (2023) Efficient
  view path planning for autonomous implicit reconstruction.
\newblock In: \emph{ICRA}. pp. 4063--4069.

\bibitem[{Zhang et~al.(2024{\natexlab{a}})Zhang, Sandstr{\"o}m, Zhang, Patel,
  Van~Gool and Oswald}]{zhang2024glorie}
Zhang G, Sandstr{\"o}m E, Zhang Y, Patel M, Van~Gool L and Oswald MR
  (2024{\natexlab{a}}) Glorie-slam: Globally optimized rgb-only implicit
  encoding point cloud slam.
\newblock \emph{arXiv preprint arXiv:2403.19549} .

\bibitem[{Zhang et~al.(2024{\natexlab{b}})Zhang, Zhang, Kuang and
  Zhang}]{zhang2024nerf}
Zhang J, Zhang F, Kuang S and Zhang L (2024{\natexlab{b}}) Nerf-lidar:
  Generating realistic lidar point clouds with neural radiance fields.
\newblock In: \emph{AAAI}.

\bibitem[{Zhang et~al.(2021{\natexlab{a}})Zhang, Luan, Wang, Bala and
  Snavely}]{zhang2021physg}
Zhang K, Luan F, Wang Q, Bala K and Snavely N (2021{\natexlab{a}}) Physg:
  Inverse rendering with spherical gaussians for physics-based material editing
  and relighting.
\newblock In: \emph{CVPR}. pp. 5453--5462.

\bibitem[{Zhang et~al.(2020)Zhang, Riegler, Snavely and
  Koltun}]{zhang2020nerf++}
Zhang K, Riegler G, Snavely N and Koltun V (2020) Nerf++: Analyzing and
  improving neural radiance fields.
\newblock \emph{arXiv preprint arXiv:2010.07492} .

\bibitem[{Zhang et~al.(2018)Zhang, Isola, Efros, Shechtman and
  Wang}]{zhang2018unreasonable}
Zhang R, Isola P, Efros AA, Shechtman E and Wang O (2018) The unreasonable
  effectiveness of deep features as a perceptual metric.
\newblock In: \emph{CVPR}. pp. 586--595.

\bibitem[{Zhang et~al.(2021{\natexlab{b}})Zhang, Srinivasan, Deng, Debevec,
  Freeman and Barron}]{zhang2021nerfactor}
Zhang X, Srinivasan PP, Deng B, Debevec P, Freeman WT and Barron JT
  (2021{\natexlab{b}}) Nerfactor: Neural factorization of shape and reflectance
  under an unknown illumination.
\newblock \emph{ACM TOG} : 1--18.

\bibitem[{Zhao et~al.(2024{\natexlab{a}})Zhao, Yang, Mao, Bao and
  Cui}]{zhao2024pnerfloc}
Zhao B, Yang L, Mao M, Bao H and Cui Z (2024{\natexlab{a}}) Pnerfloc: Visual
  localization with point-based neural radiance fields.
\newblock In: \emph{AAAI}.

\bibitem[{Zhao et~al.(2024{\natexlab{b}})Zhao, Ivanovic and
  Mehr}]{zhao2024distributed}
Zhao H, Ivanovic B and Mehr N (2024{\natexlab{b}}) Distributed nerf learning
  for collaborative multi-robot perception.
\newblock \emph{arXiv preprint arXiv:2409.20289} .

\bibitem[{Zhao et~al.(2025)Zhao, Ivanovic and Mehr}]{zhao2025ramen}
Zhao H, Ivanovic B and Mehr N (2025) Ramen: Real-time asynchronous multi-agent
  neural implicit mapping.
\newblock \emph{arXiv preprint arXiv:2502.19592} .

\bibitem[{Zhi et~al.(2021{\natexlab{a}})Zhi, Laidlow, Leutenegger and
  Davison}]{zhi2021place}
Zhi S, Laidlow T, Leutenegger S and Davison AJ (2021{\natexlab{a}}) In-place
  scene labelling and understanding with implicit scene representation.
\newblock In: \emph{ICCV}. pp. 15838--15847.

\bibitem[{Zhi et~al.(2021{\natexlab{b}})Zhi, Sucar, Mouton, Haughton, Laidlow
  and Davison}]{zhi2021ilabel}
Zhi S, Sucar E, Mouton A, Haughton I, Laidlow T and Davison AJ
  (2021{\natexlab{b}}) ilabel: Interactive neural scene labelling.
\newblock \emph{arXiv preprint arXiv:2111.14637} .

\bibitem[{Zhong et~al.(2023)Zhong, Albini, Jones, Maiolino and
  Posner}]{zhong2023touching}
Zhong S, Albini A, Jones OP, Maiolino P and Posner I (2023) Touching a nerf:
  Leveraging neural radiance fields for tactile sensory data generation.
\newblock In: \emph{CoRL}. pp. 1618--1628.

\bibitem[{Zhou et~al.(2023)Zhou, Kim, Wang, Florence and Finn}]{zhou2023nerf}
Zhou A, Kim MJ, Wang L, Florence P and Finn C (2023) Nerf in the palm of your
  hand: Corrective augmentation for robotics via novel-view synthesis.
\newblock In: \emph{PCVPR}. pp. 17907--17917.

\bibitem[{Zhu et~al.(2024)Zhu, Wang, Blum, Liu, Song, Pollefeys and
  Wang}]{zhu2023sni}
Zhu S, Wang G, Blum H, Liu J, Song L, Pollefeys M and Wang H (2024) Sni-slam:
  Semantic neural implicit slam.
\newblock \emph{CVPR} .

\bibitem[{Zhu et~al.(2022{\natexlab{a}})Zhu, Chen, Wu, Hou, Shi, Li, Li, Zhao
  and Zhou}]{zhu2022latitude}
Zhu Z, Chen Y, Wu Z, Hou C, Shi Y, Li C, Li P, Zhao H and Zhou G
  (2022{\natexlab{a}}) Latitude: Robotic global localization with truncated
  dynamic low-pass filter in city-scale nerf.
\newblock \emph{ICRA} .

\bibitem[{Zhu et~al.(2023)Zhu, Peng, Larsson, Cui, Oswald, Geiger and
  Pollefeys}]{zhu2023nicer}
Zhu Z, Peng S, Larsson V, Cui Z, Oswald MR, Geiger A and Pollefeys M (2023)
  Nicer-slam: Neural implicit scene encoding for rgb slam.
\newblock \emph{3DV} .

\bibitem[{Zhu et~al.(2022{\natexlab{b}})Zhu, Peng, Larsson, Xu, Bao, Cui,
  Oswald and Pollefeys}]{zhu2022nice}
Zhu Z, Peng S, Larsson V, Xu W, Bao H, Cui Z, Oswald MR and Pollefeys M
  (2022{\natexlab{b}}) Nice-slam: Neural implicit scalable encoding for slam.
\newblock In: \emph{CVPR}. pp. 12786--12796.

\end{thebibliography}

\end{document}